\documentclass[ba]{imsart}
\pubyear{TBA}
\volume{TBA}
\issue{TBA}
\arxiv{2010.00000}
\firstpage{1}
\lastpage{1}

\usepackage{bbm}
\usepackage{amsmath,comment} 
\usepackage{amsfonts} 
\usepackage{amsthm}
\usepackage{algorithm}
\usepackage{algpseudocode}
\usepackage{natbib}
\usepackage[colorlinks,citecolor=blue,urlcolor=blue,filecolor=blue,backref=page]{hyperref}
\usepackage{graphicx}
\usepackage{subcaption}

\startlocaldefs
\numberwithin{equation}{section}
\theoremstyle{plain}

\endlocaldefs

\newtheorem{theorem}{Theorem}[section]
\newtheorem{corollary}{Corollary}[theorem]
\newtheorem{lemma}[theorem]{Lemma}

\newtheorem{definition}{Definition}
\newcommand{\con}{{\,\vert \,}}

\DeclareMathOperator*{\argmax}{arg\,max}

\def\m{\mathcal}

\def\T{{\mathrm{\scriptscriptstyle T} }}

\newcommand{\be}{\begin{equs}}
\newcommand{\ee}{\end{equs}}

\numberwithin{equation}{section}
\theoremstyle{plain}

\setcitestyle{authoryear,open={(},close={)}} 

\begin{document}

\begin{frontmatter}
\title{Differentially private Bayesian tests}
\runtitle{Differentially private Bayesian tests}

\begin{aug}
\author{\fnms{Abhisek} \snm{Chakraborty}\thanksref{}\ead[label=]{first@somewhere.com}},
\author{\fnms{Saptati} \snm{Datta}\thanksref{}\ead[label=]{second@somewhere.com}}

\runauthor{Chakraborty and Datta}

\address[addr1]{Department of Statistics, Texas A\&M University, 
College Station, TX, USA
}

\thankstext{t1}{Both the authors contributed equally.}
\thankstext{t2}{There was no external or internal funding for this project.}

\end{aug}

\begin{abstract}
Differential privacy has emerged as a significant cornerstone in the realm of scientific hypothesis testing utilizing confidential data. When data are not confidential, Bayesian tests are widely used in reporting scientific discoveries, as they effectively address the key criticisms of p-values, namely lack of interpretability and inability to quantify evidence in support of competing hypotheses. In this article, we introduce a novel framework for differentially private Bayesian hypothesis testing, thereby expanding the applicability of Bayesian testing to confidential data. This framework naturally arises from a principled data-generative mechanism, ensuring that the resulting inferences retain interpretability while maintaining privacy. Further, by focusing on differentially private Bayes factors based on  test statistics, we circumvent the need to model the complete data generative mechanism and ensure substantial computational benefits. We also  provide a set of sufficient conditions to establish Bayes factor consistency under the proposed framework. Finally, the utility of the proposed methodology is showcased via several numerical experiments.
\end{abstract}

\begin{keyword}[class=MSC]
\kwd[Primary ]{60K35}
\kwd{60K35}
\kwd[; secondary ]{60K35}
\end{keyword}

\begin{keyword}
\kwd{Bayes Factors}
\kwd{Bayes Factor Consistency}
\kwd{Differential Privacy}
\kwd{Interpretable ML}
\kwd{Laplace Mechanism}
\end{keyword}

\end{frontmatter}

\section{Introduction}\label{sec:intro}
Hypothesis testing is an indispensable tool to answer scientific questions in the context of clinical trials, bioinformatics, social sciences, etc. The data within such domains often involves sensitive and private information pertaining to individuals. Researchers often bear legal obligations to safeguard the privacy of such data. In this context, differential privacy \citep{dwork2006differential} has emerged as a popular framework for ensuring privacy in statistical analyses with confidential data. Consequently, differentially private versions of numerous commonly used hypothesis tests have been developed, although exclusively from a frequentist view point. This  encompasses private adaptations of test of binomial proportions \citep{10.5555/3327144.3327334}, significance in linear regression \citep{JMLR:v24:23-0045}, goodness of fit \citep{KWAK2021}, analysis of variance \citep{swanberg2019improved}, high-dimensional normal means \citep{narayanan2022private}, to name a few. Differentially private versions of common non-parametric tests \citep{10.1145/3319535.3339821}, permutation tests \citep{kim2023differentially}, etc., have emerged as well. Another recent line of work  proposes to automatically create private versions of the existing frequentist tests in a black-box fashion \citep{kazan2023test, peña2023differentially}, utilizing the subsample-and-aggregate method, with the aggregation done by the uniformly most powerful binomial test \citep{10.5555/3327144.3327334}.

In this article, we intend to complement the existing literature via providing a novel generalised framework for differentially private Bayesian testing. This is  crucial since  p-values obtained from frequentist tests are routinely criticised for it's lack of transparency,  inability to quantify evidence in support of the null hypothesis, and inability to measure the importance of a result. Readers may refer to the American Statistical Association  statement on statistical significance and p-values \cite{wasserstein2016asa} for further discussions on this. On the other hand, Bayes factors \citep{kass1995bayes, jeffreys1961theory, morey2011bayes, rouder2009bayesian, wagenmakers2010bayesian, johnson2023bayes}  can quantify the relative evidence in favor of competing hypotheses, effectively addressing the persistent criticisms of p-values. This automatically calls for  development of differentially private Bayes Factors to test the scientific hypotheses of interest with confidentiality guarantees.


Prior to presenting the proposed methodology, we provide a concise overview of pertinent concepts in  Bayesian hypothesis testing and differential privacy, laying the groundwork for subsequent discussions.

\subsection{Bayesian hypothesis testing} 
Suppose we intend to test the null hypothesis $\mbox{H}_0:\theta\in\Theta_0$ against  the alternative hypothesis $\mbox{H}_1:\theta\in\Theta_1$.
The Bayes Factor  \citep{kass1995bayes, jeffreys1961theory, morey2011bayes, rouder2009bayesian, wagenmakers2010bayesian, johnson2023bayes} in favour of the alternative hypothesis $\mbox{H}_1$ against the null hypothesis $\mbox{H}_0$ is defined as 
\begin{align}\label{def:BF}
\mbox{BF}_{10} = \frac{\mbox{\rm P}(\mbox{\rm H}_1\con x)}{\mbox{\rm P}(\mbox{\rm H}_0 \con x)}  =  \frac{m_1(x)}{m_0(x)} \times  \frac{\mbox{\rm P}(\mbox{\rm H}_1)}{\mbox{\rm P}(\mbox{\rm H}_0)},    
\end{align}
where $\mbox{\rm P}(\mbox{\rm H}_i \con x)$ is the posterior probability of $\mbox{\rm H}_i$ given the data $x$, $\mbox{\rm P}(\mbox{\rm H}_i)$ is the  prior probability assigned to $\mbox{\rm H}_i$, and $M_i(x)$ denotes the marginal probability  assigned to the data under $\mbox{\rm H}_i, \ i =0, 1$. The marginal densities are computed via 
$$
m_i(x) = \int_{\Theta_i} \pi(x \con \theta) \pi_i(\theta) d\theta,
$$
where $\pi(x \con \theta)$ is the data generative model given the parameter $\theta$, and $\pi_i(\theta)$ is the  prior density for $\theta$ under $\mbox{\rm H}_i, \ i = 0, 1$.  Bayes factor effectively addresses the persistent criticisms regarding lack of interpretability of p-values \citep{wasserstein2016asa} in reporting outcomes of hypothesis tests, since it directly quantifies the relative evidence in favor of the competing hypotheses $\mbox{H}_0$ and $\mbox{H}_1$. Noteworthy, the value of a Bayes factor depends on the choices of $\pi_i(\theta),\ i = 0, 1$, and it is generally difficult to justify  any single default choice. Moreover, the expression of $M_i(x),\ i =0, 1$ may not be available in closed form, and require computing  a potentially high-dimensional numerical integral. Several modifications to the existing Bayes factor methodology are proposed to improve the reporting of scientific findings. Two such modifications of traditional Bayes factors, that will be utilized to construct differentially private Bayesian tests, are Bayes factors constructed based on common test statistics (BFBOTS) \citep{Johnson2005} and Bayes factors reported as a function of effect size (BFFs) \citep{Johnson2023}. We briefly review both the ideas next.

\subsection{Bayes factor based on test statistics}
To circumnavigate the issues with traditional Bayes factors, \citet{Johnson2005} proposed to define Bayes factors based on common test statistics, and put priors directly on the non-centrality parameter of the test statistics to specify the various competing hypotheses. Bayes factors are first defined in terms of standard $z$, $t$, $\chi^2$, and $F$ test statistics. The distribution of these test statistics under the null hypothesis is known. Under the alternative hypotheses, their asymptotic distributions are common non-central distributions, characterized solely by scalar-valued non-centrality parameters. As a result, the specification of the prior density under the alternative hypothesis is simplified, and no prior density is required under the null hypothesis. For example, suppose we observe data $X_1,\ldots, X_n\sim \mbox{N}(\mu, \sigma^2)$, and want to test $\mbox{H}_0:\mu=0$ against $\mbox{H}_1:\mu\neq 0$. In this context,  under $\mbox{H}_0$, the $t$-statistic $t=\frac{\sqrt{n}\Bar{X}}{\hat{\sigma}}$ follows a student's $t_{n-1}(0)$ distribution with degrees of freedom $n-1$, where $\Bar{X}$ and $\hat{\sigma}$ are the mean and standard deviation of the observed data. Under $\mbox{H}_1$ the $t$-statistic $t=\frac{\sqrt{n}\Bar{X}}{\hat{\sigma}}$ follows a (non-central) $t_{n-1}(\delta)$ distribution with non centrality parameter $\delta\geq 0$ and degrees of freedom $n-1$. In other words, the non centrality parameter $\delta = 0$ under $\mbox{H}_0$ and $\delta > 0$ under $\mbox{H}_1$. So, we can  elicit priors directly on the non-centrality parameter $\delta$ under $\mbox{H}_1$, and conduct a Bayesian testing. Moreover, if we utilise non-local priors \citep{nonlocal_hyp} on $\delta$ under $\mbox{H}_1$ instead of local priors, not only the expression of the marginal likelihoods are often available in closed form, but also it ensures quick accumulation of evidence in favor of both true $\mbox{H}_0$ and true $\mbox{H}_1$. Data-driven approaches for the prior hyper-parameter tuning \citep{johnson2023bayes, datta2024} were subsequently suggested, adding to the objectivity of such Bayesian testing methodology. 

\subsection{Bayes factor functions(BFFs)}

In \cite{Johnson2023}, the authors defined Bayes factor functions (BFFs) as a mapping from standardized effect sizes of interest to corresponding Bayes factors, or more precisely, as a mapping from prior densities centered on the standardized effect sizes of interest to the corresponding Bayes factors. For a given value of the test statistic, a range of Bayes factors based on the test statistic value is calculated by varying the prior densities elicited on the non-centrality parameter that defines the alternative hypothesis. In other words, the families of prior densities used to define these Bayes factors are indexed by the standardized effect size. Through this mapping, BFFs make the relationship between Bayes factors and prior assumptions more transparent by providing a clear interpretation of Bayes factors as a function of prior densities used in their formulation. Additionally, another crucial feature of BFFs relevant to us is that the BFFs enable the accumulation of evidence across multiple studies examining the same phenomenon.

To elucidate the construction of a BFF, let us reconsider a two-sided $t$-test as previously discussed. Let $X_1, \dots, X_n$ be independent and identically distributed as $N(\mu, \sigma^2)$, where $\sigma^2$ is unknown, and define the test statistic as $t = \sqrt{n} \bar{x} / \sigma$. Under this set up, $t \sim t_{n-1}(\sqrt{n}\pi^*)$, where $\pi^* = \mu / \sigma$ represents the standardized effect, and $\lambda = \sqrt{n}\pi^*$ is the non-centrality parameter associated with the test statistic. Under the null hypothesis, we have $\mbox{H}_0: \lambda = 0$, and no prior specification on $\lambda$ is required.
Under the alternative hypothesis $\mbox{H}_1:  \lambda \neq 0$, we may specify a normal-moment prior density \citep{Johnson2010}, as follows:
\[
\mbox{H}_1: t\mid\lambda\sim t_{n-1}(\lambda),\quad \lambda\sim j(\lambda \mid \tau^2),
\]
where
$$j(\lambda \mid \tau^2) = \frac{ \lambda^2}{\sqrt{2\pi}\tau^3} \exp\left(-\frac{\lambda^2}{2\tau^2}\right), \qquad \lambda \in \mathcal{R},$$ and $\tau^2$ is a prior hyper-parameter.
The mode of the prior density $j(\lambda \mid \tau^2)$ occurs at $\pm \sqrt{2}\tau$.  For a given value of standardized effect size $\pi^*$, if we set $\tau^2 = n\pi^{*2}/2$, then the mode of the prior density equals $\sqrt{n}\pi^*$. Consequently, the mode of the prior density aligns with the non-centrality parameter $\lambda$ of the test statistic. Utilizing this relationship, Bayes Factor Functions (BFFs) are developed to associate standardized effect sizes with corresponding Bayes factors. Specifically, for a given standardized effect size, the Bayes factors are computed using alternative prior densities that are centered on non-centrality parameters aligned with the specified standardized effect size.
We next provide a succinct overview of relevant concepts from differential privacy literature.

\subsection{Differential privacy}
Differential privacy \citep{dwork2006differential, baraheem2022survey}  presents a principled framework for safeguarding sensitive information in confidential data utilized for statistical analyses. It achieves this by ensuring that the probability of any specific output of the statistical procedure is nearly the same, regardless of the data belonging any one specific individual in the database. In this context, databases that vary only with respect to the data corresponding to a single individual are termed \emph{neighboring} databases.

\begin{definition}[Differential Privacy, \citep{dwork2006differential}] \label{def:dp}
A randomized algorithm $f:\mathcal{D}\to\mathcal{S}$ is $\varepsilon$-differentially private if for all possible outcomes  $\mbox{\rm S} \subseteq \mathcal{S}$  and
for any two neighbouring databases $\mbox{\rm D}, \mbox{\rm D}^{\prime}\in\mathcal{D}$, we have 
$
\mbox{\rm P}[f(\mbox{\rm D}) \in \mbox{\rm S}] \leq e^\varepsilon \mbox{\rm P}[f(\mbox{\rm D}^{\prime}) \in \mbox{\rm S}].
$
\end{definition}

Further, the measure of how much the modification of a single row in a database can influence the output of a query is referred to as \emph{global sensitivity}.
\begin{definition}[Global sensitivity] \label{def:gs}
The global sensitivity of a function $f$  is 
$
\mbox{\rm GS}_f = \underset{\mbox{\rm D},\mbox{\rm D}^{\prime}\in\mathcal{D}}{ \textup{max}}\ |f(\mbox{\rm D})-f(\mbox{\rm D}^{\prime})|,
$
where $\mbox{\rm D}$ and $\mbox{\rm D}^{\prime}$ are neighboring databases. 
\end{definition}

The \emph{privacy budget} $\varepsilon$ quantifies the degree of privacy protection, with smaller values of $\varepsilon$ indicating a stronger privacy guarantee. In particular, $\varepsilon$-differential privacy ensures that the privacy loss is bounded by $\varepsilon$ almost surely, which implies that the ratio of the output probabilities under neighboring datasets is constrained by $e^{\varepsilon}$ for all outcomes. As $\varepsilon$ becomes smaller, the privacy of the mechanism $f$ increases, since the output distributions under neighboring datasets $f(\rm D)$ and $f(\rm D')$ are forced to be similar \citep{Barrientos2019}.
Every differentially private algorithm achieves this via incorporating an element of randomness. One of the most widely adopted such technique is the Laplace mechanism  \citep{dwork2006differential}, that involves introducing a privacy noise sampled from Laplace distribution to the result of the query that requires privacy protection.  The scale of the  Laplace distribution employed to introduce the privacy noise, is contingent on both the privacy budget $\varepsilon$ and the global sensitivity $\mbox{\rm GS}_f$. Given any function $f$, the \emph{Laplace mechanism} is defined as
$$
\tilde{f}(\mbox{\rm D})= f(\mbox{\rm D}) + \eta, $$
where $\eta$ is a privacy noise  drawn from $\mbox{Laplace}(0, \mbox{\rm GS}_f/\varepsilon)$, and $\mbox{\rm GS}_f$ is the global sensitivity of $f$.  We now have reviewed all the necessary tools to introduce our methodology.
\subsection{Relevant literature}
There has been growing interest in developing Bayesian inferential methods under the differential privacy (DP) framework. A notable contribution in this area is by \cite{peña2023differentially}, who proposed a differentially private hypothesis test based on a subsample-and-aggregate scheme. While their method primarily aligns with frequentist principles—focusing on assessing null hypothesis fit—they also extend it to the Bayesian paradigm by computing posterior odds based on the distribution of the test statistics they introduce. However, due to the nature of their test statistic, the resulting Bayes factor lacks consistency under the alternative hypothesis. Additionally, the method relies on a subjective prior for the test's power, rendering the Bayes factor sensitive to this prior specification.
To address these limitations, \cite{Pena2024} introduced an alternative differentially private test that uses truncated likelihood ratio statistics or truncated Bayes factor, effectively bounding the global sensitivity of the Bayes factor. While this approach improves privacy guarantees, it raises interpretability concerns, as the truncation is not derived from the underlying generative model.

In the context of linear regression, \cite{Amitai_Reiter_2018} proposed two approaches for Bayesian inference based on posterior probabilities of regression coefficients. The first approach assumes a normal approximation to the posterior and requires unbiased posterior estimates of the coefficients, making it suitable primarily for large sample sizes. The second approach aggregates posterior probabilities using Fisher’s method and assumes that the resulting statistic follows a $\chi^2$ distribution. This assumption fails under the alternative hypothesis. Moreover, their method involves clipping posterior probabilities at a fixed threshold, which may suppress evidence against the null. Importantly, their work does not focus specifically on Bayesian hypothesis testing.

Other notable contributions include \cite{Bernstein2019}, who proposed a differentially private method for posterior sampling of regression parameters, although their work does not directly address hypothesis testing. In parallel, \cite{Dimitrakakis2017, Heikkila2019, Hu2023} developed various techniques for drawing posterior samples under differential privacy constraints. These methods typically rely on the assumption of a bounded likelihood and require a privacy budget that scales with the number of posterior samples. Consequently, when the privacy budget is small, such approaches may yield unreliable Monte Carlo approximations.

Drawing inspiration from these works, we propose an objective Bayesian testing procedure that is consistent under the true hypothesis.

\subsection{Our contributions}
The key contributions in the article are three-fold. 
First, to the best of our knowledge, we introduce the first-ever differentially private objective Bayesian testing framework and ensure the consistency of the Bayes factor under the true model. This development is important in light of the American Statistical Association's statement on statistical significance and p-values \citep{wasserstein2016asa}, which warns against making scientific decisions based solely on p-values, due  to their lack of transparency, inability to quantify evidence supporting the null hypothesis, and failure to measure the practical significance of a result. While Bayes factor-based tests offer a solution to many of the limitations associated with p-values, extending these tests to a differentially private setting—while preserving their inherent interpretability—presents substantial challenges. To address this, we integrate widely-used strategies in differentially private testing, such as sub-sample and aggregate methods and truncated test functions, within the proposed data-generative mechanism. This careful integration ensures that the resulting Bayesian testing procedures emerge naturally from the data generation models, maintaining the interpretability of the resulting inferences.

Second, we introduced differentially private Bayes factors that rely on widely-used test statistics, extending the work of \citet{Johnson2005} to a privacy-preserving framework. In this approach, we place priors directly on the non-centrality parameter of the test statistic and employ data-driven techniques for tuning prior hyperparameters. This strategy eliminates the need to model the entire data-generating process and avoids placing priors on potentially high-dimensional model parameters. As a result, the method offers significant computational advantages when calculating marginal likelihoods, which typically involve intractable integrals in fully parametric Bayesian testing frameworks. Furthermore, we establish a set of sufficient conditions to ensure the consistency of the Bayes factor \citep{chib2016, Chatterjee2020} within the proposed framework.

Finally, we critically leverage the proposed Bayes factor  based on test statistics framework, to offer a general template for constructing size-\(\alpha\) differentially private Bayesian tests that adhere to a predefined privacy budget, provided the investigator has pre-specified the effect size of interest. Additionally, we propose numerical schemes for data-driven hyperparameter tuning. Specific examples under particular hypothesis testing scenarios are presented. Nevertheless, the applicability of the proposed schemes extends beyond the Bayes factors based on test statistics discussed in this article.

In a nutshell, the key differences between our contribution and the existing works of \cite{peña2023differentially, Pena2024} and \cite{Amitai_Reiter_2018} are as follows:
\begin{enumerate}
\item  We introduce Bayes factors for hypothesis testing within an objective Bayesian framework, which eliminates the dependence on subjective prior elicitation. This ensures that the resulting Bayes factors are interpretable  and mitigates the sensitivity to prior specifications that is present in the approaches of \cite{peña2023differentially}.
\item Unlike \cite{Pena2024}, our method does not require any direct truncation scheme to guarantee the boundedness of the General Sensitivity Parameter (GSS). This property is inherently satisfied by our model specification and prior choice. As a result, we retain full interpretability of the Bayes factor, which may be compromised in \cite{Pena2024} due to the non-generative truncation required to bound the GSS. Consequently, the Bayes factor in their work can be hard to interpret.
\item 
We evaluate the Bayes factor over a sequence of alternative hypotheses indexed by the effect size of interest, which introduces additional flexibility.
\item As outlined in our general framework, presented in Section~\ref{framework}, the proposed method extends well beyond test statistic-based Bayes factors. We emphasize Bayes factors based on test statistics primarily to avoid subjective priors on nuisance parameters. We use sufficient statistics to define Bayes factors, thereby ensuring there is no potential information loss. However, our general framework can encompass a broader class of models.
\item In contrast to \cite{Amitai_Reiter_2018}, our primary objective is hypothesis testing for the purpose of model selection. We aim to achieve both interpretability and consistency in model selection, both under the null and alternative hypotheses.
\end{enumerate}

Rest of the article is organized as follows. Section \ref{framework} introduces the proposed framework for differentially private Bayesian testing in complete generality, and presents the key features of the methodology. In Section \ref{ssec:method_teststat}, we introduce differentially private Bayesian tests based on common test statistics, discuss their asymptotic properties under various common hypothesis testing problems and discuss schemes for the hyper-parameter tuning under the proposed technology. In Section \ref{experiments_t}, we study the numerical efficacy of the proposed technology in different hypothesis testing problems with practical utility. Finally, we conclude with a discussion.

\section{General framework}\label{framework}

In this section, the objective is to lay down a hierarchical model equipped with carefully chosen priors on the parameter of interest to specify competing hypotheses, which naturally gives rise to the proposed differentially private Bayesian tests. Notably, this attribute remains elusive in the existing literature and that hinders the interpretability of the resulting inference. In particular,  many existing differentially private frequentist  testing procedures \citep{peña2023differentially, kazan2023test} rely on an ad-hoc sampling and aggregation scheme coupled with a test statistic truncation step, to formulate the desired  tests. Despite the empirical success of such methods, a significant deficiency stems from the lack of a principled probabilistic interpretation, i.e, such methods does not enable us to quantify evidence in favor of the null and alternative hypotheses. We specifically tackle this issue in the subsequent discussions. 

\subsection{Hierarchical model and prior specification}
\textcolor{blue}{For a positive integer $t$, denote $[t] :\,= \{1, \ldots, t\}$. Suppose we observe data $\mathbf{x}=(x_1,\ldots, x_n)'$ from a probability distribution having density $\pi(\cdot\mid\theta),\ \theta\in\Theta$. We then randomly divide the data into $M_n$ partitions $\mathbf{x}^{(i)}, i\in[M_n]$.
We can assume that the data in each of the $M_n$ partitions arise from
\begin{align}\label{eqn:hm1}
    \mathbf{x}^{(i)}\sim \prod_{j=1}^{n_{i}} \pi(\mathbf{x}^{(i)}_j\mid\theta^{(i)}), \quad\text{independently for}\ i\in[M_n], 
\end{align}
 such that $\sum_{i=1}^{M_n} n_i = n$. The above data generating scheme comes in handy to formalize the \emph{sample and aggregate} scheme under the proposed setup.} 
 \textcolor{blue}{
In privacy-preserving inference, this subsample--and--aggregate structure plays a crucial role in controlling the global sensitivity of the resulting quantity of interest, e.g, test statistic, Bayes factor. By splitting the dataset into multiple partitions, computing intermediate quantities within each subset, and then aggregating these results, one can approximate the full-sample quantity of interest. This construction provides the foundation for the proposed differentially private Bayes factor, as it allows the privacy noise to scale with the sensitivity of the partition-level statistics rather than the entire dataset.}


\textcolor{blue}{Next, we posit the priors on the parameter of interest $\theta$ under the competing hypotheses. To that end, we assume that
\begin{align}\label{eqn:hm2}
    &\theta^{(i)}\mid\tau_{0,i}, \tau_{1,i}\sim (1-\omega_n)\ \pi_{0}(\theta^{(i)} \mid \tau_{0,i}^2)\ +\ \omega_n\ \pi_{1}(\theta^{(i)} \mid \tau_{1,i}^2) \quad \text{under}\ \mbox{H}_0,\notag\\
    & \theta^{(i)}\mid\tau_{0,i}, \tau_{1,i}\sim\ \omega_n\ \pi_{0}(\theta^{(i)}\mid \tau_{0,i}^2)\ +\ (1 - \omega_n)\ \pi_{1}(\theta^{(i)} \mid \tau_{1,i}^2)\quad \text{under}\ \mbox{H}_1,
\end{align}
independently for $i\in[M_n]$, where  $\{\pi_{i}(\cdot),\ i = 0, 1\}$ are the prior probability distributions, $(\tau_{0,i}, \tau_{1,i})$ are prior hyper-parameters, and $\omega_n\in (0, \frac{1}{2})$ is a hyper-parameter that ensures that the Bayes factor for testing $\mbox{H}_0$ against $\mbox{H}_1$ is bounded. We will later on show that $\omega_n$ depends on the sample size $n$ and hence find it necessary to index it with $n$. If the Bayes factor for testing $\mbox{H}_0$ against $\mbox{H}_1$ were to be unbounded, the global sensitivity parameter of the Bayes factor becomes unbounded too. An immediate fix against it would involve truncating the log Bayes factor beyond an interval of the form $[-a_n, a_n], a_n>0$ or more generally $[L_n,U_n]$ with $L_n<U_n$, akin to the approach in \cite{Pena2024}. 
If one were to adopt the framework of \cite{Pena2024}, impose non-local priors on the parameters of interest, and implement an objective Bayesian hyper-parameter tuning strategy similar to that used in the present work, one would obtain a Bayes factor with operating characteristics comparable to the differentially private Bayes factor developed here. However, such a construction would lack the interpretability afforded by our formulation, which arises from a hierarchical model with carefully specified priors that directly encode the competing hypotheses and thereby yield the proposed differentially private Bayesian tests in a principled manner.
Upon closer look, truncating the log Bayes factor in principle means that the maximum evidence that we can obtain in favor of $\mbox{H}_0$ (or $\mbox{H}_1$) against $\mbox{H}_1$ (or $\mbox{H}_0$) is bounded from above -- we precisely impose this constraint a-priori via the hierarchical specification \eqref{eqn:hm2}, thereby eliminating the need for truncation/post-processing. 
For each partition $\mathbf{x}^{(i)}, \ i\in[M_n]$,  we record the notations  for the integrals
    $$m_k(\mathbf{x}^{(i)}\mid \tau_{k,i}^2) = \int_\Theta \bigg[\prod_{j=1}^{n_j}\pi(\mathbf{x}^{(i)}_j\mid\theta^{(i)})\bigg]\ \pi_{k}(\theta^{(i)}\mid \tau_{k,i}^2)\ d\theta^{(i)}, \quad k = 0, 1.$$
Given \(\omega_n, \tau_{0,i}, \tau_{1,i}\), and assuming independence, the marginal \(m_k(\mathbf{x} \mid \tau_{k,i}^2)\) is the product of the marginals across each partition, \(m_k(\mathbf{x}^{(i)} \mid \tau_{k,i}^2, \omega_n)\). Consequently, since the Bayes factor is the ratio of these marginal densities with equal prior probabilities on the hypotheses, it becomes the product of the individual Bayes factors for each partition. Aggregation of evidence across partition is further detailed in Section \ref{sec:key_prop}. In summary, the mixture priors on $\{\theta^{(i)},\ i\in[M_n]\}$ under the competing hypotheses are instrumental in formalizing the \emph{truncation} scheme necessary to ensure that the global sensitivity parameter of the privacy mechanism is bounded, while preserving the interpretability of the resulting Bayes factors. One may criticize the appearance of the parameter $\omega_n$ in our formulation. To that end,  we carefully introduce an asymptotic regime to analyze the proposed methodology, so that the effect of $\omega_n$ washes away as the sample size diverges to $\infty$. }

The model and prior specified in equations \eqref{eqn:hm1}-\eqref{eqn:hm2}, together with the assumption that 
\begin{align}\label{eqn:hm3}
    \mbox{P}(\mbox{H}_0) = \mbox{P}(\mbox{H}_1) = 0.5,
\end{align}
apriori, completely describe our generative model of interest.



\subsection{Properties}\label{sec:key_prop}
\textcolor{blue}{We note that for each partition $\mathbf{x}^{(i)}, \ i\in[M_n]$,  the quantity
    $m_k(\mathbf{x}^{(i)}\mid \tau_{k,i}^2),\ k = 0, 1$,
can be interpreted as the marginal likelihood of the data partition  $\mathbf{x}^{(i)}$ under the hypothesis $\mbox{H}_k$ when $\omega_n =0$. Under this set up, as we eluded to earlier, the Bayes factor for testing $\mbox{H}_0$ against $\mbox{H}_1$ can be unbounded, and consequently the global sensitivity parameter of the privacy mechanism is unbounded too. This renders the development of differentially private Bayesian tests non-trivial, and throughout the article we shall assume that the hyper-parameter $\omega_n> 0$.} Under the alternative hypothesis $\mbox{H}_1$   in \eqref{eqn:hm1}-\eqref{eqn:hm3}, the marginal distribution of the data partition  $\mathbf{x}^{(i)}$ is expressed as
\[
m^t_1(\mathbf{x}^{(i)} \mid \tau_{0,i}^2,\tau_{1,i}^2 ,\omega_n) =  \omega_n\ m_0(\mathbf{x}^{(i)} \mid \tau_{0,i}^2)\ + \ (1-\omega_n)\ m_1(\mathbf{x}^{(i)} \mid \tau_{1,i}^2), \quad i\in[M_n];
\]
and under the null hypothesis $\mbox{H}_0$, it is given by:
\[
m^t_0(\mathbf{x}^{(i)}\mid \tau_{0,i}^2,\tau_{1,i}^2 ,\omega_n) = (1-\omega_n)\ m_0(\mathbf{x}^{(i)}\mid \tau_{0,i}^2)\ +\ \omega_n\ m_1(\mathbf{x}^{(i)}\mid \tau_{1,i}^2),  \quad i\in[M_n].
\]
Consequently, in light of the data partition  $\mathbf{x}^{(i)}$,  the Bayes factor against the null hypothesis $\mbox{H}_0$ is expressed as
\begin{align}\label{eqn:BF}
    &\mbox{BF}^t_{10}(\mathbf{x}^{(i)} \mid \tau_{0,i}^2,\tau_{1,i}^2 ,\omega_n) \notag\\
    =&\frac{\omega_n  + (1-\omega_n)\big[m_1(\mathbf{x}^{(i)}\mid\tau_{1,i}^2)\ /\ m_0(\mathbf{x}^{(i)}\mid\tau_{0,i}^2)\big]}{(1-\omega_n)  + \omega_n \big[m_1(\mathbf{x}^{(i)}\mid\tau_{1,i}^2)\ /\ m_0(\mathbf{x}^{(i)}\mid\tau_{0,i}^2)\big] }, \quad i\in[M_n].
\end{align}
We proceed by listing out the key properties of the Bayes factor against the null hypothesis  $\mbox{BF}^t_{10}(\mathbf{x}^{(i)} \mid \tau_{0,i}^2,\tau_{1,i}^2 ,\omega_n)$ for the data partition $\mathbf{x}^{(i)}$.

\begin{lemma}\label{prop1}
    For $\omega_n\in(0, 1/2)$, $\mbox{\rm BF}^t_{10}(\mathbf{x}^{(i)} \mid \tau_{0,i}^2,\tau_{1,i}^2 ,\omega_n)$ is bounded between $[\frac{\omega_n}{1-\omega_n}, \frac{1-\omega_n}{\omega_n}]$.
\end{lemma}


\begin{lemma}\label{prop2}
Under \eqref{eqn:hm1}-\eqref{eqn:hm3}, the  Bayes Factor against the null hypothesis $\mbox{H}_0$ combined across all the partitions $\mathbf{x}^{(i)},\ i\in[M_n]$ is expressed as
    \begin{align*}
        \mbox{\rm BF}^t_{10}(\mathbf{x} \mid \tau_{0,i}^2,\tau_{1,i}^2 ,\omega_n) =\prod_{i=1}^{M_n} \mbox{\rm BF}^t_{10}(\mathbf{x}^{(i)} \mid \tau_{0,i}^2,\tau_{1,i}^2 ,\omega_n).
    \end{align*}
\end{lemma}

\begin{corollary}
 Lemma \ref{prop1} together with Lemma \ref{prop2} implies that, the combined Bayes Factor against the null hypothesis $\mbox{H}_0$ across all the partitions $\mathbf{x}^{(i)},\ i\in[M_n]$, $\mbox{\rm BF}^t_{10}(\mathbf{x}\mid \tau_{0,i}^2,\tau_{1,i}^2 ,\omega_n)$ is bounded between $\left(\frac{\omega_n}{1-\omega_n}\right)^{M_n}$ and $ \left(\frac{1-\omega_n}{\omega_n}\right)^{M_n}$ for $\omega_n\in(0, 1/2)$.   
\end{corollary}

In the sequel, we construct differential private Bayesian tests based on the randomised mechanism, denoted by $f$, that map a data set $\mathbf{x}$ to the average log Bayes factor 
\begin{align}\label{eqn:avg_BF}
   f_{10}(\mathbf{x}) = \frac{1}{M_n}\log\mbox{BF}^t_{10}(\mathbf{x}\mid \tau_{0,i}^2,\tau_{1,i}^2 ,\omega_n) = \frac{1}{M_n}\sum_{i=1}^{M_n} \log\mbox{BF}^t_{10}(\mathbf{x}^{(i)} \mid \tau_{0,i}^2,\tau_{1,i}^2 ,\omega_n).
\end{align}
We take refuge to common differential privacy mechanism of adding  calibrated privacy noise to the randomised mechanism $f$ to ensure differential privacy of the inferential procedure. For every data set $\mathbf{x}$, since the log Bayes factor in each of the partitions $\log\mbox{BF}^t_{10}(\mathbf{x}^{(i)} \mid \tau_{0,i}^2,\tau_{1,i}^2 ,\omega_n)$ is bounded between $[-a_n, a_n]$ with 
$
a_n =- \log(\omega_n/1-\omega_n),
$ 
the global sensitivity  of $\log\mbox{BF}^t_{10}(\mathbf{x}^{(i)} \mid \tau_{0,i}^2,\tau_{1,i}^2 ,\omega_n)$ is $2a_n$. Consequently, the global sensitivity  of $\mbox{GS}_f$ of the log Bayes factor averaged over the partitions, i.e of $f$, is  $(2a_n/M_n)$. We now have all the necessary ingredients to introduce our privacy preserving mechanism.

To that end, let $\mathcal{D}$ denote the collection of all possible data sets. We consider a random mechanism $\mbox{H}: \mathcal{D} \rightarrow \mathbf{R}$ expressed as
\begin{equation}\label{eqn:avg_privateBF_general}
    \mbox{H}_{10}(\mathbf{x}) = f_{10}(\mathbf{x}) + \eta, \quad \mathbf{x}\in \mathcal{D},
\end{equation}
where $\eta$ is  a $\mbox{Laplace}(0, \mbox{GS}_f/\varepsilon)$ distributed privacy noise. A test based on $\mbox{H}_{10}(\mathbf{x})$ is then obtained as
\begin{equation}\label{eqn:avg_privateBFtest_general}
\phi(\mathbf{x}) =
\begin{cases}
1 & \text{if}\ \mbox{H}_{10}(\mathbf{x})\geq \gamma_{\alpha, \varepsilon}(n),\\
0 &\text{otherwise},
\end{cases}
\end{equation}
for some cut-off $\gamma_{\alpha, \varepsilon}(n)$, depending on the sample size $n$ and size $\alpha$ of the test. Here, $\phi(\mathbf{x})=1$ signifies rejection of $\mbox{H}_0$/ acceptance of $\mbox{H}_1$, and if $\phi(\mathbf{s})=0$, accept $\mbox{H}_0$.
\begin{theorem}
    The function $\phi(\mathbf{x})$ for testing $\mbox{H}_0$ against $\mbox{H}_1$, as described in \eqref{eqn:avg_BF}-\eqref{eqn:avg_privateBFtest_general}, is $\varepsilon$-differentially private.
\end{theorem}

\begin{proof}
    The global sensitivity $\mbox{GS}_f$ of $f(\mathbf{x})$ is $\frac{2a_n}{M_n}$. Then, defining $\mbox{H}(\mathbf{x}) = f(\mathbf{x}) + \eta$, where $\eta$ follows  $\mbox{Laplace}(0, \mbox{GS}_f/\varepsilon)$ completes the proof, by definition.
\end{proof}

Next, we develop differentially private Bayes factors based on widely-used test statistics, and  crucially leverage the proposed framework to provide a systematic approach for selecting \(\gamma_{\alpha, \varepsilon}(n)\) that maintains a predetermined privacy budget, assuming the investigator specifies the effect size of interest in advance.

\section{Differentially private Bayes factors based on test statistics}\label{ssec:method_teststat}

The general framework presented in Section \ref{framework} can be employed to devise fully parametric Bayesian tests that preserves differential privacy. However, given the limitations of Bayes factors derived from fully parametric models, as outlined in the introduction, and in order to ensure straightforward preservation of \((\varepsilon, 0)\)-differential privacy in the proposed mechanism, we instead develop differentially private Bayes factors based on test statistics. 
 On top of computational simplicity and objectivity, a major benefit of representing Bayes factors in terms of test statistics that is crucial for us   is that they can be formulated as a function of the standardized effect size \citep{Johnson2023, datta2024}, if we specify the competing hypotheses in \eqref{eqn:hm2} using a mixture of a spike at $0$ and an appropriately chosen first order non-local prior \citep{Johnson2010, Johnson2023} on the non-centrality parameter of the appropriate test statistic. Moreover, such non-local priors contain a single scale parameter that control the mode of the prior densities that in turn can be utilized to define the null and alternative hypotheses. By determining the prior mode using a function of the standardized effect size (analogous to BFFs.), we can derive Bayes factors for a sequence of null and alternative hypothesis.  In particular, the hyper parameter $\tau^2_{1,i}$ in \eqref{eqn:hm2} can be objectively determined based on the standardized effect size of interest to express our Bayes factors as functions of the sequence of null and alternatives. This enables us to determine a privatized cut-off $\gamma_{\alpha, \varepsilon}(n)$ that varies for each sequence of null and  alternatives, while adhering to pre-specified privacy budget.

Beyond aiding the computation of the cut-off value $\gamma_{\alpha, \varepsilon}(n)$, such that $(\epsilon, 0)$-differential privacy is ensured, the choice of priors throughout the rest of the article is further guided by two other considerations. First, the carefully chosen non-local priors ensure that both the marginal likelihoods under the two competing hypotheses of interest are available in closed form. This eliminates the computational expenses associated with the numerical evaluation of the otherwise intractable multiple integrals. Secondly, and perhaps more important, the prior specification ensures rapid accumulation of evidence in favor of the true hypothesis.

We shall now formally introduce the differentially private Bayes factor based on test statistics. We note that,  the partition specific log Bayes factor based on test statistic $\log\mbox{BF}^t_{10}(\mathbf{x}^{(i)} \mid \tau_{0,i}^2,\tau_{1,i}^2 ,\omega_n)$, as shown later in Section \ref{sec:BFBOTS_DP},  depends on  $\mathbf{x}^{(i)}$ only through the partition specific test statistic, denoted by $s^{(i)},\ i\in[M_n]$. To make it explicit, we use the notation $\log\mbox{BF}^t_{\rm stat, 10}(s^{(i)} \mid \tau_{0,i}^2,\tau_{1,i}^2 ,\omega_n)$ instead of $\log\mbox{BF}^t_{10}(\mathbf{x}^{(i)} \mid \tau_{0,i}^2,\tau_{1,i}^2 ,\omega_n)$ in the rest of the section. Similarly, we shall use $f_{\rm stat, 10}(\mathbf{s})$ and $\mathrm{H}_{\rm stat,10}(\mathbf{s})$ to denote the non-privatized and privatized average log Bayes factors, instead of $f_{10}(\mathbf{x})$ and $\mbox{H}_{10}(\mathbf{x})$, respectively, where $\mathbf{s} = (s^{(1)},\ldots, s^{(M_n)})^{\T}$. With that, the Bayesian test function based on test statistic $\mathrm{H}_{\rm stat,10}(\mathbf{s})$ is  expressed as
\begin{equation}\label{eqn:avg_privateBF}
    \mathrm{H}_{\rm stat,10}(\mathbf{s}) = f_{\rm stat, 10}(\mathbf{s}) + \eta, \quad \mathbf{s}\in \mathcal{D},
\end{equation}
where $\eta$ is  a $\mbox{Laplace}(0, \mbox{GS}_f/\varepsilon)$ distributed privacy noise. Similar to the general framework, a test based on $ \mathrm{H}_{\rm stat,10}(\mathbf{s})$ is then obtained as
\begin{equation}\label{eqn:avg_privateBFtest}
\phi(\mathbf{s}) =
\begin{cases}
1 & \text{if}\ \mathrm{H}_{\rm stat,10}(\mathbf{s})\geq \gamma_{\alpha, \varepsilon}(n),\\
0 &\text{otherwise},
\end{cases}
\end{equation}
for some cut-off $\gamma_{\alpha, \varepsilon}(n)$, depending on the sample size $n$ and size $\alpha$ of the test. Here, $\phi(\mathbf{s})=1$ signifies rejection of $\mbox{H}_0$/ acceptance of $\mbox{H}_1$, and if $\phi(\mathbf{s})=0$, accept $\mbox{H}_0$.

Next, given the sample size $n$, privacy mechanism, the  user defined parameters $(\varepsilon, a_n, M_n)$,  and the size  $\alpha$ of the test, we determine the cut off $\gamma_{\alpha, \varepsilon}(n)$ by satisfying the \emph{size constraint}
\begin{align}\label{eqn:BF_cutoff}
   \Xi_{0}(M_n\mid \varepsilon, \alpha, n) = \mbox{E}\big[\phi(\mathbf{s})\mid \mbox{\rm H}_0] = \mbox{P}_{\mbox{\rm H}_0}\big[\mbox{\rm H}(\mathbf{s}) \geq \gamma_{\alpha, \varepsilon}(n)\big] = \alpha.
\end{align}
Exact calculation of the cut off $\gamma_{\alpha, \varepsilon}(n)$ may not be amenable under certain cases. However, we can always obtain $\gamma_{\alpha, \varepsilon}(n)$ via a Monte Carlo simulation.  We present the formalized algorithm to determine the cutoff value $\gamma_{\alpha, \varepsilon}(n)$ in Algorithm \ref{algorithm1}. Determination of the cut off is further demonstrated in the experiment section. 

\subsection{Selection of size $\alpha$ cut-off $\gamma_{\alpha, \varepsilon}(n)$}

Given the privatized average logarithm of Bayes factor based on test statistic $\mathrm{H}_{\rm stat, 10}$ as defined in \eqref{eqn:avg_privateBFtest}, our goal in Algorithm \ref{algorithm1} is to  compute the associated size $\alpha$ cut-off $\gamma_{\alpha, \varepsilon}$ via a Monte Carlo scheme. Such Monte Carlo approaches are ubiquitous in literature  in finding cut-offs of test functions; see \cite{Barrientos2019}  for an example in context of differentially private frequentist tests.

\textcolor{blue}{The inputs of the Algorithm \ref{algorithm1} are (a) the privacy budget $\varepsilon$, (b) the number of partitions \( M_n \), and truncation level of the partition specific log Bayes factors \( a_n \),  (c) the standardised effect size of interest $\pi^*$, and finally (d) the number of iterations in the Monte-Carlo simulation \( N \). To find the cut-off  $\gamma_{\alpha, \varepsilon}(n)$,  we note that the partition specific common test statistic $s^{(i)}$  follows a non-central distribution, denoted by $\mbox{S}_{\nu_i}(\lambda^{(i)})$,  under the null hypothesis in \eqref{eqn:hm1}-\eqref{eqn:hm2}, where 
$$\lambda^{(i)}\sim (1-\omega_n)\delta_0 + \omega_n \pi_{1}(\lambda^{(i)} \mid \tau_{i}^2)$$ 
and $\nu_i$ is a constant completely determined by the size of the $i$-th partition of the data. 
With this insight, at the iteration $k\in\{1,\ldots, N\}$ of the Monte Carlo scheme,   we first independently draw the non-centrality parameter $\lambda^{(i)}$ from a prior distribution with density $(1-\omega_n)\delta_0 + \omega_n \pi_{1}(\lambda^{(i)} \mid \tau_{i}^2)$ and  pseudo test statistics \( s_k^{(i)} \)  from a non-central distribution  $\mbox{S}_{\nu_i}(\lambda^{(i)}),\ i\in[M_n]$. Crucially, the prior hyper-parameter $\tau_{i}^2$ is completely determined by the effect size of interest $\pi^\star$ and size of the $i$-th partition $n_i$.
For example, for a $t$-test, we draw the pseudo test statistic $s_k^{(i)}\equiv t_k^{(i)}$ from  non-central  $t_{|n_i| - 1}(\lambda^{(i)})$ distributions with non-centrality parameter $\lambda^{(i)}$, where 
$$\lambda^{(i)} \sim (1-\omega_n) \delta_0 + \omega_n J(\tau_i^2),\quad \tau_i = \sqrt{n_i} \pi^\star,$$ 
where $J(\tau_i^2)$ denotes a normal moment density of order 1, the form of which is given by,
\begin{equation}
J( x \mid \tau^2 ) = \frac{x^2}{\sqrt{2 \pi} \tau^3  } \exp\left( -\frac{x^2}{2\tau^2} \right),\quad x\in\mathbf{R}.
\end{equation}
The normal moment prior \cite{Johnson2010, Johnson2023} has two key features that make it particularly well-suited for Bayesian hypothesis testing. First, it is a nonlocal prior, meaning it assigns zero probability density at the null value (typically zero). This property enforces a clear separation between the null and alternative hypotheses by penalizing parameter values close to the null, thereby reducing false positives and increasing the evidence in favor of the null when it is true. Second, the normal moment prior improves model selection consistency, especially in high-dimensional or sparse settings. Its structure ensures that, as sample size increases, the Bayes factor reliably favors the true model—whether it corresponds to the null or the alternative—leading to more robust and interpretable inference.}

For a $\chi^2$-test, we draw the test statistic $s_k^{(i)}\equiv h^{(i)}$ from a non-central  $\chi^2_\nu(\lambda^{(i)})$, where 

$$\lambda^{(i)} \sim (1-\omega_n)\ \delta_0 + \omega_n\ \ \mbox{\rm G}((k/2) + 1,\ 1/2\tau_i^2),\quad  \tau_i^2 = \frac{n_i \pi^{\star'}\pi^\star}{2},$$ 
where $\mathrm{G}(a, b)$ denotes a gamma density with shape parameter $a_n$ and rate parameter $b$.
For an $F$-test,  we draw the test statistic $f^{(i)}$ from a non-central $F_{p, n_i - p}(\lambda^{(i)})$ for a $p\geq 1$. Then, based on $\mathbf{s}_k = ( s_k^{(1)}, \ldots, s_k^{(M_n)})^{\T}$, we calculate the pseudo non-private truncated log Bayes factor  \( \mathrm{f}_{\rm stat, 10}(\mathbf{s}_k) \) via Equation \eqref{eqn:avg_BF}. Then, a  privacy noise $\eta$ generated from  \( \mbox{Laplace}(0, 2a_n/\varepsilon M_n) \) is added to \( \mathrm{f}_{\rm stat, 10}(\mathbf{s}_k) \) to calculate the pseudo private truncated log Bayes factor $\mbox{H}_{\rm stat, 10}(\mathbf{s}_k)$ via Equation \eqref{eqn:avg_privateBF}. 
 In step (3), we set the size $\alpha$ cut-off  \( \gamma_{\alpha,\varepsilon}(n) \) to the $100 (1-\alpha) \%$ quantile the $\{\mbox{H}_{\rm stat, 10}(\mathbf{s}_k),\ k\in[N]\}$ values. It is important to note that we are sampling the test statistics from their conditional distributions, \( S_\nu(\lambda) \), under the null hypothesis, where the mode of the slab part of the prior is set by the standardized effect size \( \pi^\star \). In this process, we are not sampling pseudo data, but solely the test statistics which only depend on the partition specific sample size and the standardized effect size of interest. This ensures the preservation of differential privacy of the proposed mechanism. 

\textcolor{blue}{
We note that for a broad class of models, a sufficient test statistic with a tractable null distribution may not exist, even though the full data-generating distribution under both the null and alternative hypotheses is known. In such settings, the general framework described in Section~\ref{framework} can be applied directly to the full dataset, without reduction to a test statistic. An explicit algorithm for computing the corresponding Bayes factor cutoff in this full-data setting is provided in Section~2 of the supplementary material.
}

\begin{algorithm}
\caption{(Bayes factor cut-off)}\label{algorithm1}
\begin{algorithmic}[1]
\State \textbf{Input.} (i) The number of partitions $M_n$, (ii) the  truncation level $a_n$ of the partition specific log Bayes factors, (iii) privacy parameter $\varepsilon$, (iv) \( N \): The total number of the Monte-Carlo simulations $N$ to estimate the size $\alpha$ cut-off $\gamma_{\alpha,\varepsilon}(n)$, (v) Standardized effect size of interest  $\pi^\star \in (0, 1)$.
    \For{Standardized effect size of interest $\pi^\star$}
\For{\( k = 1 \) to \( N \)}

(i) Compute of $\tau^2_i, i = 1, 2,\ldots, M_n$ in terms of $\pi^\star$ for each of the $M_n$ partitions of the data. For example, for a $z$ or $t$ test $\tau^2_i = \frac{n_i \pi^{\star\prime}\pi^{\star}}{2} $ for the $i$-th partition. 
\footnotemark

 (ii) Draw independent samples of the non-centrality parameter, 
    \[
    \lambda^{(i)} \sim (1-\omega_n)\delta_0 + \omega_n \pi_{1}(\lambda^{(i)} \mid \tau_{i}^2),
    \]
    where $\pi_{1}(\lambda^{(i)} \mid \tau_{i}^2)$ is a non-local moment prior distribution, $i \in [M_n]$. The prior hyper-parameter $\tau_{i}^2$ is completely determined by  standardized effect size of interest  $\pi^\star$ and the size of the $i$-th partition $n_i$.

   (ii) Draw test statistics $s_k^{(i)}$ from $\mbox{S}_{\nu_i}(\lambda^{(i)})$ independently for $i \in [M_n]$.

 (iii) Compute $\mbox{\rm BF}^t_{\rm stat, 10}(s_k^{(i)} \mid \tau_i^2, \omega_n)$ for $i \in [M_n]$.

 (iv) Compute 
    \[
    f_{\rm stat, 10}(\mathbf{s}_k) = \frac{1}{M_n} \sum_{i=1}^{M_n} \log \mbox{\rm BF}^t_{\rm stat, 10}(s_k^{(i)} \mid \tau_i^2, \omega_n).
    \]

   (v) Generate \( \eta \) from Laplace \( (0, \frac{2a_n}{\varepsilon M_n}) \).

   (vi) Compute \( \mbox{H}_{\rm stat, 10}(\mathbf{s}_k) = f_{\rm stat, 10}(\mathbf{s}_k) + \eta \).
    \EndFor
    \State \textbf{Output.} Compute the size $\alpha$ cut-off \( \gamma_{\alpha,\varepsilon}(n) \), such that $100 (1-\alpha) \%$ of the $\{\mbox{H}_{\rm stat, 10}(\mathbf{s}_1), \ldots, \mbox{H}_{\rm stat, 10}(\mathbf{s}_N)\}$ values are greater than \( \gamma_{\alpha, \varepsilon}(n) \). 
\EndFor


\end{algorithmic}

\end{algorithm}
\footnotetext{Choice of $\tau^2_i$ for various tests($z, t, \chi^2$ and $F$) can be found in Table 1 of \cite{Johnson2023}.}
\subsection{Hyperparameter tuning }\label{sec:hyperparameter_tuning}

For the use of differentially private Bayesian tests, the concerned parties initially establish the desired privacy level $\varepsilon$. Following this determination, we need to select appropriate values for the parameters $M_n$, given $a_n =- \log(\omega_n/1-\omega_n) = k n^\beta$ with fixed $ 0 <\beta < 1, 0 < k \leq 1$.  This iterative process will allow us to experiment with diverse values of $M_n$, thereby striking a nuanced balance between attaining the desired privacy level and  asymptotic behavior of the Bayes factors. In this article, we consider the case where  we do not have the autonomy to select specific value of $\varepsilon$, but possess the discretion to choose $M_n$. Analysts, endowed with the autonomy of choosing the privacy budget $\varepsilon$,  may iteratively employ the described methodology, for varying the values of $\varepsilon$.

Given the sample size $n$, a privacy mechanism, a privacy budget $\varepsilon$, standardized effect size of interest $\pi^\star$  and the size  $\alpha$, we determine the Bayes factor cut off $\gamma_{\alpha, \varepsilon}(n)$ using \eqref{eqn:BF_cutoff} via a Monte Carlo simulation, for varying values of  $M_n$ given $a_n$. Now, let $\mathcal{M}$ denote the space all possibles pairs $M_n$ under consideration. 
We operate under the regime that $M_n = \zeta n$, $0 < \zeta \leq 1$ given $a_n = k n^\beta$. The constraints $0 <\beta < 1$  and $0 < k \leq 1$ ensure  that the $\mbox{GS}_f$ decreases as the sample size increases. The order assumption on $a_n$, or equivalently $\omega_n$, enables us to dictate the rate at which the hypotheses specified in Theorems \ref{Th:1}-\ref{Th:2} (Theorems 3.3 and 3.4 of the Supplementary Material) approach the usual/\textcolor{blue}{non-privatized null hypotheses significance testing framework}. This, in turn, enables us to dictate the convergence rate of the privatized Bayes factor. Specifically, slower the order of $a_n$, lower would be the rate of convergence. Refer to Corrollary \ref{lemma_combined_BF} and Theorem \ref{main_prop} for details.
The optimal choices $M_n$ is obtained via \emph{maximising power} of the test
$$
   \argmax_{M_n\in\mathcal{M}} \Xi_{1}(M_n\mid \varepsilon, \alpha, n)
$$
where 
$$\Xi_{1}(M_n\mid \varepsilon, \alpha, n) = \mbox{E}\big[\phi(\mathbf{s})\mid \mbox{\rm H}_1] = \mbox{P}_{\mbox{\rm H}_1}\big[\mbox{\rm H}(\mathbf{s}) \geq \gamma_{\alpha, \varepsilon}(n)\big].
$$
The numerical strategy for the hyper parameter tuning in further demonstrated in Section \ref{experiments_t} via examples.

\subsection{Bayes factors based on test statistics and convergence rates of Bayes factors}\label{sec:BFBOTS_DP}
In this section, we present the closed form expressions for differentially private Bayes factors based on common $t$,$\chi^2$ and $F$ test statistics.
Further, we study the asymptotic properties of the proposed  Bayes factors, adopting the popular notion of \emph{Conventional Bayes factor consistency} \citep{chib2016, Chatterjee2020}.
Such asymptotic properties must be satisfied to ensure that, as the sample size increases, the correct model has a faster rate of posterior contraction as a function of sample size in comparison to an incorrect model/complex model.   Results on consistency of Bayes factors based on test statistics \citep{a40ded0c-a9d6-3b1d-b191-ca0979a06298, datta2024} is not particularly new.  However, the introduction of differential privacy to this framework introduces unique challenges. To ensure the convergence of the privatized Bayes factors, it is essential to incorporate additional meaningful assumptions. This includes considerations about the distribution of test statistics under the competing hypotheses \eqref{eqn:hm2}, order of the truncation parameter $a_n$ (or equivalently $\omega_n$) and number of data partitions $M_n$ relative to the sample size, and the characteristics of the global sensitivity parameter. 

\begin{definition}[Conventional Bayes factor consistency \citep{chib2016}] For two competing models $M_l$ and $M_k$, the Bayes factor for comparing models $M_k$ and $M_l$, denoted as $\mbox{\rm BF}_{kl} = \frac{m(\mathbf{x} \mid M_k)}{m(\mathbf{x}\mid M_l)}$, is consistent if\\
(i) $\mbox{\rm BF}_{kl} \xrightarrow{p} 0$ when $M_l$ contains the true model, and\\
(ii) $\mbox{\rm BF}_{kl} \xrightarrow{p} \infty$  when $M_k$ contains the true model, where $\mathbf{x}$ is the data.
\end{definition}

Before we proceed, a few notations are in order. The confluent hyper-geometric function of order $(m, n)$ \citep{10.5555/1098650} is denoted by ${_m}F_n(a,b;z)$. The Dirac's delta maesure at $0$ is denoted by $\delta_0$. Further, $\mbox{N}(a, b)$ denotes normal distribution with a mean of $a_n$ and variance $b$, $\mbox{T}_n(\theta)$ denotes the t-distribution with degrees of freedom $\nu$  and  non-centrality parameter $\theta$, $\chi^2_n(\theta)$ denotes the chi-squared distribution with degrees of freedom $\nu$  and non-centrality parameter $\theta$, $\mbox{F}_{k,m}(\theta)$ denotes  the F-distribution with degrees of freedom $(k, m)$ and  non-centrality parameter $\theta$ , $\mbox{G}(\alpha, \theta)$ denotes a gamma distribution with a shape parameter $\alpha$ and a rate parameter $\theta$, and $\mbox{J}(\tau^2)$ represents a normal-moment density of order 1 \citep{Johnson2010}. 

\textcolor{blue}{To guarantee consistency of the Bayes factors, we operate under the asymptotic regime where the number of partitions satisfies \( M_n = \zeta n \), for some constant \( 0 < \zeta \leq 1 \), and the truncation parameter is chosen as \( a_n = k n^\beta \), with \( 0 < \beta < 1 \) and \( 0 < k \leq 1 \). These assumptions apply uniformly across all tests considered. They arise from the selection of the truncation parameter $a_n = -\log(\omega_n/1-\omega_n)$.  Recall that, we divide $n$ observations into $M_n$ partitions. This asymptotic scaling ensures that the generalized sensitivity \( \mbox{GS}_f \) vanishes as \( n \to \infty \), while preserving the rapid convergence behavior of the privatized Bayes factor.  We shall put this statement in concrete terms  in Theorem \ref{main_prop}.}

\textcolor{blue}{The generalized sensitivity $\mathrm{GS}_f$ measures the maximum change in the privatized statistic when a single observation is altered. In the proposed construction, the test statistic is computed over $M_n = \zeta n$ equally sized partitions, each containing approximately $n / M_n = 1/\zeta$ observations. As the total sample size $n$ increases, the contribution of any single observation to the aggregated (sample--aggregate) statistic is therefore diluted at rate $1/n$. Consequently, the influence of an individual data point on the privatized Bayes factor naturally diminishes as $n$ grows.
More formally, if the truncation parameter is chosen as $a_n = k n^{\beta}$ with $0 < \beta < 1$, then the sensitivity of the truncated log Bayes factor satisfies
\[
\mathrm{GS}_f = O\!\left(\frac{a_n}{n}\right)
              = O\!\left(n^{\beta - 1}\right),
\]
which converges to zero for any $\beta < 1$. This falls under the standard ``vanishing-sensitivity'' asymptotic regime in differential privacy, where the privacy cost per individual decreases as the sample size grows while the statistical power of the test continues to improve. This assumption is not only mathematically convenient but also conceptually aligned with population-level inference, in which the privacy burden per subject should shrink as more data accumulate. Similar vanishing-sensitivity regimes have been employed in recent work on differentially private hypothesis testing.}

With these notations, the Bayes Factors based on common test statistics for each partition $\mathbf{x}^{(i)},\ i\in[M_n]$ of the observed dataset are presented in Theorems \ref{Th:1}, \ref{Th:2}, and  Theorems 3.3, 3.4 of the Supplementary Material, \textcolor{blue}{in a manner consistent with the developments of \cite{Johnson2005, Johnson2010, Johnson2023}}.  

\begin{theorem}\label{Th:1}
 (\textbf{Two-sided $z$-test}) Suppose the generative  model for the test statistic $z$ under $\mbox{\rm H}_0$ and $\mbox{\rm
H}_1$ are
\begin{eqnarray}\label{prior:z}
\mbox{\rm H}_0: z \mid \lambda &\sim& \mbox{\rm N}(\lambda, 1), \qquad \lambda \mid \tau^2 \sim (1-\omega_n)\ \delta_0 + \omega_n\ \mbox{\rm J}(\tau^2)\notag \\
\mbox{\rm H}_1: z \mid \lambda &\sim& \mbox{\rm N}(\lambda,1), \qquad \lambda \mid  \tau^2 \sim \omega_n\ \delta_0 + (1 - \omega_n)\ \mbox{\rm J}(\tau^2), \quad \tau>0\notag,
\end{eqnarray}
respectively. Then, the Bayes factor in favor of the $\mbox{\rm H}_1$ is of the form
$
\mbox{\rm BF}^t_{10}(z \mid \tau^2, \omega_n) = \frac{\omega_n + (1-\omega_n) \mbox{\rm R}}{(1-\omega_n)+\omega_n \mbox{\rm R}},
$
where
$$
\mbox{\rm R} = m_1(z \mid \tau^2)/m_0(z ) = (1+\tau^2)^{ -\frac{3}{2}}\ {_1}\mbox{\rm F}_1(3/2,\ 1/2; \ \tau^2 z^2/2(1+\tau^2)).
$$
Proof is deferred to the supplementary material.

\end{theorem}

\begin{corollary}\label{lemma1}
Under the set up in Theorem \ref{Th:1}, we further assume that
(i) $n\ \mbox{\rm mod}\  M_n = 0$ and the partition specific sample sizes satisfy $n_i = n/M_n \ \forall i\in[M_n]$.
(ii) the test statistic $z^{(i)} \sim \mbox{\rm N}(\eta \sqrt{n_i}, 1)$, for some $\eta$ ; 
(iii) the hyper-parameter $\tau_i^2 = \kappa n_i$, for some $\kappa > 0$; 
(iv) $\log \left((1-{\omega_n})/{\omega_n}\right) = k n^\beta$, $ 0 <\beta < 1, 0 < k \leq 1$.  
(v) $M_n = \zeta n$, $0 < \zeta \leq 1$.
Then, under $\mathrm{H}_0$, 
$\prod_{i=1}^{M_n} \mathrm{BF}^t_{10}(z^{(i)} \mid \tau_i^2, \omega_n) = \mathrm{O}_p(c^{-n}),$
where $c$ is a positive constant.
Similarly, under $\mathrm{H}_1$, 
$\prod_{i=1}^{M_n} \mathrm{BF}^t_{01}(z^{(i)} \mid \tau_i^2, \omega_n) = \mathrm{O}_p(c^{-n}).$
This  demonstrates that the combined Bayes factor,  obtained from the partition specific Bayes factors in Theorem \ref{Th:1}, is consistent under both $\mbox{\rm H}_0$ and $\mbox{\rm H}_1$. Proof is deferred to the supplementary material.

\end{corollary}
Assumption (i) could be relaxed by permitting $n_i$ to increase linearly with $n$. This assumption is made primarily for clarity and ease of explanation. The justification for  assumption (ii) is provided by the order of commonly utilized $z$ statistics under $\mbox{H}_1$. For instance, consider the test  means of an univariate normal population $\mbox{N}(\mu, 1)$. The test statistic of interest  $z = \sqrt{n}\Bar{X}$ satisfies the assumption, with $\Bar{X}$ and $n$ being respectively  the mean and number of observations collected from the population.  Assumption (iii) stems from the selection of $\tau_i= \sqrt{n_i}\pi^*$ , where $\pi^*$ represents the standardized effect size for one-sample $z$ tests. The choice of the scale parameter $\tau_i^2$ in terms of the sample size and $\pi^*$ for various linear tests are listed in Table 1 of \cite{Johnson2023}. Assumption \emph{(iv)} and \emph{(v)} is shared across all tests, and its justification was provided at the start of this section. This selection ensures  that the $\mbox{GS}_f$ decreases as the sample size increases, while maintaining the rapid convergence of  the privatized Bayes factor.

Now, we carry on with the discussion on privatized Bayes factors based on $t$, $\chi^2$ and $F$ test statistics for each partition $\mathbf{x}^{(i)},\ i\in[M_n]$ of the observed dataset. The corresponding results for the $t$-tests are included in the main manuscript, while those for the $\chi^2$ and $F$ tests are provided in Section 3 of the Supplementary Material.

\begin{theorem}\label{Th:2}(\textbf{Two-sided $t$-test})
Suppose the generative  model for the test statistic $t$ under $\mbox{\rm H}_0$ and $\mbox{\rm H}_1$ are
\begin{eqnarray}\label{prior:t}
\mbox{\rm H}_0: t \mid \lambda &\sim& \mbox{\rm T}_\nu(\lambda), \qquad \lambda \mid  \tau^2 \sim (1-\omega_n)\ \delta_0 + \omega_n\ \mbox{\rm J}(\tau^2)\notag \\
\mbox{\rm H}_1: t \mid \lambda &\sim&
 \mbox{\rm T}_\nu(\lambda), \qquad \lambda \mid  \tau^2 \sim \omega_n\ \delta_0 + (1 - \omega_n)\ \mbox{\rm J}(\tau^2)\notag, \quad \tau>0,
 \end{eqnarray}
respectively. Then, the Bayes factor in favor of the $\mbox{\rm H}_1$ is of the form
$
\mbox{\rm BF}^t_{10}(t \mid \tau^2, \omega_n) = \frac{\omega_n + (1-\omega_n) \mbox{\rm R}}{(1-\omega_n)+\omega_n \mbox{\rm R}},
$
where 
 $$
\mbox{\rm R} = \frac{1}{(1+\tau^2)^{\frac{3}{2}}} {_2}F_{1}\Bigg( \frac{3}{2},\frac{\nu + 1}{2},\frac{1}{2},\frac{t^2\tau^2}{(t^2+\nu)(1+\tau^2)}\Bigg).
$$

Proof is deferred to the supplementary material.
 \end{theorem}
  
\begin{corollary}\label{lemma2}
Under the set up in Theorem \ref{Th:2}, we further assume that 
(i) $n\ \mbox{\rm mod}\  M_n = 0$ and the partition specific sample sizes satisfy $n_i = n/M_n \ \forall i\in[M_n]$.
(ii) the test statistic $t^{(i)} \sim \mbox{\rm T}_\nu(\eta \sqrt{n_i})$,  for some $\eta$;  
(ii) the hyper-parameters $\nu_i = \delta n_i - u$ and  
(iii) $\tau_i^2 =\kappa n_i$, for some $\kappa > 0$, 
(iv) $\log \left((1-{\omega_n})/{\omega_n}\right) = k n^\beta$, $ 0 <\beta < 1, 0 < k \leq 1$. 
(v) $M_n = \zeta n$, $0 < \zeta \leq 1$.
Then, under $\mathrm{H}_0$, 
$\prod_{i=1}^{M_n} \mathrm{BF}^t_{10}(t^{(i)} \mid \tau_i^2, \omega_n) = \mathrm{O}_p(c^{-n}),$
where $c$ is a positive constant. Similarly, under $\mathrm{H}_1$, $\prod_{i=1}^{M_n} \mathrm{BF}^t_{01}(t^{(i)} \mid \tau_i^2, \omega_n) = \mathrm{O}_p(c^{-n}).$
This  demonstrates that the combined Bayes factor,  obtained from the partition specific Bayes factors in Theorem \ref{Th:2}, is consistent under both $\mbox{\rm H}_0$ and $\mbox{\rm H}_1$. Proof is deferred to the supplementary material.

\end{corollary}
Assumption \emph{(ii)}  follows from the degrees of freedom of frequently used $t$ statistic. For instance, consider the test  means of an univariate normal population $\mbox{N}(\mu, \sigma^2)$, where $\sigma^2$ is unknown. Then, the $t$-statistic follows a $t$ distribution with degrees of freedom $n-1$. Rest of assumptions follow from considerations described after Corollary \ref{lemma1}. 

{\color{blue}Closed-form Bayes factor expressions for the $\chi^2$ and $F$ test statistics, together with their corresponding consistency properties, are provided in Section~3 of the Supplementary Material. Theorem~3.3 and Corollary~3.3.1 of the Supplementary present the explicit Bayes factor expression for the $\chi^2$ test and establish its convergence rate respectively. Likewise, Theorem~3.4 and Corollary~3.4.1 give the corresponding Bayes factor expression and convergence results for the $F$ test, respectively. We note that when $z^2 = h$, the numerical value of the Bayes factor computed from Theorem~\ref{Th:1} is identical to the value obtained using the $\chi^2$ test with $k = 1$ degree of freedom (see Theorem~3.3 of the Supplement). Similarly, when $t^2 = f$, the value of the Bayes factor based on Theorem~\ref{Th:2} coincides with that obtained from the $F$ test with $(a = 1, b)$ degrees of freedom (see Theorem~3.4 of the Supplement). The full derivations of these equivalences appear in \cite{Johnson2023} and are beyond the scope of the current manuscript.
}

\begin{corollary}\label{lemma_combined_BF}
Under the conditions in corollaries \ref{lemma1}-\ref{lemma2} and noting that the privacy noise $\eta = \mathrm{o}_p(1)$, it follows that the privatized combined logarithm of the Bayes factor, $\mbox{\rm M}_n\times \mathrm{H}_{\rm stat,10}(\mathbf{s})=\mathrm{O}_p( n)$ under $\mbox{\rm H}_0$. Similarly, $\mbox{\rm M}_n\times \mathrm{H}_{\rm stat,01}(\mathbf{s})=\mathrm{O}_p( n)$ under $\mbox{\rm H}_1$. 
\end{corollary}

Besides Bayes factor consistency, another desirable characteristic of the our privatized Bayes factors is that  both the privatized and non-privatized Bayes factors will yield similar inference as the sample size diverges to $\infty$. Comparable results are presented in  \citet{Barrientos2019}, in the context of differentially private frequentist tests for assessing significance of linear regression.
Denoting the true non-privatized Bayes factor against the null hypothesis for the $i$-th data partition by $R^{(i)}$, $i \in [M_n]$, we characterize the discrepancy between the average privatized weight of evidence $\mathrm{H}_{\rm stat,10}(\mathbf{s})$ and average non-privatized weights of evidence $\frac{1}{M_n} \sum_{i=1}^{M_n} \log(R^{(i)})$ in terms of  the probability $P\left(\left| \mathrm{H}_{\rm stat,10}(\mathbf{s}) - \frac{1}{M_n} \sum_{i=1}^{M_n} \log(R^{(i)}) \right| > c\right)$ for some $c>0$ \citep{Barrientos2019}. Under appropriate conditions, Theorem \ref{main_prop} states that, the discrepancy between the average privatized weight of evidence  and the average true non-privatized weight of evidence converges in probability to $0$. 

\begin{theorem}\label{main_prop}
Assume that the conditions  in corollaries \ref{lemma1}-\ref{lemma2} hold. Further, assume $m_1(s^{(i)})$ and $m_0({s}^{(i)})$ are strictly positive for each $i \in [M_n]$.
Then, 
\begin{equation*}
\left| \mathrm{H}_{\rm stat,10}(\mathbf{s}) - \frac{1}{M_n} \sum_{i=1}^{M_n} \log(R^{(i)}) \right| \xrightarrow{\mathbb{P}} 0,\ \mbox{as} \ n \rightarrow \infty, \quad \text{where}\quad R^{(i)} = \frac{m_1(s^{(i)} \mid \tau_{1,i}^2)}{m_0({s}^{(i)} \mid \tau_{0,i}^2)}, \ i\in[M_n].
\end{equation*}
Proof is deferred to the supplementary material.
\end{theorem}

In summary,  Theorem \ref{main_prop} formalises that  both the privatized and non-privatized Bayes factors will favor the same hypothesis with probability converging to $1$ when the sample size diverges to $\infty$. {\color{blue}Corollary~\ref{lemma_combined_BF} and Theorem~\ref{main_prop} remain valid for the $\chi^2$ and $F$ test settings, provided the respective conditions specified in Corollary~3.3.1 and Corollary~3.4.1 of the Supplement are satisfied.}


\textcolor{blue}{ We conclude the section with a remark.
While the proposed prior specification  in Equation \ref{eqn:hm2} assumes a symmetric mixing weight $\omega_{1,n}=\omega_{2,n}=\omega_n$, an alternative and more general formulation is given by 
\begin{align}
    &\theta^{(i)}\mid\tau_{0,i}, \tau_{1,i}\sim (1-\omega_{1, n})\ \pi_{0}(\theta^{(i)} \mid \tau_{0,i}^2)\ +\ \omega_{1, n}\ \pi_{1}(\theta^{(i)} \mid \tau_{1,i}^2) \quad \text{under}\ \mbox{H}_0,\notag\\
    & \theta^{(i)}\mid\tau_{0,i}, \tau_{1,i}\sim\ \omega_{2, n}\ \pi_{0}(\theta^{(i)}\mid \tau_{0,i}^2)\ +\ (1 - \omega_{2, n})\ \pi_{1}(\theta^{(i)} \mid \tau_{1,i}^2)\quad \text{under}\ \mbox{H}_1,
\end{align}
In this case, the non-private Bayes factor is bounded by 
\[
\left[\frac{\omega_{1,n}}{1-\omega_{2,n}},\;
      \frac{1-\omega_{2,n}}{\omega_{1,n}}\right],
\]
and one may define 
\[
a_n = \log\!\left(\frac{1-\omega_{2,n}}{\omega_{1,n}}\right),
\qquad 
b_n = \log\!\left(\frac{\omega_{1,n}}{1-\omega_{2,n}}\right),
\qquad 
\mbox{GS}_f = a_n - b_n,
\]
thereby allowing separate tuning of $(a_n)$ and $(b_n)$ rather than a single sensitivity sequence. Although this extension offers additional flexibility, the theoretical development becomes algebraically more involved. The symmetric specification used in this paper, equivalently $b_n=-a_n$ and $\mbox{GS}_f=2a_n$, leads to the closed-form expression 
\[
\omega_n = \frac{1}{1+e^{k n^\beta}}, 
\qquad 0<\beta<1,\; 0<k\le 1,
\]
which enables a more transparent presentation of the Bayes factor consistency results. For this reason, we adopt the symmetric model in the main exposition, while noting that the general case can be treated analogously with no change to the final asymptotic conclusions.
}
\section{Experiment: Test for Normal means ($z$ or $t$-test)}\label{experiments_t}


Suppose a confidential database contains samples $x_1, \ldots, x_n$ generated from a univariate  $\mbox{N}(\mu, \sigma^2)$ with unknown mean $\mu$ and unknown variance $\sigma^2$. The database releases response to an analyst's queries,  upon privatization via Laplace mechanism with a fixed privacy budget $\varepsilon$. Without the loss of generality, we assume $\varepsilon \in\{1, 1.5, 2\}$. In this context, we intend to utilize differentially private Bayes factors based on $t$-statistics, to test the null hypothesis  against the alternative as described in Theorem \ref{Th:2}, for an assumed  effect size of interest $\pi^{\star}=0.01$.
We set $a_n = \log \left((1-{\omega_n})/{\omega_n}\right) = k n^\beta$ with $\beta = 1- 10^{-2}$ and $k = 1$, in line with the assumptions in Theorem \ref{Th:2}.
For the purposes of producing the power curves via numerical experiments, suppose the data  $x_1, \ldots, x_n$ in the confidential database is  generated from $\mbox{N}(\mu, 1)$, where $\mu\in\{\pm 0.01,\pm 0.02, \ldots,\pm 1\}.$ The sample size $n$ varies in an increasing grid $\{25, 50, 200, 300, 500, 1000\}$. 

Recall that, in real life application, we shall only have access to privatized log Bayes factor upon privatization via Laplace mechanism with privacy budget $\varepsilon$, the sample size $n$, and the effect size of interest $\pi^{\star}=0.01$.
For the sake of demonstration,  for a fixed value of the hyper-parameter $M_n$ given $a_n$, we first showcase the numerical schemes to determine the size $\alpha$ cut-off for the proposed differentially private Bayes factor. Given the sample size $n$, a privacy mechanism, privacy budget $\varepsilon$,  effect size of interest $\pi^{\star}=0.01$,  and the size  $\alpha$ of the test, we determine the cut off $\gamma_{\alpha, \varepsilon}(n)$ by solving the equation \eqref{eqn:BF_cutoff}, utilizing Algorithm \ref{algorithm1} based on Monte Carlo simulations. 

\begin{figure}[!htbp]
\begin{center}
\includegraphics[width=12cm, height = 4cm]{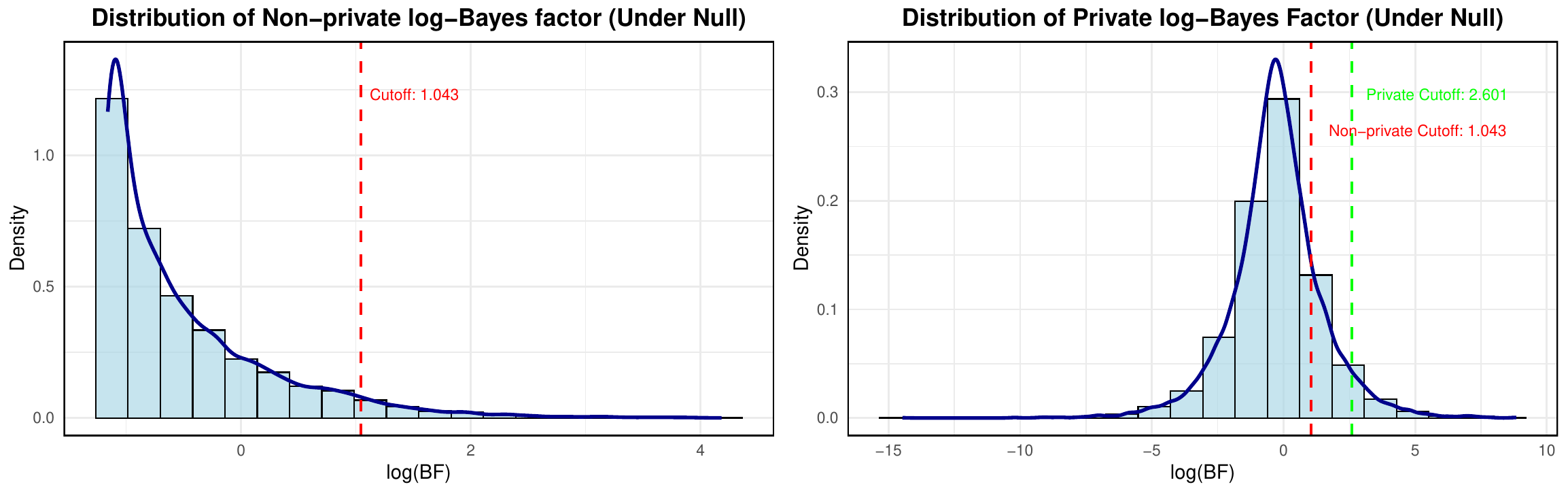} 
\caption{\textbf{Determining size $\alpha$ Bayes factor cut-off ($t$-test).} We present the distribution of log-Bayes factor in non-private and private, under $\mbox{H}_0$.  Non-private and private Bayes factor cut-offs, given size of the test $\alpha = 0.05$, sample size $n=100$ and privacy budget $\varepsilon = 1$,  with fixed hyper-parameters $M_n = 5$ given $a=3$.}\label{fig:size_t}
\end{center}
\end{figure}

In Figure \ref{fig:size_t}, we present the distribution of log-Bayes factor, under the null hypothesis, in both non-private and private set up with privacy budget $\varepsilon = 1$.  In particular, we overlay the non-private and private Bayes factor cut-offs, given size $\alpha = 0.05$, sample size $n=100$ and privacy budget $\varepsilon = 1$,  with fixed hyper-parameter $M_n= 5$ given $a= 3$.
Due to the insertion of the privacy noise, the null distribution of the differentially private log Bayes factor differ from  the null distribution of the non-private log Bayes factor, and the size $\alpha$ cut-off for differentially private log Bayes factor shifts to the right compared to the size $\alpha$ cut-off for  the non-private log Bayes factor, as expected.

\textcolor{blue}{Next, we discuss the numerical scheme to tune the hyper parameter $M_n\in\mathcal{M}$ given $a_n$.  For each value of the   $M_n$, given $(n, a_n)$, privacy mechanism, $\varepsilon$, and size  $\alpha$, we can determine the Bayes factor cut off $\gamma_{\alpha, \varepsilon}(n)$, as described earlier.  Then, the optimal choices of the hyper parameters $M_n$ is obtained via maximizing power of the privatized Bayes factor based test over $\mathcal{M}=\{2,\ldots,10\}$. We again accomplish this  via Monte Carlo simulations, similar to Algorithm \ref{algorithm1}. To that end, for each $M_n\in \mathcal{M}$ and assumed effect size $\pi^{\star}$ of interest, we  draw the pseudo test statistic $s_k^{(i)}\equiv t_k^{(i)}$ from  non-central  $t_{|n_i| - 1}(\lambda^{(i)})$ distributions with non-centrality parameter $\lambda^{(i)}$, where $$\lambda^{(i)} \sim \omega_n \delta_0 + (1- \omega_n) J(\tau_i^2),\ \tau_i = \sqrt{n_i} \pi^\star,\ i\in [M_n]$$ and $\omega_n$ satisfies the condition (iv) in \eqref{lemma1} with $k = 1$ and $\beta = 1 - 10^{-2}$. We then choose the $\hat{M}_n$ that maximizes power of the privatised Bayes factor  with  size $\alpha$ cut-off $\gamma_{\alpha, \varepsilon}(n)$. For tuning $M_n$,  it is important to note that we are only sampling the pseudo test statistics from their conditional distributions under the alternative hypothesis, where the slab part of prior mode is set by the  standardized effect size \( \pi^\star \) of interest. In this process, we are not sampling pseudo data, but solely the test statistics which only depend on the partition specific sample size and the standardized effect size of interest. This ensures the preservation of differential privacy budget of the proposed mechanism.}

\textcolor{blue}{We also note that the hyper-parameter tuning scheme remains applicable under a \emph{local} prior specification. The proposed tuning scheme does not exploit any of the separation  unique to the non-local prior $J(\tau_i^2)$; therefore, replacing $J(\tau_i^2)$ with a local prior leaves the tuning procedure unchanged.}

\begin{figure}[!htbp]
\centering

\begin{subfigure}[t]{0.4\textwidth}
\includegraphics[width=6cm, height=5cm]{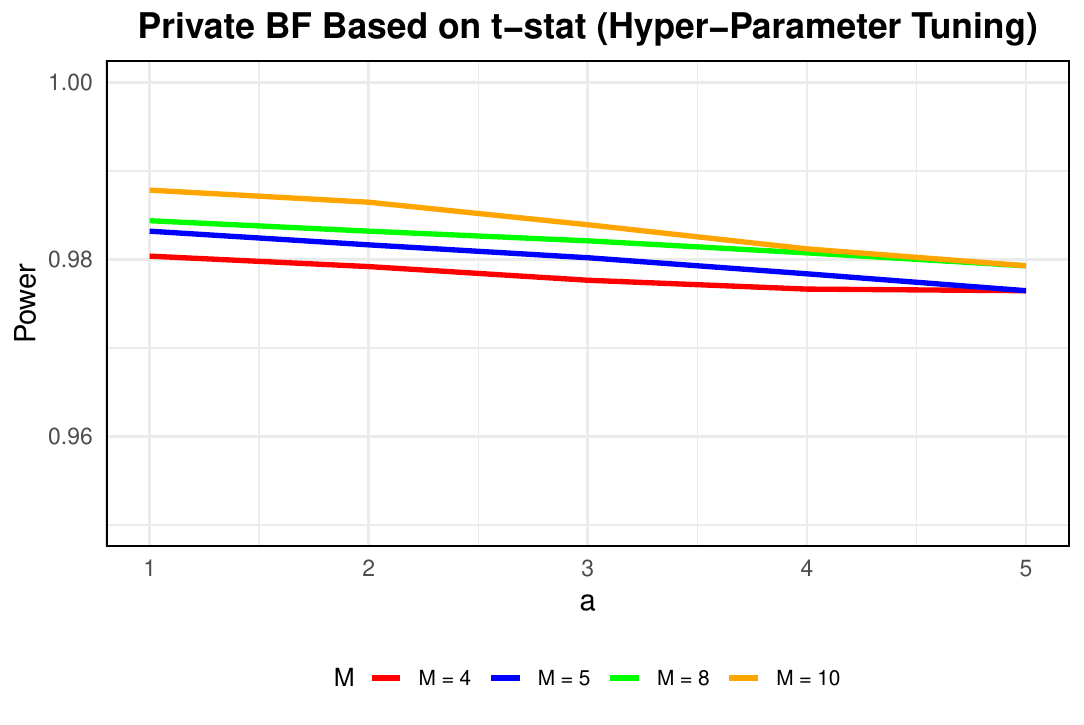} 
\caption{\textbf{Hyper-parameter tuning in size $\alpha$ Bayesian test ($t$-test).} The power of the size $\alpha$ privatized Bayesian test for sample size $n=100$ and privacy budget $\varepsilon = 1$,  with varying values of hyper-parameters $M_n$ at different fixed values of $a_n$.}
\label{fig:hpt_t}
\end{subfigure}
\hfill
\begin{subfigure}[t]{0.4\textwidth}
\includegraphics[width=6cm, height=5cm]{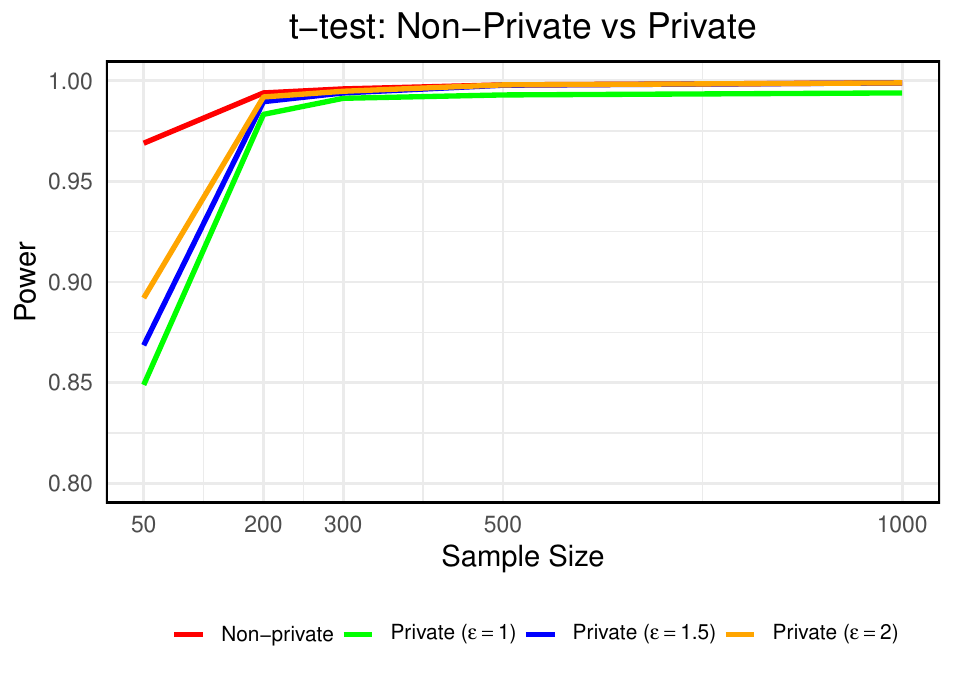} 
\caption{\textbf{Power analysis of size $\alpha$ non-private and private Bayesian $t$ test under non-local slab prior.} Comparison of the size $\alpha$ non-private Bayes factor based on $t$-statistic, and size $\alpha$ private Bayes factor based on $t$-statistic with hyper-parameters set at $\hat{M}_n$, for varying privacy budget $\varepsilon\in\{1, 1.5, 2\}$.}
\label{fig:power_t}
\end{subfigure}

\vspace{0.5cm}

\begin{subfigure}[t]{0.4\textwidth}
\includegraphics[width=\linewidth, height=5cm]{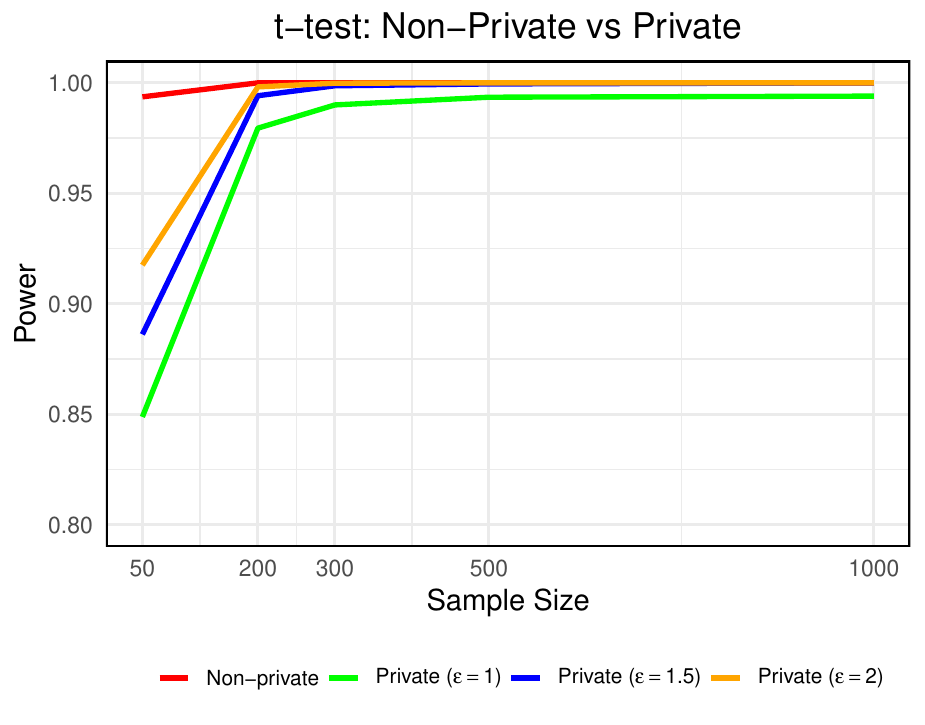}
\caption{\textcolor{blue}{\textbf{Power analysis of size $\alpha$ non-private and private Bayesian  $t$ test under true data generating mechanism with mixture prior on $\lambda$.} Comparison of the size $\alpha$ non-private Bayes factor based on $t$-statistic, and size $\alpha$ private Bayes factor based on $t$-statistic with hyper-parameters set at $\hat{M}_n$, for varying privacy budget $\varepsilon\in\{1, 1.5, 2\}$.}}
\label{fig:power_t_misspecified}
\end{subfigure}
\hfill
\begin{subfigure}[t]{0.4\textwidth}
\includegraphics[width=\linewidth, height=5cm]{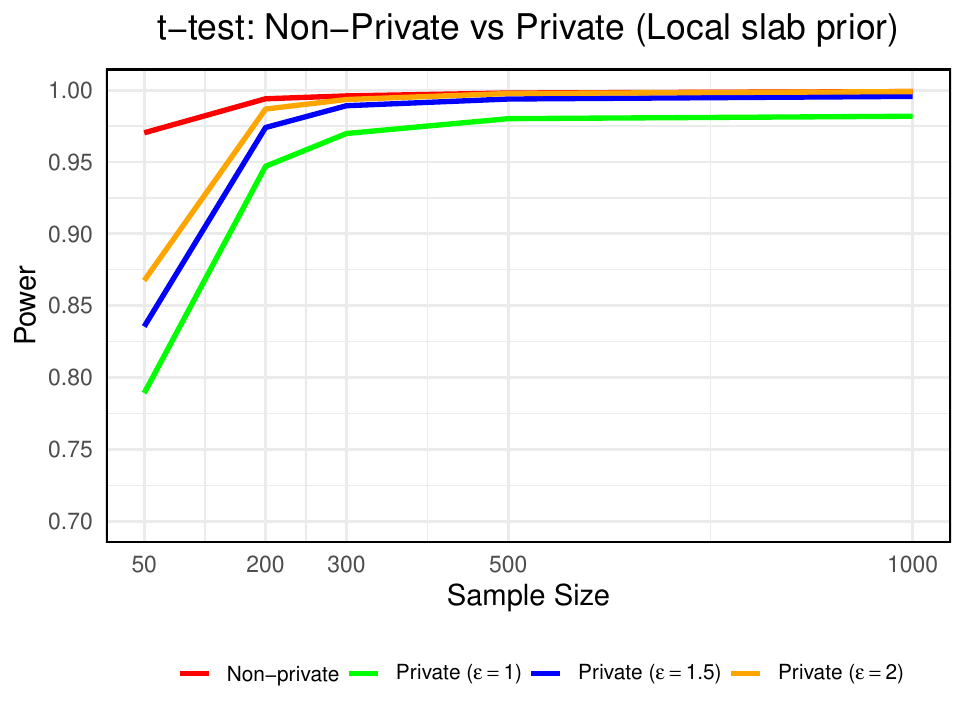}
\caption{\textbf{Power analysis of size $\alpha$ non-private and private Bayesian $t$ test under local slab prior.} Comparison of the size $\alpha$ non-private Bayes factor based on $t$-statistic, and size $\alpha$ private Bayes factor based on $t$-statistic with hyper-parameters set at $\hat{M}_n$, for varying privacy budget $\varepsilon\in\{1, 1.5, 2\}$.}
\label{fig:power_t_local}
\end{subfigure}

\caption{Comparison of non-private and private Bayesian $t$-tests under different prior specifications and privacy budgets.}
\label{fig:ttest_2x2}
\end{figure}

\textcolor{blue}{In this simulation, $\pi^{\star}\in\{\pm 0.01, \pm 0.02,\ldots,\pm 1\}$ describes the values of $\mu$ under the alternative hypothesis. Consequently, we choose the $\hat{M}_n$ that maximizes power based on $N=1000$ Monte Carlo simulations where the effect size of interest in each Monte carlo simulation is sampled uniformly from $\{\pm 0.01, \pm 0.02,\ldots,\pm 1\}$. In specific applications, the analysts usually have a pretty good idea about the effect size of interest, and the set of values of $\pi^\star$ that describes the alternative hypothesis should be updated accordingly. Refer to Figure \ref{fig:hpt_t} for a demonstration of the hyper-parameter tuning scheme.}

Finally, we compare the size $\alpha$ private Bayesian test with the size $\alpha$ non-private Bayes factor  based on $t$-statistics \citep{johnson2023bayes},  with respect to power. We consider an increasing grid of sample sizes $n\in\{25, 50, 100, 200, 500\}$. For each sample size, we proceed as described previously for determining the size $\alpha$ cut-offs and tuning the hyper-parameters $M_n$. 
\textcolor{blue}{Figure \ref{fig:power_t} presents the comparison of power of the proposed differentially private Bayesian test based on $t$-statistic with varying values of the privacy budget $\varepsilon$ and hyper-parameter tuning scheme described previous three paragraphs, with the non-private Bayesian test based on $t$-statistic proposed in \citep{Johnson2023}.} As expected, we sacrifice on power to ensure privacy, but the difference in the power of the private and non-private Bayesian tests diminishes as we increase sample size.  Further, for a fixed sample size, as the value of the privacy budget $\varepsilon$ increases,  the loss in power to ensure privacy decreases, as expected.

As a sensitivity analysis, we next present simulation results under a data generating mechanism that assumes the mixture prior on  $\lambda$. In this context, we intend to utilize differentially private Bayes factors based on $t$-statistics, to test the null hypothesis  against the alternative as described in Theorem \ref{Th:2}, for an assumed  effect size of interest $\pi^{\star}=0.01$:
\begin{eqnarray}
\mbox{\rm H}_0: z \mid \lambda &\sim& \mbox{\rm N}(\lambda, 1), \qquad \lambda \mid \tau^2 \sim (1-\omega_n)\ \delta_0 + \omega_n\ \mbox{\rm J}(\tau^2)\notag \\
\mbox{\rm H}_1: z \mid \lambda &\sim& \mbox{\rm N}(\lambda,1), \qquad \lambda \mid  \tau^2 \sim \omega_n\ \delta_0 + (1 - \omega_n)\ \mbox{\rm J}(\tau^2), \quad \tau>0\notag,
\end{eqnarray}
For the purposes of producing the power curves via numerical experiments, suppose the data  $x_1, \ldots, x_n$ in the confidential database is  generated from $\mbox{N}(\mu, 1)$, where $$\mu\in\{\pm 0.01,\pm 0.02, \ldots,\pm 1\},$$ with probability $(1 - \omega_n)$ and from $\mbox{N}(0, 1)$ with probability $ \omega_n$. Like earlier, We set $\log \left((1-{\omega_n})/{\omega_n}\right) = k n^\beta$ with $\beta = 1- 10^{-2}$ and $k = 1$, in line with the assumptions in Theorem \ref{Th:2}. The sample size $n$ still varies in an increasing grid $\{25, 50, 200, 300, 500, 1000\}$.  The  power comparison results are presented in Figure \ref{fig:power_t_misspecified}.
As earlier, we sacrifice on power to ensure privacy, but the difference in the power of the private and non-private Bayesian tests diminishes as we increase sample size.  Further, for a fixed sample size, as the value of the privacy budget $\varepsilon$ increases,  the loss in power to ensure privacy decreases, as expected.

Finally, we conduct a series of simulation studies to evaluate the utility of employing non-local priors in comparison to local priors. This investigation is motivated by the suggested use of the $g$-prior in the context of differentially private hypothesis testing for linear regression, since the $g$-prior is a local prior \cite{Pena2024}.  
\textcolor{blue}{Figure \ref{fig:power_t_local} presents the comparison of power of the proposed differentially private Bayesian test based on $t$-statistic with varying values of the privacy budget $\varepsilon$ and hyper-parameter tuning scheme described previous tow paragraphs, with the differentially private Bayesian test with local prior \citep{Pena2024}.}
When contrasted with the power curves under the non-local prior shown in Figure~\ref{fig:power_t}, it is evident that employing a non-local prior facilitates more rapid accumulation of evidence in favor of the true alternative hypothesis $\mbox{H}_1$. Consequently, this leads to substantially higher statistical power relative to the use of a local prior.

Similarly, a confidential database may have access to samples $x_1, \ldots, x_n$ from a univariate normal distribution $\mbox{N}(\mu, 1)$ with a known variance. In this context, we would utilize differentially private Bayes factors based on $z$-statistics to test the null hypothesis against the alternative. Due to the similarity of this setup with the  test of univariate normal means with unknown variance, we do not provide further numerical results in this context.

{\color{blue} Numerical experiments for the $\chi^2$ and $F$ tests are presented in Section~4 of the Supplementary Material.
}


\section{Gender differences in PHQ-8 scores from Distress Analysis Interview Corpus-Wizard of Oz 
}\label{sec:daic}

The DAIC-WOZ (Distress Analysis Interview Corpus – Wizard of Oz) dataset is a targeted subset of the larger DAIC (Distress Analysis Interview Corpus), designed to support research in the identification of mental health conditions such as depression, anxiety, and post-traumatic stress disorder (PTSD). The dataset is accessible upon request via the \href{https://dcapswoz.ict.usc.edu/}{\textcolor{blue}{DAIC-WOZ}} platform. This data collection effort was part of a broader research project aimed at developing automated systems capable of detecting both verbal and non-verbal cues associated with psychological distress. What makes DAIC-WOZ distinctive is its use of “Ellie,” a virtual interviewer. Although Ellie appears autonomous, her interactions were actually controlled in real time by a human operator in a separate location using a methodology commonly referred to as the "Wizard of Oz" technique. Each session in the dataset includes synchronized audio and video recordings along with detailed questionnaire responses. 

To assess the severity of depressive symptoms, each participant’s data is labeled using the PHQ-8 (Patient Health Questionnaire-8) score. The PHQ-8 is a validated self-report instrument similar to the PHQ-9, but it excludes the item on suicidal ideation due to ethical considerations. PHQ-8 scores range from 0 to 24 and are categorized as follows: 0–4 indicates none or minimal depression, 5–9 corresponds to mild depression, 10–14 represents moderate depression, 15–19 suggests moderately severe depression, and 20–24 reflects severe depression. For more details on the PHQ-8 questionnaire, refer to Figure~1 of the Supplement.

\begin{figure}[!htbp]
\centering
\includegraphics[width=12cm, height=6cm]{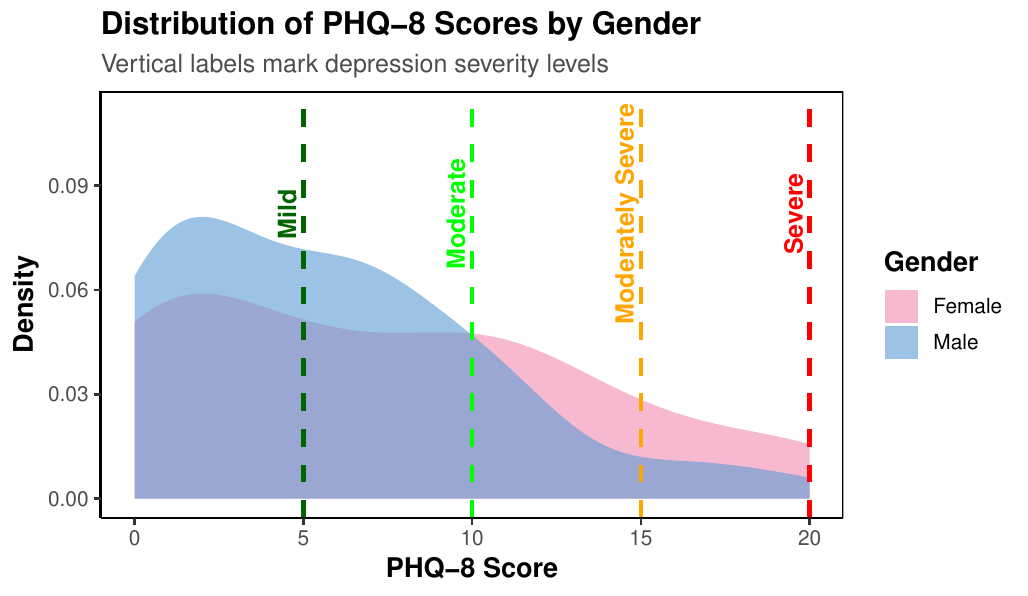}
\caption{\emph{\textbf{DAIC-WOZ database.}    The plot illustrates the gender-specific densities of the PHQ-8 scores, revealing potential differences between genders. }\label{daic_data}}
\end{figure}

The dataset comprises \( n = 107 \) participants, including 44 females and 63 males. Figure~\ref{daic_data} presents the gender-specific density  of PHQ-8 scores, revealing potential differences in the distributional patterns between males and females. These differences may reflect underlying variations in the experience, expression, or reporting of depressive symptoms across genders. It is plausible that biological gender contributes to these disparities to an extent, potentially through interactions with hormonal, neuro-biological, or psychosocial factors known to influence mental health. 

\begin{figure}
\centering
\includegraphics[width=12cm, height=6cm]{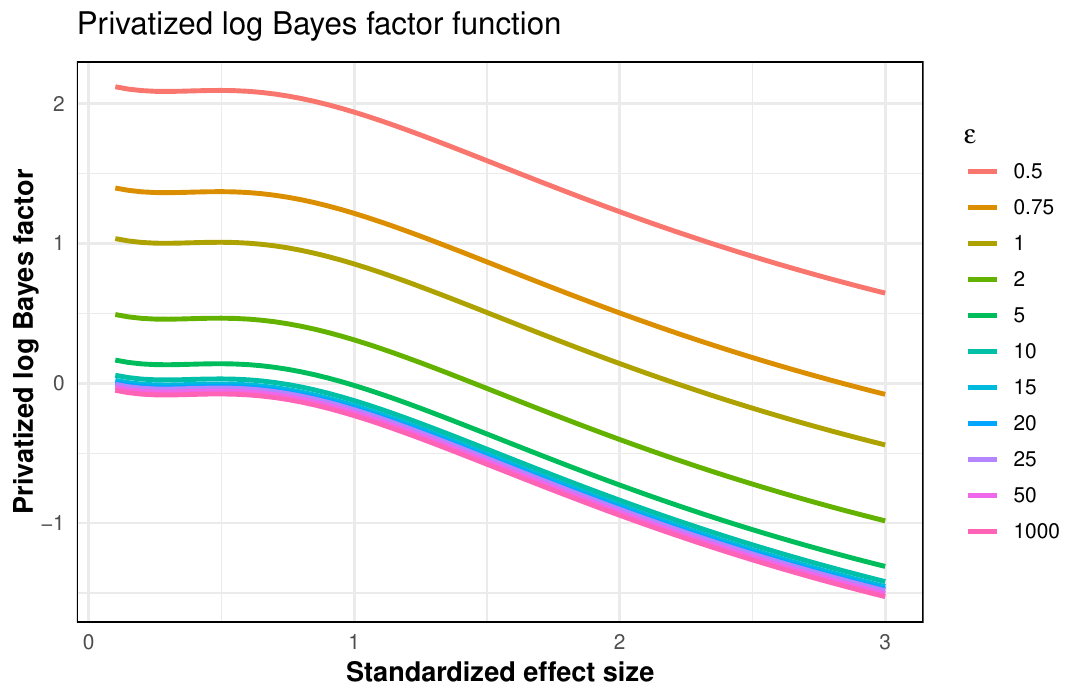}
\caption{\textcolor{blue}{\emph{\textbf{DAIC-WOZ database.}    Privatized log Bayes factor against the null hypotheses, as a function of standardized effect size, for varying values of privacy parameter $\varepsilon$.  }}\label{daic_bff}}
\end{figure}

Given the observed skewness in gender-specific PHQ-8 scores, we apply a logarithmic transformation to the variable. Let the confidential DAIC dataset contain log-transformed PHQ-8 scores for males and females, denoted by \( \{x_{g,1}, \ldots, x_{g,n_g}\} \), which are assumed to be independently drawn from a univariate normal distribution \( \mathcal{N}(\mu_g, \sigma^2) \), where \( g \in \{\mathrm{male}, \mathrm{female}\} \).  Our objective is to conduct hypothesis testing using differentially private Bayes factors based on two-sample \( t \)-statistics, to evaluate the null hypothesis \( \mbox{H}_0 : \mu_{\mathrm{male}} = \mu_{\mathrm{female}} \) against \( \mbox{H}_1 : \mu_{\mathrm{male}} \neq \mu_{\mathrm{female}} \). We assume an effect size of interest defined as
\[
\pi^{\star} = \frac{\mu_{\mathrm{male}} - \mu_{\mathrm{female}}}{\sigma} \in \{0.1, 0.2, \ldots, 3\}.
\]

In this framework, responses to analysts' queries are released through a differential privacy mechanism. Specifically, the database employs the Laplace mechanism with a fixed privacy budget \( \varepsilon \). Without loss of generality, we consider \( \varepsilon \in \{0.5, 1, 2, 5, 10, 15, 20,  \\  25, 1000\} \). We recall that, given the sample size $n=107$, a privacy mechanism, privacy budget $\varepsilon$,  effect size of interest $\pi^{\star}$,  and the size  $\alpha=0.05$ of the test, one can determine the cut-off $\gamma_{\alpha, \varepsilon}(n)$ of the size $\alpha$ Bayesian test,  by solving the equation \eqref{eqn:BF_cutoff}, utilizing Algorithm \ref{algorithm1} based on Monte Carlo simulations. 

Next, we discuss the numerical scheme to tune the hyper parameter $M_n\in\mathcal{M}$.  For each value of the   $M_n$, given $(n, a_n)$ as in Section \ref{experiments_t}, privacy mechanism, $\varepsilon$, and size  $\alpha$, we can determine the Bayes factor cut off $\gamma_{\alpha, \varepsilon}(n)$, as described earlier.  Then, the optimal choices of the hyper parameters $M_n$ is obtained via maximizing power of the privatized Bayes factor based test over $\mathcal{M}=\{1,\ldots,5\}$. We again accomplish this  via Monte Carlo simulations, similar to Algorithm \ref{algorithm1}. To that end, for each $M_n\in \mathcal{M}$ and assumed effect size $\pi^{\star}$ of interest, we  draw the pseudo test statistic $s_k^{(i)}\equiv t_k^{(i)}$ from  non-central  $t_{|n_{1i}| + |n_{2i}| - 1}(\lambda^{(i)})$ distributions with non-centrality parameter $\lambda^{(i)}$, where $$\lambda^{(i)} \sim \omega_n \delta_0 + (1- \omega_n) J(\tau_i^2),\ \tau_i = \sqrt{|n_{1i}| + |n_{2i}|} \pi^\star,\ i\in [M_n]$$ and $\omega_n$ satisfies condition in lieu with the condition (iv) in \eqref{lemma1} with $k = 1$ and $\beta = 1 - 10^{-2}$. We then choose the $\hat{M}_n$ that maximizes power of the privatized Bayes factor  with  size $\alpha$ cut-off $\gamma_{\alpha, \varepsilon}(n)$. For tuning $M_n$,  it is important to note that we are only sampling the pseudo test statistics from their conditional distributions under the alternative hypothesis, where the slab part of prior mode is set by the  standardized effect size \( \pi^\star \) of interest. In this process, we are not sampling pseudo data, but solely the test statistics which only depend on the partition specific sample size and the standardized effect size of interest. This ensures the preservation of differential privacy budget of the proposed mechanism.

In this case study, $\pi^{\star}\in\{\pm 0.1, \pm 0.2,\ldots,\pm 3\}$ describes the values of parameter of interest under the alternative hypothesis. 
In specific applications, the analysts usually have a pretty good idea about the effect size of interest, and the set of values of $\pi^\star$ that describes the alternative hypothesis should be updated accordingly. Finally, we compare the performance of size-\( \alpha \) private Bayesian tests, based on \( t \)-statistics, across varying values of the privacy budget \( \varepsilon \), in terms of the strength of evidence against the null hypothesis. 
For each value of \( \varepsilon \), we follow the procedure described previously to determine the size-\( \alpha \) decision thresholds and to tune the hyperparameters \( M_n \). The comparison results are presented in Figure~\ref{daic_bff}. For small effect sizes of interest, the test yields moderate evidence against the null hypothesis. However, as the assumed effect size increases, the evidence against the null diminishes sharply. This underscores that, for small effect sizes of interest, there are notable gender-specific differences in log PHQ-8 scores. However, these differences are not discernible when considering only relatively larger effect sizes. Furthermore, increasing the privacy budget \( \varepsilon \) leads to a consistent decrease in the strength of evidence against the null hypothesis, which is expected given the additional noise introduced by stronger privacy constraints.

\section{Discussion}
In scientific applications where data are not confidential, Bayesian hypothesis tests are commonly employed in reporting outcomes. The widespread adoption of Bayesian tests stems from their ability to address critical limitations of p-values, particularly  lack of interpretability and  failure to provide a quantitative measure of evidence supporting competing hypotheses. In this article,
we presented a novel differentially private Bayesian hypothesis testing methodology, coherently embedding popular tools from existing literature (e.g., subsample and aggregate, truncation of the test function, etc.) within a generative model framework. In contrast to existing frequentist approaches, the proposed method enables us to systematically accumulate evidence in support of the true hypothesis—a feature desirable in reporting scientific discoveries utilizing confidential data. To mitigate the computational complexities associated with computing marginal likelihoods of the competing hypotheses arising from fully parametric models, we devised our differentially private Bayes factors based on commonly used test statistics. Under  appropriate asymptotic regime, we derive the consistency rates of the proposed differentially private Bayes factors. 

\textcolor{blue}{
An important direction for future work concerns the development of privacy-preserving, data-driven procedures for selecting hyperparameters in frquentist or Bayesian hypothesis testing. In the present article, the choice of the hyperparameter $M_n$ is intentionally not data-dependent, since such dependence would consume additional privacy budget. This design choice is consistent with current practice in the differential privacy literature (e.g., \cite{Barrientos2019}). Fors instance, in the context of the $t$-test setting  Section~4, we adopt a simulation-based calibration strategy: for each candidate value $M_n\in\mathcal{M}$ given $a_n$, and for fixed sample size $n$, privacy parameter $\varepsilon$, and nominal size $\alpha$, we compute the Bayes factor rejection threshold $\gamma_{\alpha,\varepsilon}(n)$ and select $\hat{M}_n$ as the maximizer of the empirical power of the privatized Bayes factor test over the grid $\mathcal{M}$. Importantly, the simulation draws only from the conditional distribution of the test statistics under the alternative, and does not require sampling pseudo data, thereby preserving the differential privacy budget of the mechanism.
}

Although the finite sample frequentist operating characteristics of the proposed tests are explored via several numerical experiments, 
concrete theoretical study of the finite sample properties of the proposed tests is beyond the scope of the current article and presents an interesting avenue for future inquiry. Moreover, developing differentially private Bayesian tests tailored to scenarios beyond those described in this work—such as sequential testing \citep{SBF2016} and multiple comparison \citep{BERRY1999215} —holds significant practical utility.

\section*{Description of the supplementary material}
Supplementary material to ``Differentially private Bayesian tests" \citep{Supplement2025} contains  proof of theoretical results and remarks from Sections \ref{sec:key_prop} and \ref{ssec:method_teststat} in the main document. {\color{blue} It also includes the closed-form Bayes factor expressions for the $\chi^2$ and $F$ test statistics, together with the corresponding convergence rates. In addition, Section~4 of the Supplementary Material presents numerical experiments for the $\chi^2$ and $F$ tests, along with log-scale visualizations for the $z$, $t$, $\chi^2$, and $F$-based Bayes factor evaluations in Section~5.
} 

\begin{acks}[Acknowledgments]
There was no external or internal funding for this work.
\end{acks}

\bibliographystyle{ba}
\bibliography{paper-ref,references}

\end{document}


\begin{frontmatter}
\title{\ Supplementary material to\\
``Differentially private Bayesian tests"}
\runtitle{}

\begin{aug}
\author{\fnms{Abhisek} \snm{Chakraborty}}
\and
\author{\fnms{Saptati} \snm{Datta}}

\runauthor{Chakraborty, Datta}

\address[addr1]{Department of Statistics, Texas A\&M University, 
College Station, TX, USA}

\thankstext{1}{Both the authors contributed equally.}

\end{aug}

\end{frontmatter}

The Supplementary Material provides complete proofs of the theoretical results and remarks from Sections~2 and~3 of the main paper, as well as an algorithm for computing a privatized cutoff based on the full dataset. It also contains the Bayes factors corresponding to the $\chi^2$ and $F$ test statistics, together with their convergence rates. Finally, numerical studies for the $\chi^2$ and $F$ tests, along with log-scale visualizations of the figures from Section~4 of the main paper and those pertaining to the $\chi^2$ and $F$ experiments, are presented at the end of the document.


\section{Proofs of lemmas and corollaries of Section 2}

 \begin{proof}[\emph{Proof of Lemma 2.1:}]
  If $\frac{m_1(\mathbf{x}^{(i)}\mid\tau_{1,i}^2)}{m_0(\mathbf{x}^{(i)}\mid\tau_{0,i}^2)} = 0$, then $\mbox{BF}^t_{10}(\mathbf{x}^{(i)} \mid \tau_{0,i}^2,\tau_{1,i}^2 ,\omega_n) = \omega_n/(1 -\omega_n)$. Additionally, if $\frac{m_1(\mathbf{x}^{(i)}\mid\tau_{1,i}^2)}{m_0(\mathbf{x}^{(i)}\mid\tau_{0,i}^2)} =\infty$, then $\mbox{BF}^t_{10}(\mathbf{x}^{(i)} \mid \tau_{0,i}^2,\tau_{1,i}^2 ,\omega_n) = (1- \omega_n)/\omega_n$. Next, we define a map
  \begin{align*}
      g(t) = \frac{\omega_n \ +\ (1-\omega_n) t}{(1-\omega_n) \ +\ \omega_n t}, \quad t\in[0, \infty).
  \end{align*}
  Note that, the derivative of $g(t)$,
  \begin{align*}
      g^{\prime}(t) = \frac{1 - 2\omega_n}{(1 \ -\ \omega_n \ +\ \omega_n t)^2} > 0
  \end{align*}
  for $\omega_n\in(0, 1/2)$. So, $g(t)$ is increasing in $t\in[0, \infty)$, for $\omega_n\in(0, 1/2)$. This  implies that the minimum and maximum  attainable evidence against the null hypothesis is  $\frac{\omega_n}{1-\omega_n}$ and $\frac{1-\omega_n}{\omega_n}$, respectively.
\end{proof}

\begin{proof}[\emph{Proof of Lemma 2.2}]
Given $\tau_{0,i}$, $\tau_{1,i}$ and $\omega_n\in(0, 1/2)$, the prior distributions on $\theta^{(i)}$ are independent across data partitions $\mathbf{x}^{(i)},\ i\in[M_n]$. \textcolor{blue}{Define $\tau^2_0 = \{\tau^2_{0,i}\}_{i=1}^{M_n}$, $\tau^2_1 = \{\tau^2_{1,i}\}_{i=1}^{M_n}$}.Consequently, the joint marginal $m_k^t(\mathbf{x} \mid \tau^2_k, \omega_n) = \prod_{i=1}^M m_k^t(\mathbf{x}^{(i)} \mid \tau^2_k, \omega_n)$, where $k\in\{0, 1\}$. Further,  we note that
\begin{align*}
    \mbox{BF}^t_{10}(\mathbf{x} \mid \tau_{0}^2,\tau_{1}^2 ,\omega_n)
    &= \frac{m_1^t(\mathbf{x} \mid \tau_{0}^2,\tau_{1}^2 ,\omega_n)}{m_0^t(\mathbf{x} \mid \tau_{0}^2,\tau_{1}^2 ,\omega_n)} \notag\\
    &= \frac{\prod_{i=1}^M m_1^t(\mathbf{x}^{(i)}\mid \tau_{0,i}^2,\tau_{1,i}^2 ,\omega_n)}{\prod_{i=1}^M m_0^t(\mathbf{x}^{(i)}\mid \tau_{0,i}^2,\tau_{1,i}^2 ,\omega_n)} \\
    &= \prod_{i=1}^M \mbox{BF}^t_{10}(\mathbf{x}^{(i)} \mid \tau_{0,i}^2,\tau_{1,i}^2 ,\omega_n).
\end{align*}
This concludes the proof.
\end{proof}

\section{\textcolor{blue}{Algorithm to compute privatized Bayes factor based on the full data}}
\textcolor{blue}{We note that for a broad class of models, a sufficient test statistic with a tractable null distribution may not exist, even though the full data-generating distribution under both the null and alternative hypotheses is known. In such settings, the general framework described in Section 2 in the main document can be applied directly to the full dataset, without reduction to a test statistic. An explicit algorithm for computing the corresponding Bayes factor cutoff in this full-data setting is provided below by Algorithm \ref{algorithm2_full}. The workflow to compute Bayes factor cut off using the whole data is given in Algorithm \ref{algorithm2_full}. The hyper-parameters $(\tau^2_{0,i}, \tau^2_{1,i})$ should be chosen in a data-independent manner to ensure privacy.}

\begin{algorithm}[ht!]

\caption{\textcolor{blue}{(Bayes factor cut-off for full-data)}}\label{algorithm2_full}
\begin{algorithmic}[1]
\State \textbf{Input.} (i) The number of partitions $M_n$, (ii) the  truncation level $a_n$ of the partition specific log Bayes factors, (iii) privacy parameter $\varepsilon$, (iv) \( N \): The total number of the Monte-Carlo simulations $N$ to estimate the size $\alpha$ cut-off $\gamma_{\alpha,\varepsilon}(n)$

\For{\( k = 1 \) to \( N \)}

\noindent (i) Compute of $\tau^2_{1, i},\ i = 1, 2,\ldots, M$ in terms of the effect size of interest $\pi^\star$ for each of the M partitions of the data. In particular, one needs to equate $\tau_{1, i},\ i = 1, 2,\ldots, M$ to the standardized effect size of interest.  \\

\noindent(ii) Given hyper-parameters $(\tau^2_{0,i}, \tau^2_{1,i})$, draw independent samples of the parameter of interest, 
    \[
    \theta^{(i)} \sim (1-\omega_n)\pi_{0}(\theta^{(i)} \mid \tau_{0,i}^2) + \omega_n \pi_{1}(\theta^{(i)} \mid \tau_{1,i}^2),
    \]
    where $\pi_{k}(\theta^{(i)} \mid \tau_{k,i}^2),\ k = 0,1$ is a prior density supported on $\Theta_k$, the parameter space under $H_k$.\\
    
\noindent (iii) Draw i.i.d. samples $\mathbf{x}_k^{(i)}=\{x_{j,k}^{(i)}\}_{j=1}^{n_i}$ from the density  $\pi(x \mid \theta^{(i)})$,  under $\mbox{H}_0$ independently for $j= 1, \ldots,n_i, i \in [M_n] $.\\

\noindent (iv) Compute $\mbox{\rm BF}^t_{ 10}(\mathbf{x}_k^{(i)} \mid \tau^2_{0,i}, \tau^2_{1,i} \omega_n)$ for $i \in [M_n]$.\\

\noindent(v) Define $\mathbf{x}_k = \{\mathbf{x}_k^{(i)}\}_{i=1}^{M_n}$. Compute 
\[
    f_{ 10}(\mathbf{x}_k) = \frac{1}{M_n} \sum_{i=1}^{M_n} \log \mbox{\rm BF}^t_{ 10}(\mathbf{x}_k^{(i)} \mid \tau^2_{0,i}, \tau^2_{1,i} \omega_n).
\]

\noindent(vi) Generate \( \eta \) from Laplace \( (0, \frac{2a_n}{\varepsilon M_n}) \).\\

\noindent(vii) Compute \( \mbox{H}_{ 10}(\mathbf{x}_k) = f_{10}(\mathbf{x}_k) + \eta \).
    \EndFor
    \State \textbf{Output.} Compute the size $\alpha$ cut-off \( \gamma_{\alpha,\varepsilon}(n) \), such that $100 (1-\alpha) \%$ of the $\{\mbox{H}_{ 10}(\mathbf{x}_1), \ldots, \mbox{H}_{ 10}(\mathbf{x}_N)\}$ values are greater than \( \gamma_{\alpha, \varepsilon}(n) \). 


\end{algorithmic}

\end{algorithm}

\textcolor{blue}{
Suppose we observe data $x_1,\ldots, x_n$ independently from a normal distribution $N(\mu, \sigma^2)$, where both $\mu$ and $\sigma^2$ are unknown. 
We are interested in constructing a privatized test for 
\[
\mbox{H}_0: \mu \in \Theta_0 
\qquad \text{versus} \qquad 
\mbox{H}_1: \mu \in \Theta_1,
\]
where $(\Theta_0, \Theta_1)$ forms a partition of the parameter space~$\Theta$. 
The defining feature of a non-local prior is that, under hypothesis~$\mbox{H}_k$, the prior on the parameter of interest~$\mu$ should assign little to no mass to values of parameter that are consistent with $\mbox{H}_{k'}$, where $k \neq k'$. 
Accordingly, we can specify the prior under $\mbox{H}_k$ as a density supported on $\Theta_k$, denoted by $\pi_k(\mu)$. 
For example, one may take
\[
\pi_k(\mu) \propto \mbox{N}(0, \tau^2)\times \mathbb{I}(\mu \in \Theta_k),
\]
where  $\tau^2$ is a prior scale parameter. 
This construction trivially satisfies the definition of a non-local prior, since each prior $\pi_k(\mu)$ has support only within $\Theta_k$. 
The advantage of specific non-local priors such as the \emph{normal moment} and \emph{normal inverse moment} priors \citep{johnson2023bayes, datta2024} lies in their analytical tractability: 
they often yield closed-form expressions for the Bayes factor, which facilitates computational simplicity. 
However, the key operating characteristics of non-local priors are determined by their defining separation property, rather than by any particular parametric form. 
With the specification above, one can directly adopt our general framework presented in Section~2 to construct differentially private Bayesian tests.
 }

\textcolor{blue}{Now suppose we are interested in constructing a privatized test based on test statistic for 
\[
\mbox{H}_0: \mu \in \Theta_0 
\qquad \text{versus} \qquad 
\mbox{H}_1: \mu \in \Theta_1,
\]
where $(\Theta_0, \Theta_1)$ forms a partition of the parameter space~$\Theta$. Note that, under both the hypothesis, the test statistic $t$ follows a Student's t-distribution with non-centrality parameter $\delta\geq0$. Hence, the above hypothesis testing problem reduced to
\[
\mbox{H}_0: \delta \in \Theta^{\delta}_0 
\qquad \text{versus} \qquad 
\mbox{H}_1: \delta \in \Theta^{\delta}_1,
\]
where $(\Theta^{\delta}_0, \Theta^{\delta}_1)$ is a partition of $\mathbf{R}^{+}\cup\{0\}$. Accordingly, we can specify the prior under $\mbox{H}_k$ as a density supported on $\Theta^{\delta}_k$, denoted by $\pi_k(\delta)$. 
For example, one may take
\[
\pi_k(\delta) \propto \mbox{N}(0, \tau^2)\times \mathbb{I}(\delta \in \Theta^{\delta}_k),
\]
where  $\tau^2$ is a prior scale parameter. 
This construction trivially satisfies the definition of a non-local prior, since each prior $\pi_k(\delta)$ has support only within $\Theta^{\delta}_k$. 
With the specification above, one can directly device a differentially private Bayes test based on test statistic.}

\section{Proofs of theorems and lemmas in Section 3}

There is precedence in the literature for investigating the frequentist operating characteristics of Bayesian tests, 
including the study of the null distribution of Bayes factors and the determination of size-$\alpha$ cutoffs; 
see, for example, \citep{Zhou2018,Zoh2018}. 
Similarly, a substantial body of work has examined the large-sample properties of Bayes factors—often referred to as 
\emph{Bayes factor consistency} \citep{Chatterjee2020}. 
In the same spirit, we aim to develop analogous theoretical results for our proposed framework in the present manuscript. 
From a practical standpoint, if one wishes to compare the power of Bayesian tests, 
it is natural to first determine the corresponding size-$\alpha$ cutoffs for the tests.
\begin{theorem}\label{Th:1}
 (\textbf{Two-sided $z$-test}) Suppose the generative  model for the test statistic $z$ under $\mbox{\rm H}_0$ and $\mbox{\rm
H}_1$ are
\begin{eqnarray}\label{prior:z}
\mbox{\rm H}_0: z \mid \lambda &\sim& \mbox{\rm N}(\lambda, 1), \qquad \lambda \mid \tau^2 \sim (1-\omega_n)\ \delta_0 + \omega_n\ \mbox{\rm J}(\tau^2)\notag \\
\mbox{\rm H}_1: z \mid \lambda &\sim& \mbox{\rm N}(\lambda,1), \qquad \lambda \mid  \tau^2 \sim \omega_n\ \delta_0 + (1 - \omega_n)\ \mbox{\rm J}(\tau^2), \quad \tau>0\notag,
\end{eqnarray}
respectively. Then, the Bayes factor in favor of the $\mbox{\rm H}_1$ is of the form
$
\mbox{\rm BF}^t_{10}(z \mid \tau^2, \omega_n) = \frac{\omega_n + (1-\omega_n) \mbox{\rm R}}{(1-\omega_n)+\omega_n \mbox{\rm R}},
$
where
$$
\mbox{\rm R} = m_1(z \mid \tau^2)/m_0(z ) = (1+\tau^2)^{ -\frac{3}{2}}\ {_1}\mbox{\rm F}_1(3/2,\ 1/2; \ \tau^2 z^2/2(1+\tau^2)).
$$

\end{theorem}
\begin{proof}[ Proof of Theorem 3.1]
According to the prior specification in Proposition 3.1,

\begin{eqnarray}\label{supp:m0(z)}
      m_0(z)= \frac{1}{\sqrt{2\pi}}\exp\big(-\frac{z^2}{2}\big),
\end{eqnarray}
and,
\begin{eqnarray}\label{supp:m1(z)}
     m_1(z\mid \tau^2) &=& 
     \frac{\exp\big(-\frac{z^2}{2(1+\tau^2)}\big)}{\sqrt{2\pi}(2\tau^2)^{\frac{3}{2}}\Gamma(\frac{3}{2})} \int_{-\infty}^{\infty} \lambda^{2} \exp\big(-\frac{1}{2a}(\lambda -az)^2\big) \ d\lambda \nonumber \\
    & = & \frac{\exp\big(-\frac{z^2}{2(1+\tau^2)}\big)}{(2\tau^2)^{\frac{3}{2}}\Gamma(\frac{3}{2})\sqrt{\pi}} a^{\frac{3}{2}}\Gamma(r+\frac{1}{2}){_1}\mbox{F}_1\Big(-1, \frac{1}{2}, -\frac{\tau^2 z^2}{2(1+\tau^2)}\Big) \nonumber \\
    & = &\frac{\exp\big(-\frac{z^2}{2(1+\tau^2)}\big)}{\sqrt{2\pi}(1+\tau^2)^{\frac{3}{2}}} {_1}\mbox{F}_1\Big(-1,\frac{1}{2},-\frac{\tau^2 z^2}{2(1+\tau^2)}\Big) \nonumber\\
    & =& \frac{\exp\big(-\frac{z^2}{2}\big)}{\sqrt{2\pi}(1+\tau^2)^{\frac{3}{2}}} {_1}\mbox{F}_1\Big(\frac{3}{2},\frac{1}{2},\frac{\tau^2 z^2}{2(1+\tau^2)}\Big).
\end{eqnarray}
 Using equation (2.4) in the main document, \eqref{supp:m0(z)} and \eqref{supp:m1(z)}, we get the Bayes factor in Proposition 3.1.
\end{proof}
\begin{corollary}\label{lemma1}
Under the set up in Theorem \ref{Th:1}, we further assume that
(i) $n\ \mbox{\rm mod}\  M_n = 0$ and the partition specific sample sizes satisfy $n_i = n/M \ \forall i\in[M_n]$.
(ii) the test statistic $z^{(i)} \sim \mbox{\rm N}(\eta \sqrt{n_i}, 1)$, for some $\eta$ ; 
(iii) the hyper-parameter $\tau_i^2 = \kappa n_i$, for some $\kappa > 0$; 
(iv) $\log \left((1-{\omega_n})/{\omega_n}\right) = k n^\beta$, $ 0 <\beta < 1, 0 < k \leq 1$.  
(v) $M_n = \zeta n$, $0 < \zeta \leq 1$.
Then, under $\mathrm{H}_0$, 
$\prod_{i=1}^M \mathrm{BF}^t_{10}(z^{(i)} \mid \tau_i^2, \omega_n) = \mathrm{O}_p(c^{-n}),$
where $c$ is a positive constant.
Similarly, under $\mathrm{H}_1$, 
$\prod_{i=1}^{M_n} \mathrm{BF}^t_{01}(z^{(i)} \mid \tau_i^2, \omega_n) = \mathrm{O}_p(c^{-n}).$
This  demonstrates that the combined Bayes factor,  obtained from the partition specific Bayes factors in Theorem \ref{Th:1}, is consistent under both $\mbox{\rm H}_0$ and $\mbox{\rm H}_1$. 
\end{corollary}
\begin{proof}[ Proof of Corollary 3.1.1] For the sake of simplicity, denote $z^{(i)} = z $ for some $i \in [M]$ and $n_i = \frac{n}{M}$.
Consider testing  $\mbox{H}_0^{'} : z \sim \mbox{N}(0,1)$ versus $\mbox{H}_1^{'}: z \mid \lambda \sim \mbox{N}(\lambda,1), \ \lambda \mid \sigma^2, \tau^2 \sim \mbox{\rm J}(\tau^2)$. See supplementary material of \citet{datta2024}.

When $\mbox{H}_1^{'}$ is true, we have
\begin{eqnarray}
\frac{m_1(z \mid \tau^2)}{m_0(z)}
 & = & \frac{1}{(1+\tau^2)^{\frac{3}{2}}}\sum_{i=0}^{\infty} \frac{(r+\frac{1}{2})^{(i)}}{(\frac{1}{2})^{(i)}i!} \Bigg(\frac{\tau^2 z^2}{2(1+\tau^2)}\Bigg)^i \nonumber \\
 & \geq & \frac{1}{(1+\tau^2)^{\frac{3}{2}}}\sum_{i=0}^{\infty} \frac{1}{i!} \Bigg(\frac{\tau^2 z^2}{2(1+\tau^2)}\Bigg)^i =  \frac{1}{(1+\frac{\kappa n}{M})^{\frac{3}{2}}} \exp\Bigg(\frac{\tau^2 z^2}{2(1+\tau^2)}\Bigg) \nonumber. 
\end{eqnarray}
This implies,  
\begin{equation}
  \frac{m_0(z)}{m_1(z \mid \tau^2)} = \mathrm{O}_p\left(\exp\left(-\frac{cn}{M}\right)\right) =\mathrm{O}_p(1), \quad \text{for some } c > 0, 
\end{equation}
since $M_n = \zeta n$. Hence, when the alternative under our privatized framework is true,
 \begin{eqnarray}
   \mbox{BF}^t_{01}(z \mid \tau^2, \omega_n) &=& \frac{(1-\omega_n)+\omega_n\frac{m_1(z \mid \tau^2)}{m_0(z)}}{\omega_n+ (1-\omega_n)\frac{m_1(z \mid \tau^2)}{m_0(z )}}
  =  \frac{\frac{(1-\omega_n)}{\omega_n}+ \frac{m_1(z \mid \tau^2)}{m_0(z)}}{1+ \frac{(1-\omega_n)}{\omega_n}(\frac{m_1(z \mid \tau^2)}{m_0(z )})}
 =  \mathrm{O}_p(1), \   \end{eqnarray} 
where $0<\beta<1$. 

On the other hand, when the $\mbox{H}_0^{'}$ is true, we have
\begin{eqnarray}
 \frac{m_1(z \mid \tau^2)}{m_0(z)}
 =  \frac{1}{(1+\frac{\kappa n}{M})^{\frac{3}{2}}}
{_1}\mbox{F}_1\Bigg(\frac{3}{2},\frac{1}{2}; \frac{\tau^2 z^2}{2(1+\tau^2)}\Bigg)  =  O_p(\left(\frac{n}{M}\right)^{-\frac{3}{2}}) =  \mathrm{O}_p(1).
\end{eqnarray}
When the null under our privatized framework is true,
\begin{eqnarray}
   \mbox{BF}^t_{10}(z \mid \tau^2, \omega_n) = \frac{\omega_n+ (1-\omega_n)\frac{m_1(z \mid \tau^2)}{m_0(z )}}{(1-\omega_n)+\omega_n\frac{m_1(z \mid \tau^2)}{m_0(z)}}
  =  \frac{ \frac{\omega_n}{(1-\omega_n)}+ (\frac{m_1(z \mid \tau^2)}{m_0(z )})}{1 +  \frac{\omega_n}{(1-\omega_n)}\frac{m_1(z \mid \tau^2)}{m_0(z)}}
  =  \mathrm{O}_p(1).   \end{eqnarray} 
  
Finally, since, \(M_n = \zeta n\), under \(\mbox{\rm H}_0\), we have
\[
\prod_{i=1}^M \mathrm{BF}^t_{10}(z^{(i)} \mid \tau_i^2, \omega_n) = \mathrm{O}_p(c^{-n}),
\]
and under \(\mbox{\rm H}_1\), we have
\[
\prod_{i=1}^M \mathrm{BF}^t_{01}(z^{(i)} \mid \tau_i^2, \omega_n) = \mathrm{O}_p(c^{-n}).
\]



\end{proof}

\begin{theorem}\label{Th:2}(\textbf{Two-sided $t$-test})
Suppose the generative  model for the test statistic $t$ under $\mbox{\rm H}_0$ and $\mbox{\rm H}_1$ are
\begin{eqnarray}\label{prior:t}
\mbox{\rm H}_0: t \mid \lambda &\sim& \mbox{\rm T}_\nu(\lambda), \qquad \lambda \mid  \tau^2 \sim (1-\omega_n)\ \delta_0 + \omega_n\ \mbox{\rm J}(\tau^2)\notag \\
\mbox{\rm H}_1: t \mid \lambda &\sim&
 \mbox{\rm T}_\nu(\lambda), \qquad \lambda \mid  \tau^2 \sim \omega_n\ \delta_0 + (1 - \omega_n)\ \mbox{\rm J}(\tau^2)\notag, \quad \tau>0,
 \end{eqnarray}
respectively. Then, the Bayes factor in favor of the $\mbox{\rm H}_1$ is of the form
$
\mbox{\rm BF}^t_{10}(t \mid \tau^2, \omega_n) = \frac{\omega_n + (1-\omega_n) \mbox{\rm R}}{(1-\omega_n)+\omega_n \mbox{\rm R}},
$
where 
 $$
\mbox{\rm R} = \frac{1}{(1+\tau^2)^{\frac{3}{2}}} {_2}F_{1}\Bigg( \frac{3}{2},\frac{\nu + 1}{2},\frac{1}{2},\frac{t^2\tau^2}{(t^2+\nu)(1+\tau^2)}\Bigg).
$$

 \end{theorem}
\begin{proof}[ Proof of Theorem 3.2]

When the alternative is true, the marginal distribution of the test statistic is

\begin{align}\label{eqn:0}
    m_1(t\mid \tau^2,r) 
    &= m_0(t)\Bigg[c\sum_{k=0}^\infty \frac{\Big(\frac{\nu+1}{2}\Big)^{(k)}}{(\frac{1}{2})^{(k)}k!}\Bigg(\frac{t^2}{2(t^2+\nu)}\Bigg)^k \times\nonumber\\
    &\quad  \int_{-\infty}^{\infty}\lambda^{2(r+k)}\exp\Big(-\frac{\lambda^2(\tau^2+1)}{2\tau^2}\Big) \, d\lambda + \nonumber \\ 
    &\quad  c\frac{\sqrt{2}t\Gamma\Big(\frac{\nu}{2}+1\Big)}{\sqrt{t^2+\nu}\Gamma\Big(\frac{\nu+1}{2}\Big)} \times \nonumber \\
    &\quad\sum_{k=0}^{\infty}\frac{\Big(\frac{\nu+1}{2}\Big)^{(k)}}{\big(\frac{3}{2}\big)^{(k)}k!}\Bigg(\frac{t^2}{2(t^2+\nu)}\Bigg)^k \times \nonumber \\
    &\quad  \int_{-\infty}^{\infty}\lambda^{2+2k+1}\exp\Big(-\frac{\lambda^2(1+\tau^2)}{2\tau^2}\Big) \, d\lambda \Bigg]
\end{align}

    Let,
    \begin{eqnarray}
         I_1 
        & = & \int_{-\infty}^{\infty}\lambda^{2(1+k)}\exp\Big(-\frac{\lambda^2(\tau^2+1)}{2\tau^2}\Big) \ d\lambda \nonumber \\
        & = & 2\int_{0}^{\infty}\lambda^{2(1+k)}\exp\Big(-\frac{\lambda^2(\tau^2+1)}{2\tau^2}\Big) \ d\lambda \nonumber \\
        & = & \Gamma\Big(k+\frac{3}{2}\Big)\Big(\frac{2\tau^2}{1+\tau^2}\Big)^{k+\frac{3}{2}}
    \end{eqnarray}
    Assuming $\lambda^2=u$,
    \begin{eqnarray}
         I_2 
        & = &\int_{-\infty}^{\infty} \lambda^{2+2k+1} \exp\Big(-\frac{\lambda^2(1+\tau^2)}{2\tau^2}\Big) \ d\lambda \nonumber \\ 
        & = & 0
    \end{eqnarray}
    Substituting in \eqref{eqn:0}, the marginal density is
   \begin{align}\label{supp:m1(t)}
    &m_1(t\mid \tau^2) = m_0(t)\Bigg[ \frac{1}{(1+\tau^2)^{\frac{3}{2}}}  {_2}F_1\left(\frac{v+1}{2},\frac{3}{2},\frac{1}{2},\frac{\tau^2 t^2}{(t^2+\nu)(1+\tau^2)}\right) \Bigg],
\end{align}
where $m_0(t)$ denotes the density of a central $t$ distribution with $\nu$ degrees of freedom.
    Hence the Bayes factor  based on two sided $t$ statistic is 
   \begin{align}
    &BF_{10}(t\mid \tau^2) = \frac{1}{(1+\tau^2)^{\frac{3}{2}}} {_2}F_1\left(\frac{v+1}{2},\frac{3}{2},\frac{1}{2},\frac{\tau^2 t^2}{(t^2+\nu)(1+\tau^2)}\right) 
    \end{align}

    Using equation (2.4) in the main document, and \eqref{supp:m1(t)}, we get the Bayes factor in Proposition 3.2.
\end{proof}
\begin{corollary}\label{lemma2}
Under the set up in Theorem \ref{Th:2}, we further assume that 
(i) $n\ \mbox{\rm mod}\  M_n = 0$ and the partition specific sample sizes satisfy $n_i = n/M_n \ \forall i\in[M_n]$.
(ii) the test statistic $t^{(i)} \sim \mbox{\rm T}_\nu(\eta \sqrt{n_i})$,  for some $\eta$;  
(ii) the hyper-parameters $\nu_i = \delta n_i - u$ and  
(iii) $\tau_i^2 =\kappa n_i$, for some $\kappa > 0$, 
(iv) $\log \left((1-{\omega_n})/{\omega_n}\right) = k n^\beta$, $ 0 <\beta < 1, 0 < k \leq 1$. 
(v) $M_n = \zeta n$, $0 < \zeta \leq 1$.
Then, under $\mathrm{H}_0$, 
$\prod_{i=1}^M \mathrm{BF}^t_{10}(t^{(i)} \mid \tau_i^2, \omega_n) = \mathrm{O}_p(c^{-n}),$
where $c$ is a positive constant. Similarly, under $\mathrm{H}_1$, $\prod_{i=1}^M \mathrm{BF}^t_{01}(t^{(i)} \mid \tau_i^2, \omega_n) = \mathrm{O}_p(c^{-n}).$
This  demonstrates that the combined Bayes factor,  obtained from the partition specific Bayes factors in Theorem \ref{Th:2}, is consistent under both $\mbox{\rm H}_0$ and $\mbox{\rm H}_1$. 

\end{corollary}

\begin{proof}[ Proof of Corollary 3.2.1]
For the sake of simplicity, denote $t^{(i)} = t, \nu^{(i)} = \nu, $ for some $i \in [M]$ and $ n_i = n/M$ $\forall i \in [M]$.Consider testing $\mbox{H}_0^{'} : t \sim T_\nu(0)$ vs $\mbox{H}_1^{'} : t \mid \lambda \sim T_\nu(\lambda),\  \lambda \mid  \tau^2 \sim \mbox{\rm J}(\tau^2)$ \citep{datta2024}.
 Define $\psi = \frac{(\nu+1)t^2\tau^2}{2(t^2+\nu)(1+\tau^2)}$. When the $\mbox{H}_1^{'}$ is true,
 $\psi = \mathcal{O}_p\left(c\frac{n}{M}\right),c>0$.  This implies
\begin{eqnarray}
\frac{m_1(t\mid \tau^2)}{m_0(t)}
    & = &\frac{1}{(1+\tau^2)^{\frac{3}{2}}} {_2}F_{1}\Bigg( \frac{3}{2},\frac{\nu + 1}{2},\frac{1}{2},\frac{t^2\tau^2}{(t^2+\nu)(1+\tau^2)}\Bigg)\nonumber \\
    & \approx & \frac{1}{(1+\tau^2)^{\frac{3}{2}+1}} {_1}F_{1}\Bigg(\frac{3}{2}, \frac{1}{2}, \psi\Bigg)\nonumber \\
    & = & \frac{1}{(1+\tau^2)^{\frac{3}{2}}}\sum_{i=0}^{\infty}\frac{(\frac{3}{2})^{(i)}}{(\frac{1}{2})^{(i)}i!} \Bigg(\frac{(\nu+1)t^2\tau^2}{2(t^2 + \nu)(1+\tau^2)}\Bigg)^i \nonumber \\
    & \geq & \frac{1}{(1+\tau^2)^{\frac{3}{2}}}\sum_{i=0}^{\infty} \Bigg(\frac{(\nu+1)t^2\tau^2}{2(t^2 + \nu)(1+\tau^2)}\Bigg)^i \frac{1}{i!}. \nonumber  
\end{eqnarray}
This implies 
\begin{equation}
  \frac{m_0(t)}{m_1(t \mid \tau^2)} = \mathrm{O}_p\left(\exp\left(-\frac{cn}{M}\right)\right) = \mathrm{O}_p(1), \quad \text{for some } c > 0.
\end{equation}
Hence,
 
  \begin{eqnarray}
   \mbox{BF}^t_{01}(t \mid \tau^2, \omega_n) &=& \frac{(1-\omega_n)+\omega_n \frac{m_1(t \mid \tau^2)}{m_0(t)}}{\omega_n + (1-\omega_n)\frac{m_1(t \mid \tau^2)}{m_0(t )}}
  =  \frac{\frac{(1-\omega_n)}{\omega_n}+ \frac{m_1(t \mid \tau^2)}{m_0(t)}}{1+ \frac{(1-\omega_n)}{\omega_n}(\frac{m_1(t \mid \tau^2)}{m_0(t )})}
  =  \mathrm{O}_p(1).  \end{eqnarray}    
  When the $\mbox{H}_0^{'}$ is true, $\psi= O_p(1)$. Then,
     \begin{eqnarray}
    \frac{m_1(t \mid \tau^2)}{m_0(t)}
       & = &\frac{1}{(1+\tau^2)^{\frac{3}{2}}} {_2}F_{1}\Bigg( \frac{3}{2},\frac{\nu + 1}{2},\frac{1}{2},\frac{t^2\tau^2}{(t^2+\nu)(1+\tau^2)}\Bigg)\nonumber \\
    & \approx & \frac{1}{(1+\tau^2)^{\frac{3}{2}+1}} {_1}F_{1}\Bigg(\frac{3}{2}, \frac{1}{2}, \psi\Bigg)\nonumber \\
     & = & \frac{1}{(1+\tau^2)^{\frac{3}{2}}}\sum_{i=0}^{\infty}\frac{(\frac{3}{2})^{(i)}}{(\frac{1}{2})^{(i)}i!} \Bigg(\frac{(\nu+1)t^2\tau^2}{2(t^2 + \nu)(1+\tau^2)}\Bigg)^i \nonumber \\
   & = & \mathrm{O}\left(\left(\frac{n}{M}\right)^{-\frac{1}{2}-1}\right)\mathrm{O}_p(1) 
    =  \mathrm{O}_p(1). \nonumber
 \end{eqnarray}
     When the null is true,

\begin{eqnarray}
   \mbox{BF}^t_{10}(t \mid \tau^2, \omega_n) &=& \frac{\omega_n + (1-\omega_n)\frac{m_1(t \mid \tau^2)}{m_0(t )}}{(1-\omega_n)+\omega_n \frac{m_1(t \mid \tau^2)}{m_0(t)}}
 =  \frac{ \frac{\omega_n}{(1-\omega_n)}+ (\frac{m_1(t \mid \tau^2)}{m_0(t )})}{1 +  \frac{\omega_n}{(1-\omega_n)}\frac{m_1(t \mid \tau^2)}{m_0(t)}}
  = \mathrm{O}_p(1).   
\end{eqnarray} 
Since, \(M_n = \zeta n\), under \(\mbox{\rm H}_0\),
\[
\prod_{i=1}^M \mathrm{BF}^t_{10}(t^{(i)} \mid \tau_i^2, \omega_n) = \mathrm{O}_p(c^{-n}),
\]
and under \(\mbox{\rm H}_1\),
\[
\prod_{i=1}^M \mathrm{BF}^t_{01}(t^{(i)} \mid \tau_i^2, \omega_n) = \mathrm{O}_p(c^{-n}).
\]

\end{proof}
\begin{theorem}\label{Th:3}
(\textbf{$\chi^2$-test})     Suppose the generative  model for the test statistic $h$ under $\mbox{\rm H}_0$ and $\mbox{\rm H}_1$ are
\begin{eqnarray} \label{prior:h}
 \mbox{\rm H}_0:  h \mid \lambda &\sim& \chi^2_k(\lambda), \quad \lambda \mid  \tau^2 \sim (1-\omega_n)\ \delta_0 + \omega_n\ \ \mbox{\rm G}((k/2) + 1,\ 1/2\tau^2), \nonumber \\
\mbox{\rm H}_1: h \mid \lambda &\sim& \chi^2_k(\lambda), \quad \lambda \mid  \tau^2 \sim \omega_n \ \delta_0 + (1-\omega_n) \ \mbox{\rm G}((k/2) + 1,\ 1/2\tau^2),\ \tau>0.\nonumber
\end{eqnarray}
Then, the Bayes factor in favor of the $\mbox{\rm H}_1$ is of the form
$
\mbox{\rm BF}^t_{10}(h \mid \tau^2, \omega_n) = \frac{\omega_n + (1-\omega_n) \mbox{\rm R}}{(1-\omega_n)+\omega_n \mbox{\rm R}},
$
where 
$$
\mbox{\rm R} = m_1(h \mid \tau^2)/m_0(h)  =  (1+\tau^2)^{-k/2-1}  {_1}\mbox{\rm F}_1((k/2)+1,\ k/2;\ (\tau^2 h)/(2(1+\tau^2)).
$$ 
\end{theorem}
\begin{proof}[ Proof of Theorem 3.3]
Note that, under the prior specifications of Proposition 3.3,
\begin{eqnarray}\label{supp:m0(h)}
    m_0(h) = \frac{1}{2^\frac{k}{2} \Gamma(\frac{k}{2})} h^{\frac{k}{2} - 1} \exp(-\frac{h}{2}),
\end{eqnarray}
and, 
\begin{eqnarray} \label{supp:m1(h)}
 m_1(h \mid \tau^2) 
&= &\sum_{i=0}^{\infty} \int \frac{h^{\frac{k}{2}+i-1} \exp(-h / 2) 2^{-i}}{i ! 2^{\frac{k}{2}+i} \Gamma\left(\frac{k}{2}+i\right)\left(2 \tau^{2}\right)^{\frac{k}{2}+1} \Gamma\left(\frac{k}{2}+1\right)} \lambda^{i+\frac{k}{2}+1-1} \exp \left(-\frac{\lambda}{2\left(1+\tau^{2}\right)}\right) d \lambda \nonumber\\
&=&\sum_{i=0}^{\infty} \frac{h^{\frac{k}{2}+i-1} \exp(-h / 2) 2^{-i}}{i ! 2^{\frac{k}{2}+i} \Gamma\left(\frac{k}{2}+i\right)\left(2 \tau^{2}\right)^{\frac{k}{2}+1} \Gamma\left(\frac{k}{2}+1\right)} \frac{\Gamma\left(\frac{k}{2}+1+i\right) 2^{\frac{k}{2}+1+i}}{\left(1+1 / \tau^{2}\right)^{\frac{k}{2}+1+i}}\nonumber \\
& = & \frac{h^{\frac{k}{2}-1} \exp(-h / 2)}{2^{\frac{k}{2}} \Gamma\left(\frac{k}{2}+1\right)\left(1+\tau^{2}\right)^{\frac{k}{2}+1}} \sum_{i=0}^{\infty} \left[\frac{\tau^{2} h}{2\left(1+\tau^{2}\right)}\right]^{i} \frac{1}{i!} \frac{\Gamma\Big(\frac{k}{2}+1+i\Big)}{\Gamma\Big(\frac{k}{2}+i\Big)}\nonumber \\
& = & \frac{h^{\frac{k}{2}-1} \exp(-h / 2)}{2^{\frac{k}{2}} \Gamma\left(\frac{k}{2}\right)\left(1+\tau^{2}\right)^{\frac{k}{2}+1}} \sum_{i=0}^\infty \frac{(\frac{k}{2}+r)^{(i)}}{(\frac{k}{2})^{(i)}}\left[\frac{\tau^{2} h}{2\left(1+\tau^{2}\right)}\right]^{i} \frac{1}{i!} \nonumber \\
& = & \frac{m_0(h)}{\left(1+\tau^{2}\right)^{\frac{k}{2}+1}} {_1}\mbox{F}_1 \Bigg(\frac{k}{2}+1, \frac{k}{2},\frac{\tau^2 h }{2(1+\tau^2)}\Bigg).
\end{eqnarray}
See the supplementary material of \citet{datta2024}.
Using equation (2.4) in the main document, \eqref{supp:m0(h)} and \eqref{supp:m1(h)}, we get the desired Bayes factor.

\end{proof}

Note that gamma densities can be viewed as scaled $\chi^2$ densities. A $\chi^2_\nu$ density with degrees of freedom $(\nu) >3$ has zero mass at the origin. This implies $\mbox{\rm G}((k/2) + 1,\ 1/2\tau^2)$ represents a non-local density for every $k \in \mathrm{N}$. The additional assumptions for corollary \eqref{lemma3} follow from considerations described after corollaries \ref{lemma1}-\ref{lemma2}. 
\begin{corollary}\label{lemma3}
Under the set up in Theorem \ref{Th:3}, we further assume that 
(i) $n\ \mbox{\rm mod}\  M_n = 0$ and the partition specific sample sizes satisfy $n_i = n/M_n \ \forall i\in[M_n]$.
(ii) the test statistic $h^{(i)} \sim \chi^2_\nu(\eta n_i)$,  for some $\eta$,
(iii)  the hyper-parameters $\nu_i = \delta n_i - u$, and 
(iv) $\tau_i^2 = \kappa n_i$, for some $\kappa > 0$, 
(v) $\log \left((1-{\omega_n})/{\omega_n}\right) = kn^\beta$ , $ 0 <\beta < 1$. 
(vi) $M_n = \zeta n$, $0 < \zeta \leq 1$.
Then, under $\mathrm{H}_0$, $\prod_{i=1}^{M_n} \mathrm{BF}^t_{10}(h^{(i)} \mid \tau_i^2, \omega_n) = \mathrm{O}_p(c^{-n}),$  where $c$ is a positive constant. Similarly, under $\mathrm{H}_1$, $\prod_{i=1}^{M_n} \mathrm{BF}^t_{01}(h^{(i)} \mid \tau_i^2, \omega_n) = \mathrm{O}_p(c^{-n}).$
This  demonstrates that the combined Bayes factor,  obtained from the partition specific Bayes factors in Theorem \ref{Th:3}, is consistent under both $\mbox{\rm H}_0$ and $\mbox{\rm H}_1$. 

\end{corollary}
\begin{proof}[ Proof of Corollary 3.3.1]
For the sake of simplicity, denote $h^{(i)} = h $ for some $i \in [M]$ and $n_i = \frac{n}{M} \forall i$. Consider testing $\mbox{H}_0^{'} : h \sim \chi^2_\nu(0)$ vs $\mbox{H}_1^{'} : h \mid \lambda \sim \chi^2_\nu(\lambda),\ \lambda \mid  \tau^2 \sim \mbox{\rm G}((k/2) + 1,\ 1/2\tau^2)$ \citep{datta2024}.

When $\mbox{H}_1^{'}$ is true,
\begin{eqnarray}
\frac{m_1(h \mid \tau^2)}{m_0(h)}
& = & \frac{1}{(1+\tau^2)^{1+\frac{k}{2}}}\sum_{i=0}^{\infty} \frac{(1+\frac{k}{2})^{(i)}}{(\frac{k}{2})^{(i)}i!} \Bigg(\frac{\tau^2 h}{2(1+\tau^2)}\Bigg)^i \nonumber \\
 & \geq & \frac{1}{(1+\tau^2)^{1+\frac{k}{2}}}\sum_{i=0}^{\infty} \frac{1}{i!} \Bigg(\frac{\tau^2 h}{2(1+\tau^2)}\Bigg)^i \nonumber \\
& = & \frac{1}{(1+\frac{\kappa n}{M})^{1+\frac{k}{2}}} \exp\Bigg(\frac{\tau^2 h}{2(1+\tau^2)}\Bigg). 
\end{eqnarray}
This implies 
\begin{equation}
  \frac{m_0(h)}{m_1(h \mid \tau^2)} = \mathrm{O}_p\left(\exp\left(-\frac{cn}{M}\right)\right) = \mathrm{O}_p(1), \quad \text{for some } c > 0.
\end{equation}
Hence, when the alternative is true,
 \begin{eqnarray}
   \mbox{BF}^t_{01}(h \mid \tau^2, \omega_n) = \frac{(1-\omega_n)+\omega_n\frac{m_1(h \mid \tau^2)}{m_0(h)}}{\omega_n+ (1-\omega_n)\frac{m_1(h \mid \tau^2)}{m_0(h )}}
  =  \frac{\frac{(1-\omega_n)}{\omega_n}+ \frac{m_1(h \mid \tau^2)}{m_0(h)}}{1+ \frac{(1-\omega_n)}{\omega_n}(\frac{m_1(h \mid \tau^2)}{m_0(h )})}
  =  \mathrm{O}_p(1),  \end{eqnarray} 
where $0<\beta<1$.

When the $\mbox{H}_0^{'}$ is true,
\begin{eqnarray}
   \frac{m_1(h \mid \tau^2)}{m_0(h)}
 & = & \frac{1}{(1+\frac{\kappa n}{M})^{1+\frac{k}{2}}}
{_1}\mbox{F}_1\Bigg(1+\frac{k}{2},\frac{k}{2}; \frac{\tau^2 z^2}{2(1+\tau^2)}\Bigg) 
 =  O_p(1).
    \end{eqnarray}
   When the null is true,
\begin{eqnarray}
   \mbox{BF}^t_{10}(h \mid \tau^2, \omega_n) &=& \frac{\omega_n+ (1-\omega_n)\frac{m_1(h \mid \tau^2)}{m_0(h )}}{(1-\omega_n)+\omega_n\frac{m_1(h \mid \tau^2)}{m_0(h)}}
  = \frac{ \frac{\omega_n}{(1-\omega_n)}+ (\frac{m_1(h \mid \tau^2)}{m_0(h )})}{1 +  \frac{\omega_n}{(1-\omega_n)}\frac{m_1(h \mid \tau^2)}{m_0(h)}}
 =  \mathrm{O}_p(1).   \end{eqnarray} 
Since, \(M_n = \zeta n\), under \(\mbox{\rm H}_0\),
\[
\prod_{i=1}^M \mathrm{BF}^t_{10}(h^{(i)} \mid \tau_i^2, \omega_n) = \mathrm{O}_p(c^{-n}),
\]
and under \(\mbox{\rm H}_1\),
\[
\prod_{i=1}^M \mathrm{BF}^t_{01}(h^{(i)} \mid \tau_i^2, \omega_n) = \mathrm{O}_p(c^{-n}).
\]
    
\end{proof}

\begin{theorem}\label{Th:4}
(\textbf{$\mbox{\rm F}$- test})  Suppose the generative  model for the test statistic $F$ under $\mbox{\rm H}_0$ and $\mbox{\rm H}_1$ are
\begin{eqnarray} \label{prior:f}
 \mbox{\rm H}_0:  f \mid \lambda &\sim& \mbox{\rm F}_{\nu,b}(\lambda), \quad \lambda \mid  \tau^2 \sim (1-\omega_n)\ \delta_0 + \omega_n\ \ \mbox{\rm G}((a/2) + 1,\ 1/2\tau^2), \nonumber \\
\mbox{\rm H}_1: f \mid \lambda &\sim& \mbox{\rm F}_{\nu,b}(\lambda), \quad \lambda \mid  \tau^2 \sim \omega_n\ \delta_0 + (1-\omega_n) \ \mbox{\rm G}((a/2) + 1,\ 1/2\tau^2),\ \tau>0.\nonumber
 \end{eqnarray}
Then, the Bayes factor in favor of the $\mbox{\rm H}_1$ is of the form
$
\mbox{\rm BF}^t_{10}(f \mid \tau^2, \omega_n) = \frac{\omega_n + (1-\omega_n) \mbox{\rm R}}{(1-\omega_n)+\omega_n \mbox{\rm R}},
$
where 
$$
\mbox{\rm R} = m_1\left(f \mid \tau^2\right)/m_0\left(f\right)  = (1+\tau^2)^{-\frac{a}{2}-1} {_2}\mbox{\rm F}_1\left(\frac{a}{2}+1,\ \frac{a+b}{2},\ \frac{a}{2};\ \frac{af\tau^2}{(1+\tau^2)(b+af)}\right). 
$$ 
\end{theorem}
\begin{proof}[ Proof of Theorem 3.4]

Note that, under the prior specifications of Proposition 3.4,
\begin{eqnarray}\label{supp:m0(f)}
     m_0(f)=\frac{1}{\beta(\frac{a}{2},\frac{b}{2})}\Big(\frac{a}{b}\Big)^{\frac{a}{2}} f^{\frac{a}{2} - 1}\Big(1+\frac{a}{b}f\Big)^{-\frac{a+b}{2}}
\end{eqnarray}
    and,   
\begin{eqnarray}\label{supp:m1(f)}
     m_1(f\mid \tau^2) 
    &=& \int_{0}^{\infty} \sum_{i=0}^{\infty} \frac{\exp \Big[-\frac{\lambda}{2}\Big] \Big(\frac{\lambda}{2}\Big)^{i} \Gamma(\frac{a}{2}+\frac{b}{2}+i)}{i!\Gamma(\frac{a}{2}+i)\Gamma(\frac{b}{2})} \Big(\frac{a}{b}\Big)^{\frac{a}{2}+i} \Big(\frac{b}{b+af}\Big)^{\frac{a}{2}+\frac{b}{2}+i} \times \nonumber \\ &  & \quad \quad \quad \quad \quad \quad \quad \quad \quad \quad \quad\quad \quad \quad f^{\frac{a}{2}+i-1} \frac{\lambda^{\frac{b}{2}+1-1} \exp\Big[-\frac{\lambda}{2\tau^2}\Big]}{\Gamma(\frac{a}{2}+1)(2\tau^2)^{\frac{a}{2}+1}} \ d\lambda \nonumber \\
    &=& \sum_{i=0}^{\infty} \frac{\Gamma(\frac{a}{2}+\frac{b}{2}+i)}{2^i i! \Gamma(\frac{a}{2}+i)\Gamma(\frac{b}{2})} \Big(\frac{a}{b}\Big)^{\frac{a}{2}+i} \Big(\frac{b}{b+af}\Big)^{\frac{a}{2}+\frac{b}{2}+i} \times \nonumber \\
    &  & \quad \quad \quad \quad \quad \quad \quad \quad \quad \quad \quad \quad \quad   f^{\frac{a}{2}+i-1}  \frac{\Gamma(i+\frac{a}{2}+1)}{\Gamma(\frac{a}{2}+1)(2\tau^2)^{\frac{a}{2}+1}} \Big(\frac{2\tau^2}{\tau^2 + 1}\Big)^{i+\frac{a}{2}+1} \nonumber \\
    & = & \frac{\Gamma\Big(\frac{a+b}{2}\Big)}{\Gamma\big(\frac{a}{2}\big)\Gamma\big(\frac{b}{2}\big)} \big(\frac{a}{b}\big)^{\frac{a}{2}}\big(1+\frac{af}{b}\big)^\frac{a+b}{2} f^{\frac{a}{2}-1}\big(\frac{1}{1+\tau^2}\big)^{\frac{a}{2}+1} \times \nonumber \\ 
    & & \quad \quad \quad \quad \quad \quad \quad \quad \quad \quad \quad \quad \quad \quad \sum_{i=0}^{\infty} \frac{\big(\frac{a+b}{2}\big)^{(i)}\big(\frac{a}{2}+1\big)^{(i)}}{\big(\frac{a}{2}\big)^{(i)}i!}\Big(\frac{af\tau^2}{(b+af)(1+\tau^2)}\Big)^i \nonumber \\
    &=& \frac{m_0(f)}{(1+\tau^2)^{\frac{a}{2}+1}} {_2}F_{1}\Big(\frac{a+b}{2}, \frac{a}{2}+1,\frac{a}{2},\frac{af\tau^2}{(b+af)(1+\tau^2)}\Big)  \nonumber. \\ 
\end{eqnarray}

See the supplementary material of \citet{datta2024}. Using equations equation (2.4) in the main document, \eqref{supp:m0(f)} and \eqref{supp:m1(f)}, we get the desired Bayes factor.
\end{proof}

The additional assumptions for corollary \eqref{lemma4} follow from considerations described after corollaries \ref{lemma1}-\ref{lemma2}. 

Finally, building on the corollaries \ref{lemma1}-\ref{lemma4} on the consistency of the combined Bayes factors, we determine the convergence rate of the combined privatized Bayes factor, $\mathrm{H}_{\rm stat,10}$
\begin{corollary}\label{lemma4} 
Under the set up in Theorem \ref{Th:4}, we further assume that 
(i) $n\ \mbox{\rm mod}\  M_n = 0$ and the partition specific sample sizes satisfy $n_i = n/M_n \ \forall i\in[M_n]$.
(ii) the test statistic $f^{(i)} \sim \mbox{\rm F}_{\nu_i,b_i}(\eta n_i)$,  for some $\eta>0$ ; 
(iii) the hyper-parameter $b_i = \delta n_i - u$ and; 
(iv) $\log \left((1-{\omega_n})/{\omega_n}\right) = kn^\beta$ , $ 0 <\beta < 1$.
(vi) $M_n = \zeta n$, $0 < \zeta \leq 1$.
Then, under $\mathrm{H}_0$, $\prod_{i=1}^{M_n} \mathrm{BF}^t_{10}(f^{(i)} \mid \tau_i^2, \omega_n) = \mathrm{O}_p(c^{-n}),$ where $c$ is a positive constant. Similarly, under $\mathrm{H}_1$, $\prod_{i=1}^{M_n} \mathrm{BF}^t_{01}(f^{(i)} \mid \tau_i^2, \omega_n) = \mathrm{O}_p(c^{-n}).$
This  demonstrates that the combined Bayes factor,  obtained from the partition specific Bayes factors in Theorem \ref{Th:4}, is consistent under both $\mbox{\rm H}_0$ and $\mbox{\rm H}_1$. 

\end{corollary}

\begin{proof}[ Proof of Corollary 3.4.1]
For the sake of simplicity, denote $f^{(i)} = f, \nu_i = \nu $ for some $i \in [M_n]$ and $n_i = \frac{n}{M_n} \forall i$.   Consider testing $\mbox{H}_0^{'} : f \sim F_{\nu,b}(0)$ vs $\mbox{H}_1^{'} : f \mid \lambda \sim F_{\nu,b}(\lambda),\  \lambda \mid  \tau^2 \sim \mbox{\rm G}((k/2) + 1,\ 1/2\tau^2)$ \citep{datta2024}.
 Define $\psi = \frac{(\nu+b)\nu f\tau^2}{2(b+\nu f)(1+\tau^2)}$. When the $\mbox{H}_1^{'}$ is true,
 $\psi = \mathcal{O}_p(cn/M_n),c>0$.  This implies
\begin{eqnarray}
\frac{m_1(f\mid \tau^2)}{m_0(f)}
    & = &\frac{1}{(1+\tau^2)^{\frac{\nu}{2}+1}} {_2}F_{1}\Bigg( \frac{\nu}{2}+1,\frac{\nu+b}{2},\frac{\nu}{2},\frac{\nu f\tau^2}{(b+\nu f)(1+\tau^2)}\Bigg)\nonumber \\
    & \approx & \frac{1}{(1+\tau^2)^{\frac{\nu}{2}+1}} {_1}F_{1}\Bigg(\frac{\nu}{2}+1, \frac{\nu}{2}, \psi\Bigg)\nonumber \\
    & = & \frac{1}{(1+\tau^2)^{\frac{\nu}{2}+1}}\sum_{i=0}^{\infty}\frac{(\frac{\nu}{2}+1)^{(i)}}{(\frac{\nu}{2})^{(i)}i!} \Bigg(\frac{(\nu+b)\nu f\tau^2}{2(b+\nu f)(1+\tau^2)}\Bigg)^i \nonumber \\
    & \geq & \frac{1}{(1+\tau^2)^{\frac{\nu}{2}+1}}\sum_{i=0}^{\infty} \Bigg(\frac{(\nu+b)\nu f\tau^2}{2(b+\nu f)(1+\tau^2)}\Bigg)^i \frac{1}{i!}. \nonumber  
\end{eqnarray}
This implies 
\begin{equation}
  \frac{m_0(f)}{m_1(f \mid \tau^2)} = \mathrm{O}_p\left(\exp\left(-\frac{cn}{M_n}\right)\right) = \mathrm{O}_p(1), \quad \text{for some } c > 0.
\end{equation}
Hence,
 
  \begin{eqnarray}
   \mbox{BF}^t_{01}(f \mid \tau^2, \omega_n) &=& \frac{(1-\omega_n)+\omega_n\frac{m_1(f \mid \tau^2)}{m_0(f)}}{\omega_n+ (1-\omega_n)\frac{m_1(f \mid \tau^2)}{m_0(f )}}
  =  \frac{\frac{(1-\omega_n)}{\omega_n}+ \frac{m_1(f \mid \tau^2)}{m_0(f)}}{1+ \frac{(1-\omega_n)}{\omega_n}(\frac{m_1(f \mid \tau^2)}{m_0(f )})}
  =  \mathrm{O}_p(1).  \end{eqnarray}    
  When the $\mbox{H}_0^{'}$ is true, $\psi= O_p(1)$. Then,
     \begin{eqnarray}
    \frac{m_1(f \mid \tau^2)}{m_0(f)}
    & = & \frac{1}{(1+\tau^2)^{\frac{\nu}{2}+1}} {_2}F_{1}\Bigg(\frac{\nu+b}{2}, \frac{\nu}{2}+1,\frac{\nu}{2},\frac{\nu f\tau^2}{(b+\nu f)(1+\tau^2)}\Bigg) \nonumber \\
    & = & \frac{1}{(1+\tau^2)^{\frac{\nu}{2}+1}} {_2}F_{1}\Bigg( \frac{\nu}{2}+1,\frac{\nu+b}{2},\frac{\nu}{2},\frac{\nu f\tau^2}{(b+\nu f)(1+\tau^2)}\Bigg)\nonumber \\
    & \approx & \frac{1}{(1+\tau^2)^{\frac{\nu}{2}+1}} {_1}F_{1}\Bigg(\frac{\nu}{2}+1, \frac{\nu}{2}, \psi \Bigg)\nonumber \\
    & = & \frac{1}{(1+\tau^2)^{\frac{a}{2}+1}}\sum_{i=0}^{\infty}\frac{(\frac{\nu}{2}+1)^{(i)}}{(\frac{\nu}{2})^{(i)}i!} \Bigg(\frac{(\nu+b)\nu f\tau^2}{2(b+\nu f)(1+\tau^2)}\Bigg)^i \nonumber \\
   & = & \mathrm{O}(\left(\frac{n}{M_n}\right)^{-\frac{\nu}{2}-1})\mathrm{O}_p(1) 
    =  \mathrm{O}_p(1). \nonumber
 \end{eqnarray}
     When the null is true,

\begin{eqnarray}
   \mbox{BF}^t_{10}(f \mid \tau^2, \omega_n) &=& \frac{\omega_n+ (1-\omega_n)\frac{m_1(f \mid \tau^2)}{m_0(f )}}{(1-\omega_n)+\omega_n\frac{m_1(f \mid \tau^2)}{m_0(f)}}
 =  \frac{ \frac{\omega_n}{(1-\omega_n)}+ (\frac{m_1(f \mid \tau^2)}{m_0(f )})}{1 +  \frac{\omega_n}{(1-\omega_n)}\frac{m_1(f \mid \tau^2)}{m_0(f)}}
  = \mathrm{O}_p(1).   
\end{eqnarray} 
Since, \(M_n = \zeta n\), under \(\mbox{\rm H}_0\),
\[
\prod_{i=1}^{M_n} \mathrm{BF}^t_{10}(f^{(i)} \mid \tau_i^2, \omega_n) = \mathrm{O}_p(c^{-n}),
\]
and under \(\mbox{\rm H}_1\),
\[
\prod_{i=1}^{M_n} \mathrm{BF}^t_{01}(f^{(i)} \mid \tau_i^2, \omega_n) = \mathrm{O}_p(c^{-n}).
\]
    
\end{proof}

\begin{proof}[ Proof of Corollary 3.2.2]

It has been established that $\prod_{i=1}^{M_n} \mathrm{BF}^t_{10}(s^{(i)} \mid \tau_i^2, \omega_n) = \mathrm{O}_p(c^{-n})$ under $\mbox{\rm H}_0$ and $\prod_{i=1}^{M_n} \mathrm{BF}^t_{01}(s^{(i)} \mid \tau_i^2, \omega_n) = \mathrm{O}_p(c^{-n})$ under $\mbox{\rm H}_1$. Given that $M_n = \zeta n$, the non-privatized weights of evidence $\frac{1}{M_n}\sum_{i=1}^{M_n}\log \left(\mathrm{BF}^t_{10}(s^{(i)} \mid \tau_i^2, \omega_n)\right) = O_p(1)$ under $\mbox{\rm H}_0$ and $\frac{1}{M_n}\sum_{i=1}^{M_n}\log \left(\mathrm{BF}^t_{01}(s^{(i)} \mid \tau_i^2, \omega_n)\right) = O_p(1)$ under $\mbox{\rm H}_1$ follow accordingly.

With $\eta \sim$ Laplace$(0,\frac{2a_n}{M_n\epsilon})$ and the condition that $a_n/M_n = kn^\beta/\zeta n \rightarrow 0$ as $n \rightarrow \infty$, applying Chebyshev's inequality \citep{resnick2014probability} ensures that for any $\delta >0$, there exists an $n_0 \in \mathbb{N}$ such that
\begin{equation}
     P(|\eta| \geq \delta) \leq \frac{2a_n}{M_n \epsilon \delta^2} = \frac{2 k n^\beta}{\zeta n \epsilon \delta^2} \rightarrow 0, \ \text{for all}\ n \geq n_0.
\end{equation} 
Under $\mbox{\rm H}_0$,

\begin{equation}
\mathrm{H}_{\rm stat,10}(\mathbf{s}) = O_p(1) + o_p(1) = O_p(1).
\end{equation}
Consequently, $M_n\mathrm{H}_{\rm stat,10}(\mathbf{s}) = \zeta n O_p(1) = O_p(\zeta n)$. Defining the privatized Bayes factor against the alternative as $\mathrm{H}_{\rm stat,01}(\mathbf{s})$ and noting that $\mathrm{H}_{\rm stat,01}(\mathbf{s}) = -\mathrm{H}_{\rm stat,10}(\mathbf{s})$, we derive under $\mbox{\rm H}_1$,

\begin{equation}
\mathrm{H}_{\rm stat,01}(\mathbf{s}) = O_p(1) + o_p(1) = O_p(1).
\end{equation}
Thus, $M_n\mathrm{H}_{\rm stat,01}(\mathbf{s}) = \zeta n O_p(1) = O_p(\zeta n)$.
\end{proof}

\begin{proof}[ Proof of Theorem 3.3]
 Let us define \(\mathrm{f}_{\rm stat,10} = \frac{1}{M_n} \sum_{i = 1}^{M_n} \log \left( \frac{\omega_n+ (1-\omega_n) R^{(i)}}{(1-\omega_n) + \omega_nR^{(i)}} \right)\), and consider $ \mathrm{H}_{\rm stat,10}(\mathbf{s}) = \mathrm{f}_{\rm stat,10} + \eta$.
Then, the \emph{triangle inequality} yields,
\begin{equation} \label{i}
   \left| \mathrm{H}_{\rm stat,10}(\mathbf{s}) - \frac{1}{M_n} \sum_{i=1}^{M_n} \log(R^{(i)}) \right| \leq \left| \mathrm{H}_{\rm stat,10}(\mathbf{s}) - \mathrm{f}_{\rm stat,10} \right| + \left| \mathrm{f}_{\rm stat,10} - \frac{1}{M_n} \sum_{i=1}^{M_n} \log(R^{(i)}) \right|.
\end{equation}
Then, due to the \emph{Chebyshev's inequality} \citep{resnick2014probability}, given any $\delta >0$ and $\rho > 0$, there exists an $n_0 \in \mathbb{N}$ such that
\begin{equation}\label{ii}
    \mbox{\rm P}\left(\left| \mathrm{H}_{\rm stat,10}(\mathbf{s}) - \mathrm{f}_{\rm stat,10} \right| \geq \delta \right) = P(|\eta| \geq \delta) \leq \frac{2a_n}{M_n \epsilon \delta^2} = \frac{2 k n^\beta}{\zeta n \epsilon \delta^2} \leq \frac{\rho}{2}, \ \text{for all}\ n \geq n_0.
\end{equation}
This follows from our assumptions $M_n = \zeta n, a = \log \left(\frac{1-\omega_n}{\omega_n}\right) = kn^\beta $  and the fact that $\eta \sim $ Laplace$(0, \frac{2a_n}{M_n\epsilon})$.

Next, we note that,  $R^{(i)} = O_p(1)$ for all $i\in[M_n]$, under both the null and alternative hypotheses . Therefore, using the fact that $\frac{\omega_n}{1-\omega_n} = \exp(-kn^\beta)$, we have

\begin{eqnarray}
    \frac{ \frac{\omega_n}{(1-\omega_n)R^{(i)}} +  1}{1 + \frac{\omega_n}{(1-\omega_n)}R^{(i)}} \xrightarrow{\mathbb{P}} 1, \ \text{for all}\ i\in[M_n] \nonumber.
\end{eqnarray}
By the \emph{Continuous Mapping Theorem} \citep{resnick2014probability},
\begin{eqnarray}
T_i := \log \left( \frac{ \frac{\omega_n}{(1-\omega_n)R^{(i)}} +  1}{1 + \frac{\omega_n}{(1-\omega_n)}R^{(i)}} \right) \xrightarrow{\mathbb{P}} 0,\ \text{for all}\ i\in[M_n]. \nonumber
\end{eqnarray}
Without loss of generality, assume $R^{(i)}$'s, or equivalently, $T_i$'s are identically and independently distributed. Note that, $$\left| \mathrm{f}_{\rm stat,10}- \frac{1}{M_n} \sum_{i=1}^{M_n} \log(R^{(i)}) \right| = \left| \frac{1}{M_n} \sum_{i = 1}^{M_n} \log \left( \frac{ \frac{\omega_n}{(1-\omega_n)R^{(i)}} +  1}{1 + \frac{\omega_n}{(1-\omega_n)}R^{(i)}} \right) \right|.$$ 
Then,
\begin{equation}\label{iii}
\begin{split}
\mbox{\rm P}\left(\left| \mathrm{f}_{\rm stat,10} - \frac{1}{M_n} \sum_{i=1}^{M_n} \log(R^{(i)}) \right| \geq \delta \right)
&= \mbox{\rm P} \left(\left| \frac{1}{M_n} \sum_{i = 1}^{M_n} \log \left( \frac{ \frac{\omega_n}{(1-\omega_n)R^{(i)}} +  1}{1 + \frac{\omega_n}{(1-\omega_n)}R^{(i)}} \right) \right| \geq \delta \right), \\
&=\mbox{\rm P}(|T_1|\geq \delta) 
\leq \frac{\rho}{2}, \quad \ \text{for all}\ n\geq n_0. 
\end{split}
\end{equation}
Combining equations \eqref{i},\eqref{ii} and \eqref{iii} completes the proof. Similarly, it can be proven that ,
\begin{equation*}
    \left| \mathrm{H}_{\rm stat,01}(\mathbf{s}) - \frac{1}{M_n} \sum_{i=1}^{M_n} \log\left(\frac{1}{R^{(i)}}\right) \right| \xrightarrow{\mathbb{P}} 0,\ \mbox{as} \ n \rightarrow \infty.
\end{equation*}
\end{proof}

\section{Experiments: $\chi^2$ and $F$ tests}
\subsection{Test for independence in contingency tables ($\chi^2$-test)}

Suppose a confidential database contains paired binary features $(x_1, y_1), \ldots, (x_n, y_n)\in\{0,1\}\times\{0,1\}$ on $n$ individuals. The database releases response to an analyst's queries,  upon privatization via Laplace mechanism with a fixed privacy budget $\varepsilon$. Without the loss of generality, we assume $\varepsilon \in\{ 1, 1.5, 2\}$. In this context, suppose we intend  to test the hypothesis that the binary features $(X,Y)$ are independent against the hypothesis that $(X,Y)$ are not independent. We utilise differentially private Bayes factors based on $\chi^2$-statistic to accomplish this. 

We wish to compare the size $\alpha$ private Bayesian test with the size $\alpha$ non-private Bayes factor  based on $\chi^2$-statistic \citep{johnson2023bayes},  with respect to power. We consider an increasing grid of sample sizes $n\in\{500, 1000, 1500, 2000\}$. For each sample size, 
under the hypothesis that the binary features $(X,Y)$ are not independent, we generate $n$ observations independently from a Multinomial distribution with $4$ classes $\{1, 2, 3, 4\}$ with cell probabilities $$\pi_1 = \pi_0 +\Delta(1, -1, -1, 1),$$ where $\pi_0 = (1/4, 1/4, 1/4, 1/4)^{\T}$ and $\Delta\in\{0.05, 0.10, 0.15, 0.20\}$.  The  classes  $\{1, 2, 3, 4\}$ denote $(X, Y) = (0, 0), (0, 1), (1, 0)$ and $(1, 1)$ respectively.  Recall that, in practical circumstances, we shall only have access to the log Bayes factor upon privatization via Laplace mechanism with $\varepsilon$, the sample size $n$, and the effect size of interest $\pi^{\star}$.

\begin{figure}
\begin{subfigure}[t]{0.48\textwidth}
\includegraphics[width=6cm, height = 6cm]{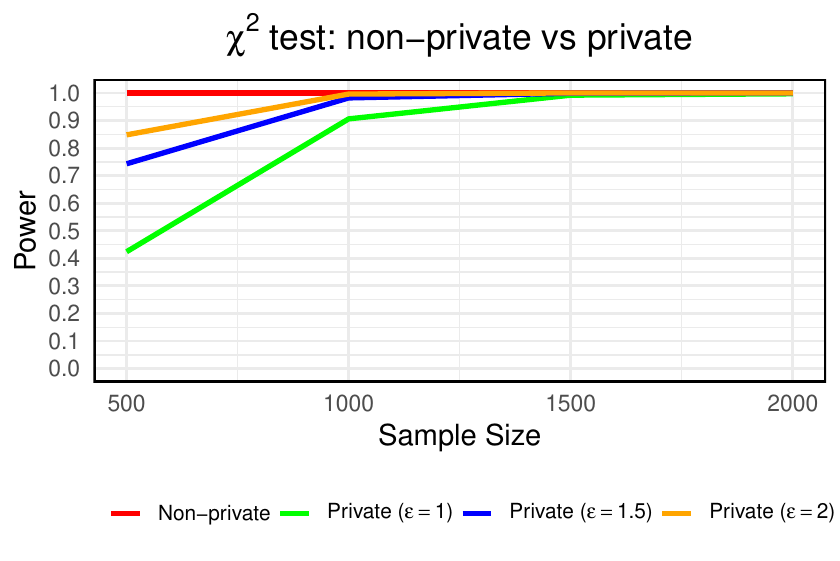} 
\caption{\textbf{Power analysis of size $\alpha$ non-private and private Bayesian $\chi^2$ test under non-local slab prior.} Comparison of the size $\alpha$ non-private Bayes factor based on $\chi^2$-statistic, and size $\alpha$ private Bayes factor based on $\chi2$-statistic with hyper-parameters set at $\hat{M}_n$ for varying values of the privacy budget $\varepsilon\in\{1, 1.5, 2\}$.}\label{fig:power_chi2}
\end{subfigure}
~
\begin{subfigure}[t]{0.48\textwidth}
\includegraphics[width=6cm, height = 6cm]{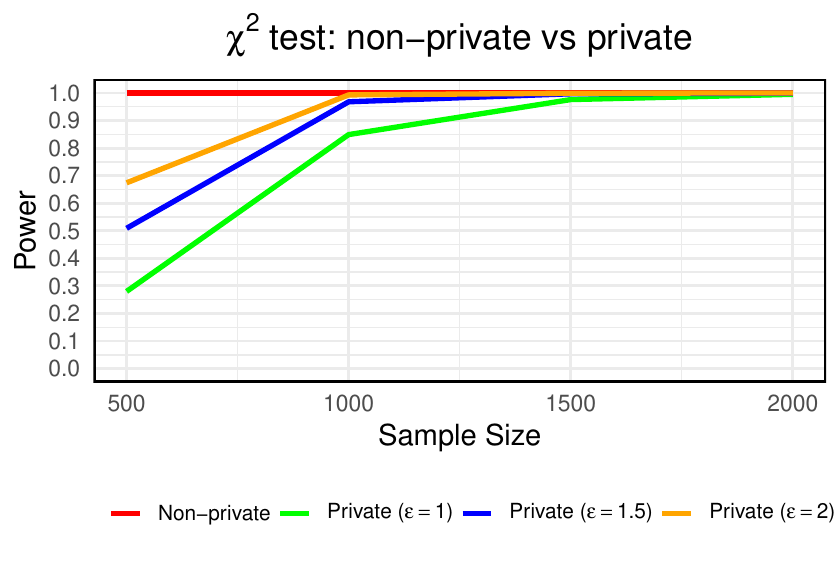} 
\caption{\textbf{Power analysis of size $\alpha$ non-private and private Bayesian $\chi^2$-test under local slab prior.} Comparison of the size $\alpha$ non-private Bayes factor based on $\chi^2$-statistic, and size $\alpha$ private Bayes factor based on $\chi2$-statistic with hyper-parameters set at $\hat{M}_n$ for varying values of the privacy budget $\varepsilon\in\{1, 1.5, 2\}$. }\label{fig:power_chi2_local}
\end{subfigure}
\caption{Comparison of non-private and private Bayesian  $\chi^2$-tests under different prior specifications and privacy budgets.}
\end{figure}

To determine the size \(\alpha\) cut-off for the differentially private Bayes factor-based test, via Algorithm 1 of the main manuscript, we introduce the standardized effect size for Multinomial tests and compute the effect size of interest. To that end, we assume that \(\pi_{1} = (\pi_{11}, \pi_{12}, \pi_{13}, \pi_{14})^{\T}\) and \(\pi_{0} =  (\pi_{01}, \pi_{02}, \pi_{03}, \pi_{04})^{\T}\), and define $u = (u_1,\ldots, u_k\})^{\T}$ such that  $u_k = \frac{\pi_{1k} - \pi_{0k}}{\sqrt{\pi_{0k}}}.$ The standardized effect size for the Multinomial test is then expressed as $\pi^\star = \sqrt{\frac{u'u}{k}}.$ For the specified values of \(\Delta \in \{0.05, 0.10, 0.15, 0.20\}\), the corresponding values of standardized effect size $\pi^\star$ are \(\{0.1, 0.2, 0.3, 0.4\}\), respectively.

For each sample size,  we carry out the hyper-parameter tuning step numerically, as described in sub-section \ref{experiments_t}. We skip  the details to avoid repetition in our exposition. The  power comparisons of the non-private and private tests  are presented in Figure \ref{fig:power_chi2}. As expected, we sacrifice on power to ensure privacy, but the difference in power of the private and non-private Bayesian tests diminishes as we increase sample size. As earlier, for a fixed sample size, as the value of the privacy budget $\varepsilon$ increases,  the loss in power to ensure privacy decreases, as expected.


Finally, we conduct a series of simulation studies to evaluate the utility of employing non-local priors in comparison to local priors.  The power comparison results are illustrated in Figure~\ref{fig:power_chi2_local}. When contrasted with the power curves under the non-local prior shown in Figure~\ref{fig:power_chi2}, it is evident that employing a non-local prior facilitates more rapid accumulation of evidence in favor of the true alternative hypothesis $\mbox{H}_1$. Consequently, this leads to substantially higher statistical power relative to the use of a local prior.

\subsection{Test for significance of linear regression ($F$-test)}

Suppose a confidential database contains information on a $p$-dimensional feature vector and a response of interest  $(x_1, y_1), \ldots, (x_n, y_n)\in\mathbf{R}^p\times\mathbf{R}$ for $n$ individuals. The database releases response to an analyst's queries,  upon privatization via Laplace mechanism with a fixed privacy budget $\varepsilon$. Without the loss of generality, we assume  $\varepsilon \in\{ 1, 1.5, 2\}$. Consider a linear regression model $y_i = \beta_0 + \beta^{\T}x_i + e_i, \ i\in[n]$, where the errors $e_i$ are independently distributed as $\mbox{N}(0, \sigma^2)$.
In this context, suppose we intend  to test whether the linear regression is significant, i.e we want to test the hypothesis that $\beta = (\beta_1,\ldots, \beta_p) = (0,0, \ldots, 0)$ versus the hypothesis that at least one of $\beta_j,\ j\in[p]$ is non-zero. We utilise differentially private Bayes factors based on $F$-statistic to accomplish this. 

We wish to compare the size $\alpha$ private Bayesian test with the non-private Bayes factor  based on $F$-statistic \citep{johnson2023bayes} using the size $\alpha$ cut-off,  with respect to power. We consider an increasing grid of sample sizes $n\in\{50, 100, 200, 500\}$. We set the number of covariates $p=2$,  set the intercept parameter $\beta_0 = 1$, and set the  regression noise variance $\sigma^2 = 0.1^2$.  For each sample size, we generate the each of covariates from univariate $\mbox{N}(0, 1)$ independently. 
Under the hypothesis that the regression is significant, we  set $(\beta_1, \beta_2)\in\{\pm 0.25, \pm 0.5, \pm 0.75, \pm 1\}^2$. Recall that, in practical circumstances, we shall only have access to a desired function of the confidential data upon privatization via Laplace mechanism with $\varepsilon$, the sample size $n$, and the   effect size of interest $\pi^{\star}$.

In order to determine the size \(\alpha\) cut-off for the differentially private Bayes factor via Algorithm 1 of the main manuscript, we first define the standardized effect size for the  linear regression problem given by, $\pi^\star = \sqrt{\frac{\beta^T(X^T X)\beta}{p\sigma^2}} = \sqrt{\frac{\beta^T(X^T X)\beta}{0.02}},$ where \(\boldsymbol{\beta} = (\beta_1, \beta_2)^T\) and \(p = 2\) is the dimension of \(\boldsymbol{\beta}\). For simplicity of exposition, we assume that $X^{T}X = I_n$, such that the expression for the standardized effect size $\pi^{\star}$ is free of the design matrix $X$. In practice, in the privatized database, the design matrix  $X$ can be orthogonalized via pre-multiplication by $\hat{\Sigma}^{-\frac{1}{2}}$, where $\hat{\Sigma} = Var(X)$. Note that, the effect size of interest is then   $\pi^\star\in\{2.5, 5, 7.5, 10\}$.

\begin{figure}
\begin{subfigure}[t]{0.48\textwidth}
\includegraphics[width=6cm, height = 5cm]{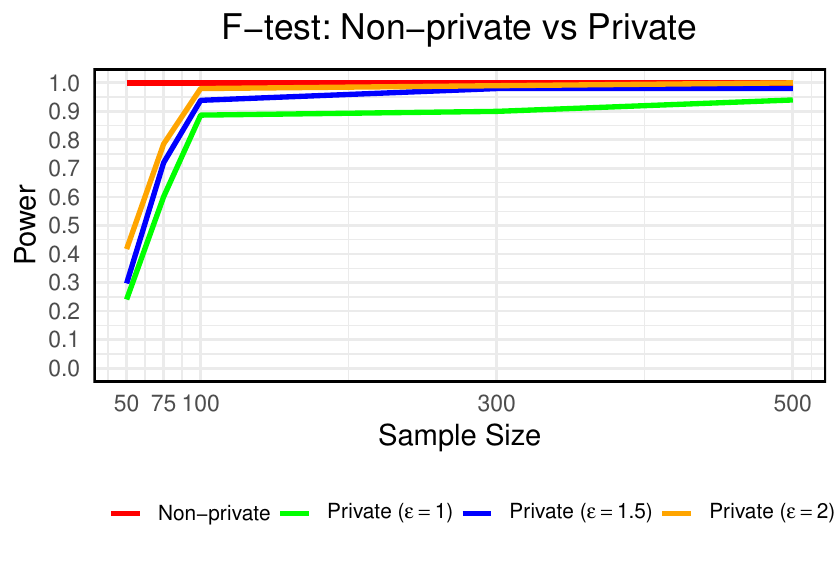} 
\caption{\textbf{Power analysis of size $\alpha$ non-private and private Bayesian $F$ test under non-local slab prior.} Comparison of the size $\alpha$ non-private Bayes factor based on $F$-statistic, and size $\alpha$ private Bayes factor based on $F$-statistic with hyper-parameters set at $\hat{M}_n$ for varying privacy budget $\varepsilon\in\{1, 1.5, 2\}$. }\label{fig:power_f}
\end{subfigure}
~
\begin{subfigure}[t]{0.48\textwidth}
\includegraphics[width=6cm, height = 5cm]{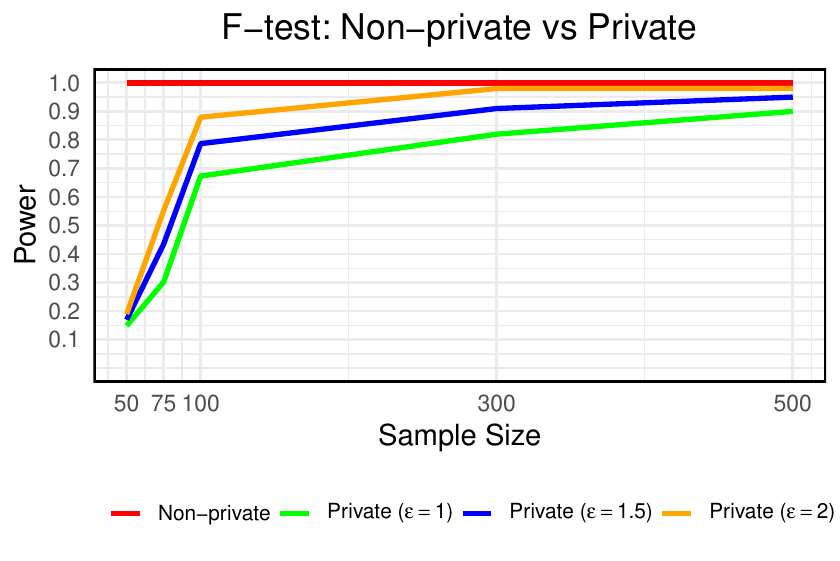} 
\caption{\textbf{Power analysis of size $\alpha$ non-private and private Bayesian $F$ test under local slab prior.} Comparison of the size $\alpha$ non-private Bayes factor based on $F$-statistic, and size $\alpha$ private Bayes factor based on $F$-statistic with hyper-parameters set at $\hat{M}_n$ for varying privacy budget $\varepsilon\in\{1, 1.5, 2\}$. }\label{fig:power_f_local}
\end{subfigure}
\caption{Comparison of non-private and private Bayesian  $F$-tests under different prior specifications and privacy budgets.}
\end{figure}

For each sample size, we  carry out the hyper-parameter tuning step numerically, as described in sub-section \ref{experiments_t}. We skip  the details for to avoid repetition in our exposition. The  power comparisons of the non-private and private tests  are presented in Figure \ref{fig:power_f}. As expected, we sacrifice on power to ensure privacy, but the difference in power of the private and non-private Bayesian tests diminishes as the sample size increases. As earlier, for a fixed sample size, as the value of the privacy budget $\varepsilon$ increases,  the loss in power to ensure privacy decreases, as expected.

Finally, we conduct a series of simulation studies to evaluate the utility of employing non-local priors in comparison to local priors. The power comparison results are illustrated in Figure~\ref{fig:power_f_local}. When contrasted with the power curves under the non-local prior shown in Figure~\ref{fig:power_f}, it is evident that employing a non-local prior facilitates more rapid accumulation of evidence in favor of the true alternative hypothesis $\mbox{H}_1$. Consequently, this leads to substantially higher statistical power relative to the use of a local prior.

\clearpage
\section{Experiment in Section 4 of the main document: plots in log scale}
\subsection{Test for Normal means ($z$ or $t$-test)}\label{experiments_t}

\begin{figure}[!htbp]
\centering

\begin{subfigure}[t]{0.4\textwidth}
\includegraphics[width=4cm, height=3cm]{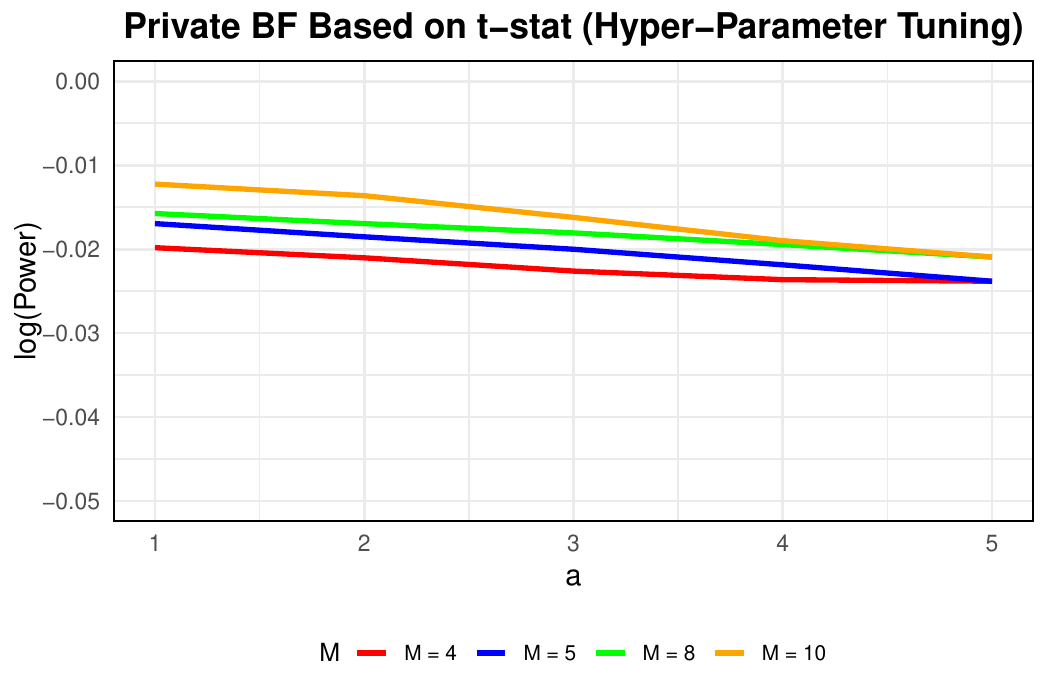} 
\caption{\textbf{Hyper-parameter tuning in size $\alpha$ Bayesian test ($t$-test).} The log-power of the size $\alpha$ privatized Bayesian test for sample size $n=100$ and privacy budget $\varepsilon = 1$,  with varying values of hyper-parameters $M_n $.}
\label{fig:hpt_t}
\end{subfigure}
\hfill
\begin{subfigure}[t]{0.4\textwidth}
\includegraphics[width=4cm, height=3cm]{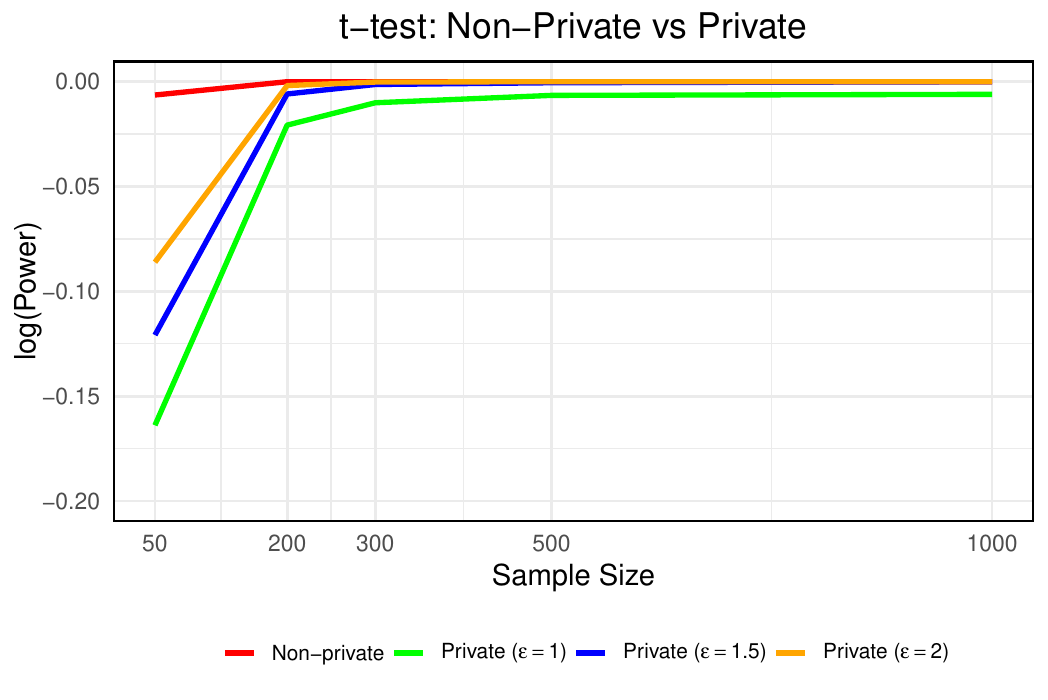} 
\caption{\textbf{Log-power analysis of size $\alpha$ non-private and private Bayesian $t$ test under non-local slab prior.} Comparison of the size $\alpha$ non-private Bayes factor based on $t$-statistic, and size $\alpha$ private Bayes factor based on $t$-statistic with hyper-parameters set at $\hat{M}_n$, for varying privacy budget $\varepsilon\in\{1, 1.5, 2\}$.}
\label{fig:power_t}
\end{subfigure}

\vspace{0.5cm}

\begin{subfigure}[t]{0.4\textwidth}
\includegraphics[width=4cm, height=3cm]{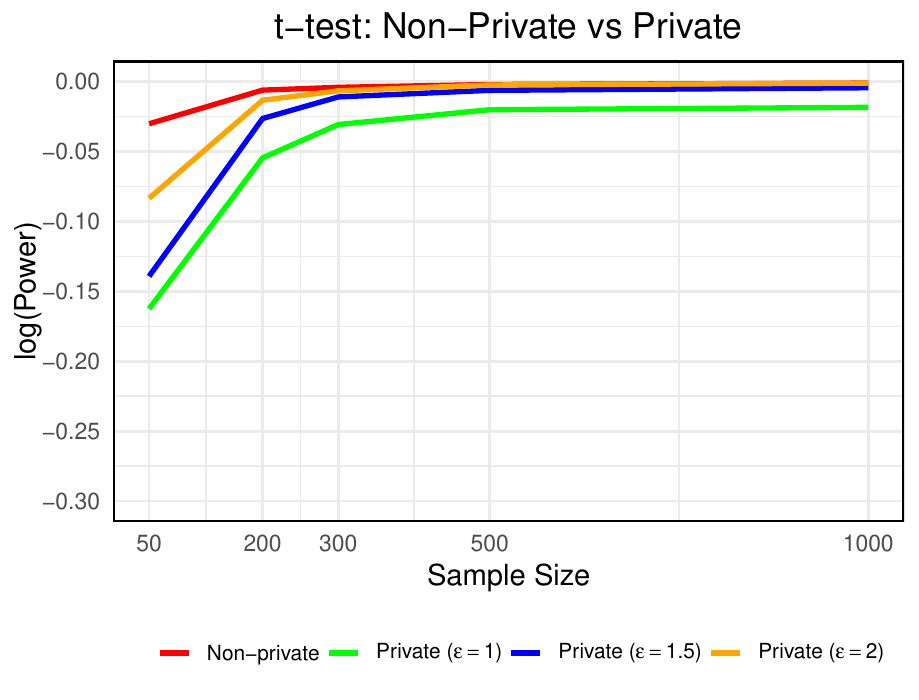}
\caption{\textbf{Log-power analysis of size $\alpha$ non-private and private Bayesian  $t$ test under non-private hypothesis.} Comparison of the size $\alpha$ non-private Bayes factor based on $t$-statistic, and size $\alpha$ private Bayes factor based on $t$-statistic with hyper-parameters set at $\hat{M}_n$, for varying privacy budget $\varepsilon\in\{1, 1.5, 2\}$.}
\label{fig:power_t_misspecified}
\end{subfigure}
\hfill
\begin{subfigure}[t]{0.4\textwidth}
\includegraphics[width=4cm, height=3cm]{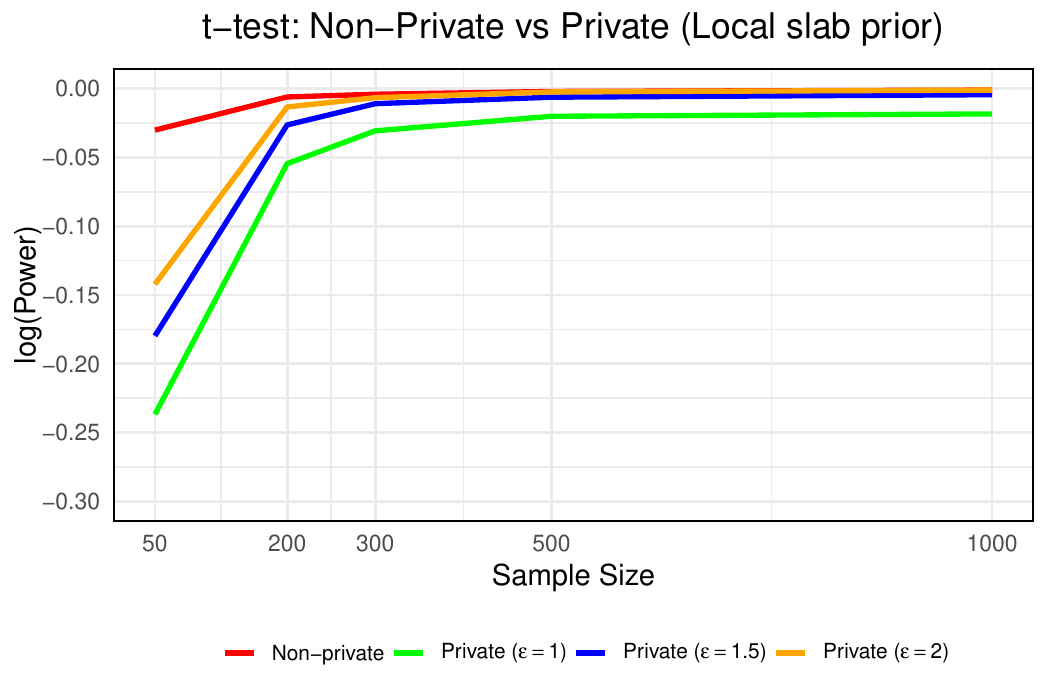}
\caption{\textbf{Log-power analysis of size $\alpha$ non-private and private Bayesian $t$ test under local slab prior.} Comparison of the size $\alpha$ non-private Bayes factor based on $t$-statistic, and size $\alpha$ private Bayes factor based on $t$-statistic with hyper-parameters set at $\hat{M}_n$, for varying privacy budget $\varepsilon\in\{1, 1.5, 2\}$.}
\label{fig:power_t_local}
\end{subfigure}

\caption{Comparison of non-private and private Bayesian $t$-tests under different prior specifications and privacy budgets.}
\label{fig:ttest_2x2}
\end{figure}

\clearpage
\subsection{Test for independence in contingency tables ($\chi^2$-test)}

\begin{figure}[!htb]
\begin{subfigure}[t]{0.48\textwidth}
\includegraphics[width=6cm, height = 6cm]{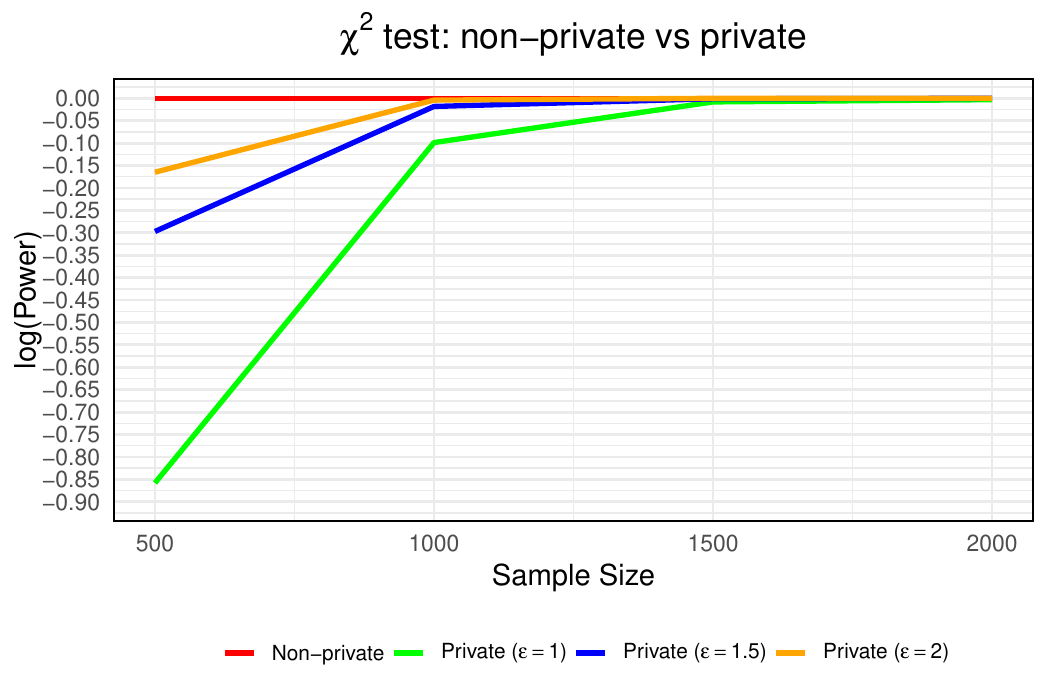} 
\caption{\textbf{Log-power analysis of size $\alpha$ non-private and private Bayesian $\chi^2$ test under non-local slab prior.} Comparison of the size $\alpha$ non-private Bayes factor based on $\chi^2$-statistic, and size $\alpha$ private Bayes factor based on $\chi2$-statistic with hyper-parameters set at $\hat{M}_n$ for varying values of the privacy budget $\varepsilon\in\{1, 1.5, 2\}$.}\label{fig:power_chi2_log}
\end{subfigure}
~
\begin{subfigure}[t]{0.48\textwidth}
\includegraphics[width=6cm, height = 6cm]{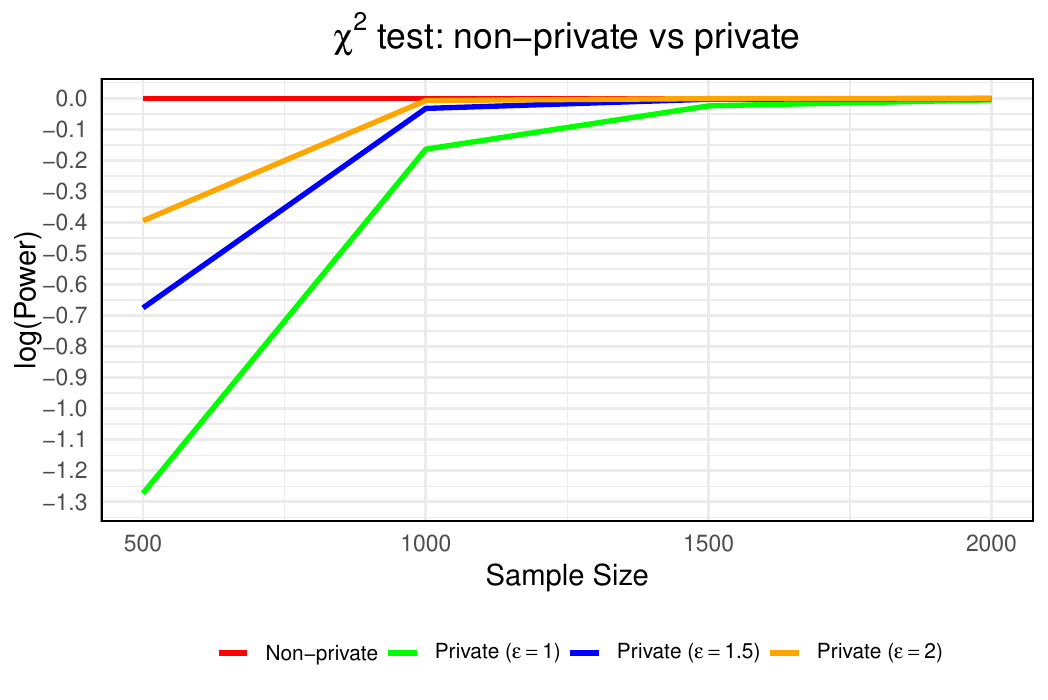} 
\caption{\textbf{Log-power analysis of size $\alpha$ non-private and private Bayesian $\chi^2$-test under local slab prior.} Comparison of the size $\alpha$ non-private Bayes factor based on $\chi^2$-statistic, and size $\alpha$ private Bayes factor based on $\chi2$-statistic with hyper-parameters set at $\hat{M}_n$ for varying values of the privacy budget $\varepsilon\in\{1, 1.5, 2\}$. }\label{fig:power_chi2_local_log}
\end{subfigure}
\caption{Comparison of non-private and private Bayesian  $\chi^2$-tests under different prior specifications and privacy budgets.}
\end{figure}

\clearpage
\subsection{Test for significance of linear regression ($F$-test)}

\begin{figure}[!htb]
\begin{subfigure}[t]{0.48\textwidth}
\includegraphics[width=6cm, height = 5cm]{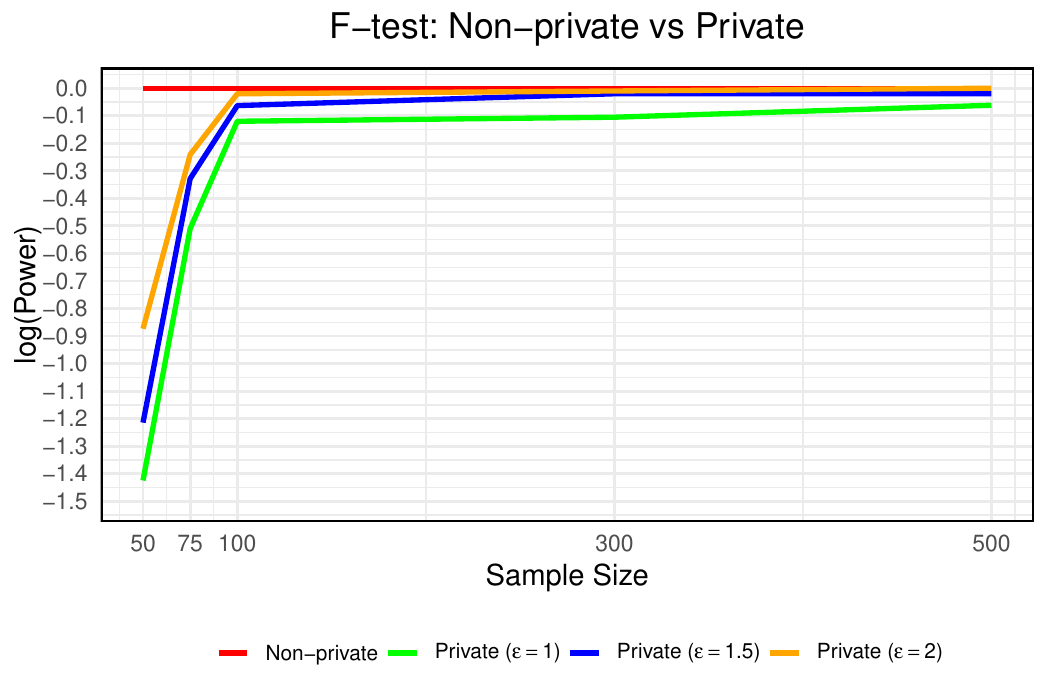} 
\caption{\textbf{Log-power analysis of size $\alpha$ non-private and private Bayesian $F$ test under non-local slab prior.} Comparison of the size $\alpha$ non-private Bayes factor based on $F$-statistic, and size $\alpha$ private Bayes factor based on $F$-statistic with hyper-parameters set at $\hat{M}_n$ for varying privacy budget $\varepsilon\in\{1, 1.5, 2\}$. }\label{fig:power_f_log}
\end{subfigure}
~
\begin{subfigure}[t]{0.48\textwidth}
\includegraphics[width=6cm, height = 5cm]{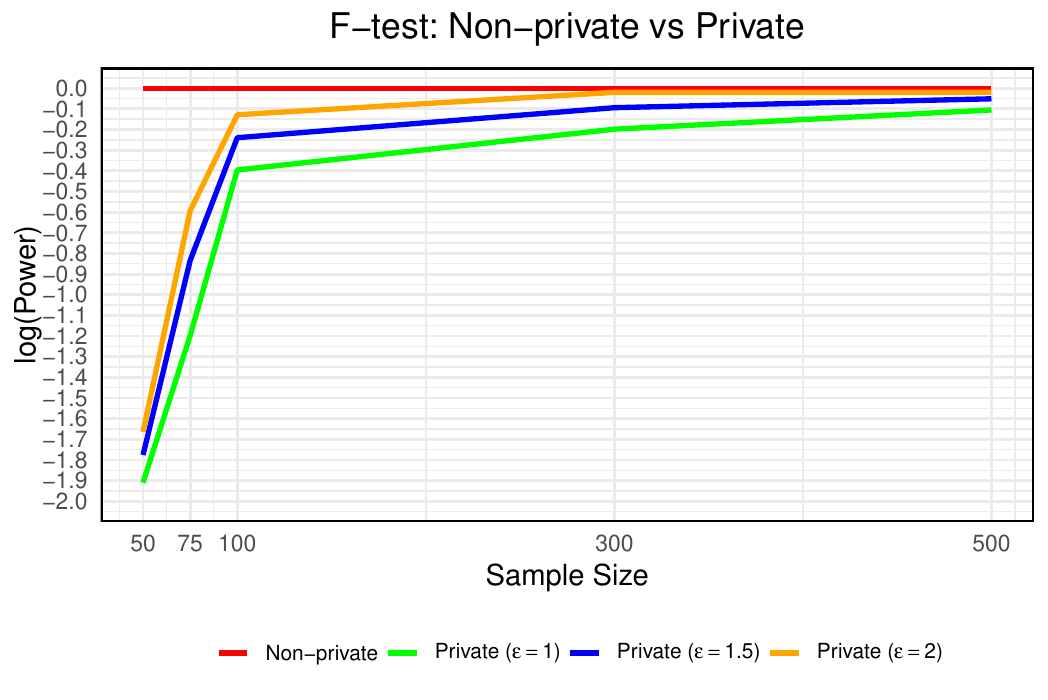} 
\caption{\textbf{Log-power analysis of size $\alpha$ non-private and private Bayesian $F$ test under local slab prior.} Comparison of the size $\alpha$ non-private Bayes factor based on $F$-statistic, and size $\alpha$ private Bayes factor based on $F$-statistic with hyper-parameters set at $\hat{M}_n$ for varying privacy budget $\varepsilon\in\{1, 1.5, 2\}$. }\label{fig:power_f_local_log}
\end{subfigure}
\caption{Comparison of non-private and private Bayesian  $F$-tests under different prior specifications and privacy budgets.}
\end{figure}

\clearpage

\section{Application}

\begin{figure}[!htb]
\begin{center}
\includegraphics[width=12cm, height =7.0cm]{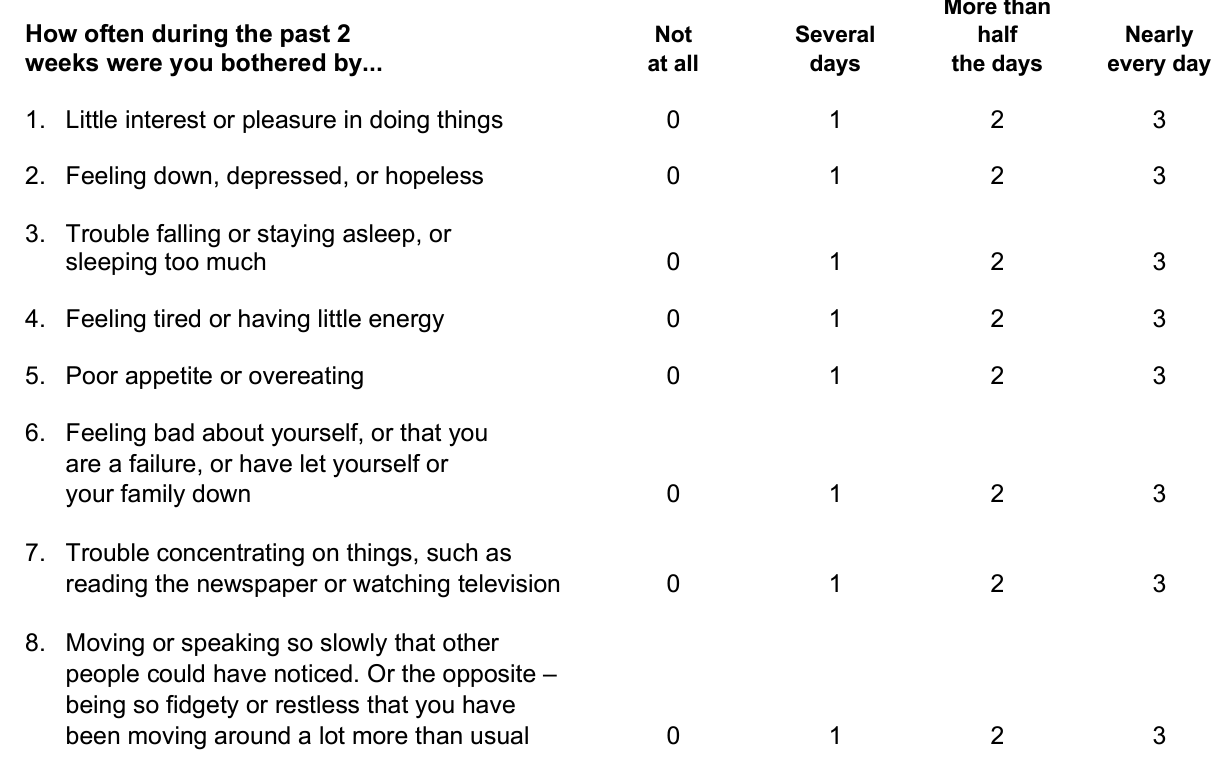}
\centering{{\caption{\emph{\textbf{Personal Health Questionnaire Depression Scale  (PHQ-8).} The PHQ-8  score is the sum of the 8 items. The PHQ-8 scores range from 0 to 24, with the following categorical interpretations: 0–4 (none/minimal depression), 5–9 (mild depression), 10–14 (moderate depression), 15–19 (moderately severe depression), and 20–24 (severe depression). }}\label{PHQ8}}}.
\end{center}
\end{figure}
\clearpage
\bibliographystyle{ba}
\bibliography{paper-ref,references}
